\documentclass[10pt]{article}

\usepackage{newtxtext,newtxmath}
\usepackage{booktabs}
\usepackage{siunitx}        % number formatting
\usepackage{longtable}
\usepackage{threeparttable}
\usepackage{caption}
\usepackage{pgfplotstable}  % CSV -> table
\pgfplotsset{compat=1.18}
\usepackage{graphicx,svg}
\usepackage{subcaption} % in preamble
\usepackage{booktabs}
\usepackage{subcaption} % in preamble
\usepackage{caption}
\usepackage{times}
\usepackage{url}
\usepackage[hidelinks]{hyperref}

\usepackage[letterpaper,margin=1in]{geometry}
\usepackage{booktabs}
\usepackage{multirow}
\usepackage{array}
\usepackage{tikz}
\usetikzlibrary{calc}
\usetikzlibrary{shapes}

\usepackage{amssymb} % for \checkmark
\DeclareUnicodeCharacter{2212}{-}

\usepackage{setspace}
\usepackage[font=small,labelfont=bf]{caption}

\DeclareCaptionFont{singlespacing}{\setstretch{1}}  % define a caption font that forces single spacing
\renewenvironment{abstract}
    {\quotation}
{\endquotation}

\date{}

\makeatletter
\renewcommand{\fnum@figure}{\textbf{Figure \thefigure}}
\renewcommand{\fnum@table}{\textbf{Table \thetable}}
\makeatother

\usepackage{scicite}
\usepackage{url}

\usepackage{xurl}
\def\scititle{Readers Prefer Outputs of AI Trained on Copyrighted Books over Expert Human Writers}
\title{\bfseries \boldmath \scititle}

\author{
	Tuhin Chakrabarty$^{1}$, 
	Jane C. Ginsburg$^{2}$,
	Paramveer Dhillon$^{3,4}$\and
	\small$^{1}$Department of Computer Science, Stony Brook University.\\
        \small$^{2}$Columbia Law School.\\
	\small$^{3}$School of Information Science, University of Michigan.\\
    \small$^{4}$MIT Initiative on the Digital Economy.\and
	\small Corresponding authors: tchakrabarty@cs.stonybrook.edu, ginsburg@law.columbia.edu,\\ \small dhillonp@umich.edu
}

\begin{document} 

\maketitle

\begin{abstract} \bfseries \boldmath
The use of copyrighted books for training AI models has led to numerous lawsuits from authors concerned about AI's ability to generate derivative content. Yet it's unclear whether these models can generate high quality literary text while emulating authors' styles/voices. To answer this we conducted a preregistered study comparing MFA-trained expert writers with three frontier AI models: ChatGPT, Claude, and Gemini in writing up to 450-word excerpts emulating 50 award-winning authors' (including Nobel laureates, Booker Prize winners, and young emerging National Book Award finalists) diverse styles. In blind pairwise evaluations by 28 MFA-trained readers (writers from top U.S. writing programs) and 516 college-educated general readers (recruited via Prolific), AI-generated text from in-context prompting was strongly disfavored by MFA-trained readers for both stylistic fidelity (odds ratio [OR]=0.16) and writing quality (OR=0.13), while college-educated general readers showed no significant preference on stylistic fidelity (OR=1.06) but favored AI-generated text for writing quality (OR=1.82). However, fine-tuning ChatGPT on individual author's complete works reversed these findings: MFA-trained readers now favored AI-generated text for stylistic fidelity (OR=8.16) and writing quality (OR=1.87), while college-educated general readers showed even stronger AI preference (stylistic fidelity OR=16.65; writing quality OR=5.42). Both reader groups preferred fine-tuned AI, but the writer-type $\times$ reader-type interaction remained significant after fine-tuning ($p=0.021$ for fidelity; $p<10^{-4}$ for quality), indicating that college-educated general readers favored AI-generated text by a wider margin than MFA-trained readers. These effects are robust under cluster-robust inference and generalize across authors and styles in author-level heterogeneity analyses. The fine-tuned outputs were rarely flagged as AI-generated (3\% rate versus 97\% for in-context prompting) by state-of-the-art AI detectors. Mediation analysis reveals this reversal occurs because fine-tuning eliminates detectable AI stylistic quirks (e.g., clich\'{e} density) that penalize in-context outputs, altering the relationship between AI detectability and reader preference. While we do not account for additional costs of human effort required to transform raw AI output into cohesive, publishable novel-length prose, the median fine-tuning and inference cost of \$81 per author represents a dramatic 99.7\% reduction compared to typical professional writer compensation. Author-specific fine-tuning thus enables non-verbatim AI writing that readers prefer to expert human writing, thereby providing empirical evidence directly relevant to copyright's fourth fair-use factor, the ``effect upon the potential market or value'' of the source works.

\end{abstract}
\noindent

\textit{Keywords: Generative AI, Copyright Law, Fair Use, Future of Work, AI Detection, AI and Society, Behavioral Science, Labor Market Impact}
\newpage

% \section{Introduction}
The U.S. publishing industry supports hundreds of thousands of jobs while generating \$30 billion in yearly revenue, contributing to the larger American copyright sectors that account for \$2.09 trillion in annual GDP contributions \cite{aap_industry_statistics_2025}. Adult Fiction and Non-Fiction books alone accounted for \$6.14 billion in 2024 \cite{PW2025}. This economically important sector now faces an unprecedented challenge: its core products have become essential training data for generative-AI systems. Models trained on well-edited books produce more coherent, accurate responses\textemdash something crucial to creating the illusion of intelligence \cite{reisner_books3_2023}. Most technology companies building AI use massive datasets of books, typically without permission or licensing \cite{samuelson2023generative}, and frequently from illegal sources. In the recent copyright lawsuit \textit{Bartz vs Anthropic} \cite{usdcc_ndca_bartz_anthropic_2025}, Judge Alsup noted that Anthropic acquired at least five million books from LibGen and two million from Pirate Library Mirror (PiLiMi). Anthropic also used Books3 (a dataset of approximately 191,000 books also used by Meta and Bloomberg to train their language models), now removed after a legal complaint by anti-piracy group, the Rights Alliance. The Bartz v. Anthropic lawsuit also revealed how Anthropic cut millions of print books from their bindings, scanned them into digital files, and threw away the originals solely for the purpose of training Claude. This unauthorized use has sparked outrage among authors \cite{gero2025creative}, triggering dozens of lawsuits against technology companies including OpenAI, Anthropic, Microsoft, Google, and Meta.

Generative AI systems such as ChatGPT that can be prompted to create new text at scale are qualitatively unlike most historical examples of automation technologies \cite{noy2023experimental}. They can now solve Olympiad-level Geometry \cite{trinh2024solving}, achieve an impressive rating of 2700 on Codeforces, one of the most challenging coding competition platforms \cite{codeforces2025}, and deliver medical guidance that meets professional healthcare standards \cite{kolata_chatbots_2024,openai_introducing_healthbench_2025}. Recent findings from Microsoft's Occupational Implications of Generative AI \cite{tomlinson2025working}, Anthropic's Economic Index \cite{handa2025economic} and OpenAI \cite{NBERw34255} reveal AI usage primarily concentrates in writing tasks. This concentration threatens creative writing professionals in particular\textemdash novelists, poets, screenwriters, and content creators who shape cultural narratives and human expression. Based on U.S. Bureau of Labor Statistics May 2023 national estimates, creative writing constitutes almost 50\% of writing jobs \cite{bls2023}, making these positions especially vulnerable to GenAI-based automation, as the writing community has warned \cite{lithub2024}.

While it is widely established that most state-of-the-art Large Language Models (LLMs) have been trained on copyrighted books, it remains unclear whether such training can produce expert-level creative writing. Past research has shown that AI cannot produce high-brow literary fiction or creative nonfiction through prompting alone when compared to professionally trained writers \cite{chakrabarty2024art}. More recent work demonstrates that AI-generated creative writing still remains characterized by clichés, purple prose, and unnecessary exposition~\cite{chakrabarty2025can}. Additionally, relying on Generative AI for creative writing reduces the collective diversity of novel content \cite{doshi2024generative}. AI often produces formulaic, mediocre creative writing because it lacks the distinctive personal voice that typically distinguishes one author from another \cite{laquintano2024ai,walsh2024does}. As Pulitzer Fiction finalist Vauhini Vara observes, {\it ``ChatGPT's voice is polite, predictable, inoffensive, upbeat. Great characters, on the other hand, aren't polite; great plots aren't predictable; great style isn't inoffensive; and great endings aren't upbeat''} \cite{Vara2023}. To address this limitation, practitioners now increasingly prompt AI systems to perform style/voice mimicry by emulating specific writers' choices \cite{chiang_why_2024}. This practice has become so common that a fantasy author recently published a novel containing an accidentally included AI prompt requesting emulation of another writer's style \cite{Tangermann2025}. While the effectiveness of such stylistic emulation remains contested, the more pressing question concerns whether style/voice mimicry genuinely improves AI-generated text quality and whether MFA-trained readers and college-educated general readers perceive these improvements as meaningful.

To address this question, we conducted a preregistered behavioral study comparing a representative sample of MFA-trained expert writers with state-of-the-art LLMs.\footnote{In the United States, a Master of Fine Arts (MFA) is typically recognized as a terminal degree for practitioners of creative writing (equivalent to a Ph.D.)---meaning that it is considered the highest degree in its field, qualifying an individual to become a professor at the university level in these disciplines.} MFA-trained writers are broadly representative of the universe of expert writers. Not all professional writers attended MFA programs, but enough did to stand in for the wider group. Historically, several award-winning authors who have written great American Novels~\cite{pw_literary_prizes} have graduated with an MFA degree in Creative Writing (See Figure~\ref{fig:mfalist} in SI) and especially from the programs we recruited our writers for this study (See Figure~\ref{fig:mfa_by_school} in SI). Eminent literary agent Gail Hochman of Brandt \& Hochman said, {\it``We look favorably on anyone who has an M.F.A., simply because it shows they're serious about their writing''~\cite{delaney2007where}.} In June 2010, the editors of The New Yorker Magazine announced to widespread media coverage their selection of 20 under 40\footnote{\url{https://www.newyorker.com/magazine/20-under-40-fiction}}---the young fiction writers who are, or will be, central to their generation; 17 out of those 20 writers had an MFA degree (See Fig~\ref{fig:MFAbreakdown} in SI). This elite sample of MFA trained expert writers provides a conservative test---if AI can compete with the best emerging talent, the disruption to average writers is likely even greater.

We selected closed-source LLMs readily accessible to users without technical expertise: GPT-4o, Claude~3.5 Sonnet, and Gemini~1.5 Pro. We also tried open-weights Llama~3.1 model but its empirical performance was not good at following long context instructions at the time of our study. Both human experts and LLMs were given the same task: write an excerpt of up to 450 words emulating the style and voice of one of 50 internationally acclaimed authors, including Nobel laureates, Booker Prize winners, and Pulitzer Prize winners, spanning multiple continents and cultures. Among others, our study includes Nobel laureates Han Kang and Annie Ernaux; Booker Prize winners Salman Rushdie, Margaret Atwood, and George Saunders; and Pulitzer Prize winners Junot Díaz and Marilynne Robinson. Beyond style emulation, our writing task requires originality under explicit content constraints. Although both AI and human writers received prompts with detailed content specifications, composing literary fiction/creative non-fiction still requires agency especially in deliberate choices of words, syntax, voice, and narrative framing to produce novel, coherent prose. Because literary fiction/creative non-fiction is inherently non-formulaic, the writing task extends beyond style imitation to creative composition, making it different from mere parody or pastiche.

Our study examines shorter (up to 450-word) excerpts for two reasons. First, an excerpt of this length isolates the voice, style, and quality that writers must sustain across any longer work. Second, in its current form, an LLM cannot write long-form narratives that automatically balance coherence, quality, thematic consistency, and plot development across thousands of words. But it is merely a matter of time before LLMs reach that capacity. It should also be noted that human novel-writing is itself iterative and compositional, making excerpt-level quality a meaningful building block. With human steering and iterative prompting, it is feasible to produce long-form fiction and non-fiction using shorter model-generated excerpts. This is already happening in real life with startups such as Sudowrite or Genre Fiction publisher Inkitt~\cite{vara_ai_romance_factory_2025} specifically focusing on helping consumers write full books using AI. In self-publishing marketplaces such as Kindle, these AI-generated books have already gained popularity~\cite{alter2026romance, jones_ai_knockoffs_2025, knibbs_scammy_ai_books_2024}.

\begin{figure*}[!htbp]
    \centering
    \includegraphics[width=0.9\textwidth]{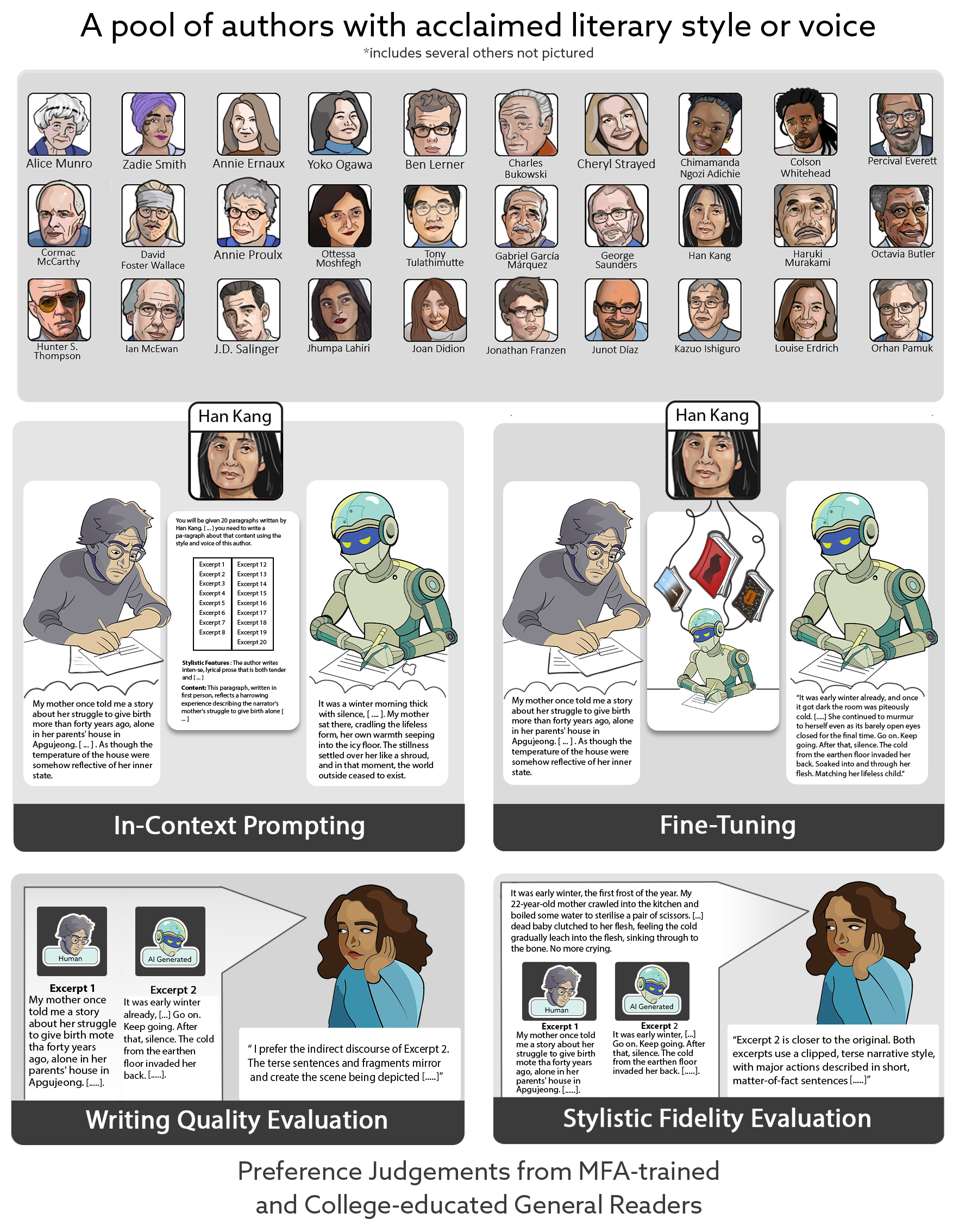} 
     \captionsetup{font={singlespacing,small},labelfont=bf}
     \caption{Figure showing our study design. (1) Select a target author and prompt. (2) Generate up to 450-word candidate excerpts from MFA experts and from LLMs under two settings: in-context prompting (instructions + few-shot examples) and author-specific fine-tuning (model fine-tuned on that author’s works). (3) Readers (MFA-trained and college-educated) perform blinded, pairwise forced-choice evaluations on two outcomes: stylistic fidelity to the target author and overall writing quality. Pair order and left/right placement are randomized on every trial.}
    \label{fig:schematic}
\end{figure*}

We tested two AI conditions: (1) in-context prompting, where models received the same instructions as human experts, and (2) fine-tuning, where models were additionally trained on each author's complete oeuvre. For authors writing in non-English languages (Han Kang, Yoko Ogawa, Annie Ernaux, Haruki Murakami), we used the same translator's work across all books to maintain voice consistency. Our MFA-trained expert writers who wrote excerpts also evaluated submissions in our study (though they never evaluated their own excerpts). We selected these writers as expert readers because their degrees serve as a strong proxy for quality in both creation and evaluation: a good writer is also likely to be a good reader, capable of discerning quality in others' writings. In addition to MFA-trained readers we also recruited college-educated general readers from Prolific. Both groups of readers performed blind pairwise evaluations~\cite{li-etal-2024-disclosure, sarkar2025ai, horton2023bias} of $\langle$Human-AI$\rangle$ excerpts on writing quality and stylistic fidelity. This design addresses three preregistered research questions: (1)~Can AI match expert performance in writing quality and stylistic fidelity across both conditions? (2)~Do both groups of readers show similar preference patterns? (3)~Does AI detectability correlate with human quality judgments, and does fine-tuning remove this correlation? Our full experimental setup is shown in Figure~\ref{fig:schematic}.

\section*{Results}

\begin{figure*}[!htbp]
    \centering
    \includegraphics[width=\textwidth]{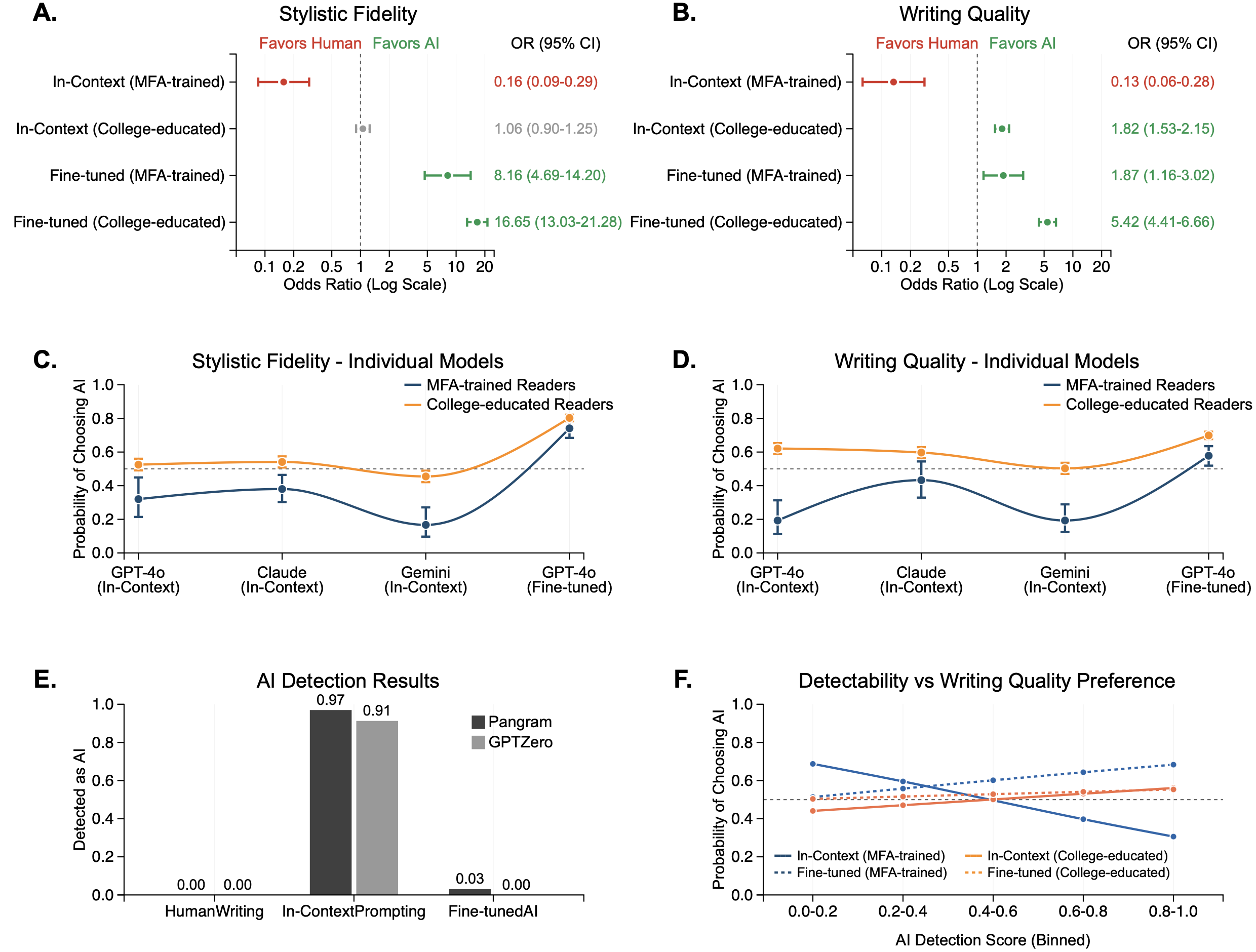}  % Note: no .svg extension
     \captionsetup{font={singlespacing,small},labelfont=bf}
     \caption{
     (A-B) Forest plots showing odds ratios (OR) and 95\% confidence intervals comparing AI-generated and human-written excerpts in pairwise evaluations of stylistic fidelity (A) and writing quality (B) where values $>$1 favor AI and values $<$1 favor humans. MFA-trained readers show preference for human writing when prompted in an in-context setting (OR = 0.16 and 0.13) but that changes when AI is fine-tuned (OR = 8.16 and 1.87). College-educated general readers show a distinct preference pattern rather than mere noise: under in-context prompting they are at parity with humans on stylistic fidelity (OR = 1.06) but prefer AI-generated text on writing quality (OR = 1.82). (C-D) Probability of choosing AI excerpts across individual language models for stylistic fidelity (C) and writing quality (D). Error bars represent 95\% confidence intervals. Dashed line indicates chance performance (50\%). (E) AI detection accuracy with chosen threshold of $\tau$=0.9 using two state-of-the-art AI detectors (Pangram and GPTZero). Human written text was never misclassified (0.00), in-context AI was detected with 97\% accuracy by Pangram and 91\% by GPTZero, but fine-tuned AI evaded detection 97\% of the time (0.03) for Pangram and 100\% of the time in case of GPTZero. (F) Relationship between AI detectability (Pangram) and preference for writing quality across detection score bins. For the in-context prompting setup, higher detection scores predict lower AI preference mainly among MFA-trained readers. After fine-tuning, that negative detectability penalty is strongly attenuated, and the model-implied slopes are flat-to-positive. Fine-tuning on an author's complete oeuvre gets rid of AI quirks while achieving expert-level performance. n = 28 MFA-trained readers, 516 college-educated general readers; 10,920 reader-level pairwise judgments/evaluations.
}

    \label{fig:preferences}
\end{figure*}

\subsection*{Overall Performance Comparisons}
Our final data consist of 10,920 reader-level pairwise judgments/evaluations, with judgments from 28 MFA-trained readers and 516 college-educated general readers. We fit a logit model for each outcome and condition and include dummies for writer type, reader group, and their interaction. We further employ CR2 cluster robust standard errors  \cite{pustejovsky2018small} clustered at the reader-level to account for within-reader correlation in ratings. Our hypotheses, outcomes, design, and analysis closely follow our OSF pre-registration (SI Sections S4-S8); minor deviations are detailed in SI Section S9.

Figure 2A–B presents the odds ratios, with corresponding predicted probabilities shown in Figure 2C–D. Under in-context prompting, MFA-trained readers demonstrated strong preference for human-written text. Odds ratios were 0.16 (95\% CI: 0.08--0.29, $p < 10^{-8}$) for stylistic fidelity and 0.13 (95\% CI: 0.06--0.28, $p < 10^{-6}$) for writing quality, indicating six- to eight-fold preferences for human excerpts (Fig.~2A--B). College-educated general readers showed no significant preference regarding stylistic fidelity (OR = 1.06, 95\% CI: 0.90--1.25, $p = 0.50$) but clearly favored AI-generated text for writing quality (OR = 1.82, 95\% CI: 1.53--2.15, $p < 10^{-10}$), selecting AI excerpts in 50--62\% of writing quality trials depending on the model (Fig.~2C--D). Inter-rater agreement reflected this divergence: MFA-trained readers achieved $\kappa = 0.58$ for stylistic fidelity and $\kappa = 0.41$ for writing quality, while college-educated readers showed minimal agreement among themselves ($\kappa = 0.10$ and $\kappa = 0.08$, respectively). Given that college-educated general readers' assessments of literary quality and style reflect inherently diverse tastes, we anticipated lower inter-rater agreement compared to MFA-trained evaluators. The writer-type $\times$ reader-type interaction was significant for both outcomes ($\chi^{2}_{(3)} = 36.6$, $p = 5.6 \times 10^{-8}$ for fidelity; $\chi^{2}_{(3)} = 46.9$, $p = 3.7 \times 10^{-10}$ for quality).

Fine-tuning on authors' complete works reversed these preferences. For MFA-trained readers, the \emph{odds} of selecting the AI excerpt were 8.16 times the odds of selecting the human excerpt for stylistic fidelity (OR \(=8.16\), 95\% CI: 4.69--14.2, \(p<10^{-12}\)) and 1.87 times the odds for writing quality (OR \(=1.87\), 95\% CI: 1.16--3.02, \(p=0.010\)). College-educated general readers showed even larger shifts in their preferences (stylistic fidelity OR = 16.65, 95\% CI: 13.03--21.28, $p < 10^{-15}$; writing quality OR = 5.42, 95\% CI: 4.41--6.66, $p < 10^{-15}$). Unlike the in-context condition, both reader groups now favored fine-tuned AI, but the magnitude of this preference differed: model-based predicted AI win probabilities were 0.74 for MFA-trained readers and 0.80 for college-educated general readers on stylistic fidelity, and 0.58 versus 0.70 on writing quality (Fig.~2C--D). The writer-type $\times$ reader-type interaction remained significant in the fine-tuned models (stylistic fidelity $\chi^{2}_{(1)} = 5.33$, $p = 0.021$; writing quality $\chi^{2}_{(1)} = 15.98$, $p = 6.4 \times 10^{-5}$), indicating that college-educated general readers favored AI-generated text by a wider margin than MFA-trained readers even after fine-tuning. Inter-rater agreement among MFA-trained readers increased ($\kappa = 0.67$ for writing quality; $\kappa = 0.54$ for stylistic fidelity), while agreement among college-educated general readers remained low ($\kappa = 0.12$ and $\kappa = 0.26$, respectively).

\begin{figure*}[!htbp]
    \centering
    \includegraphics[width=\textwidth]{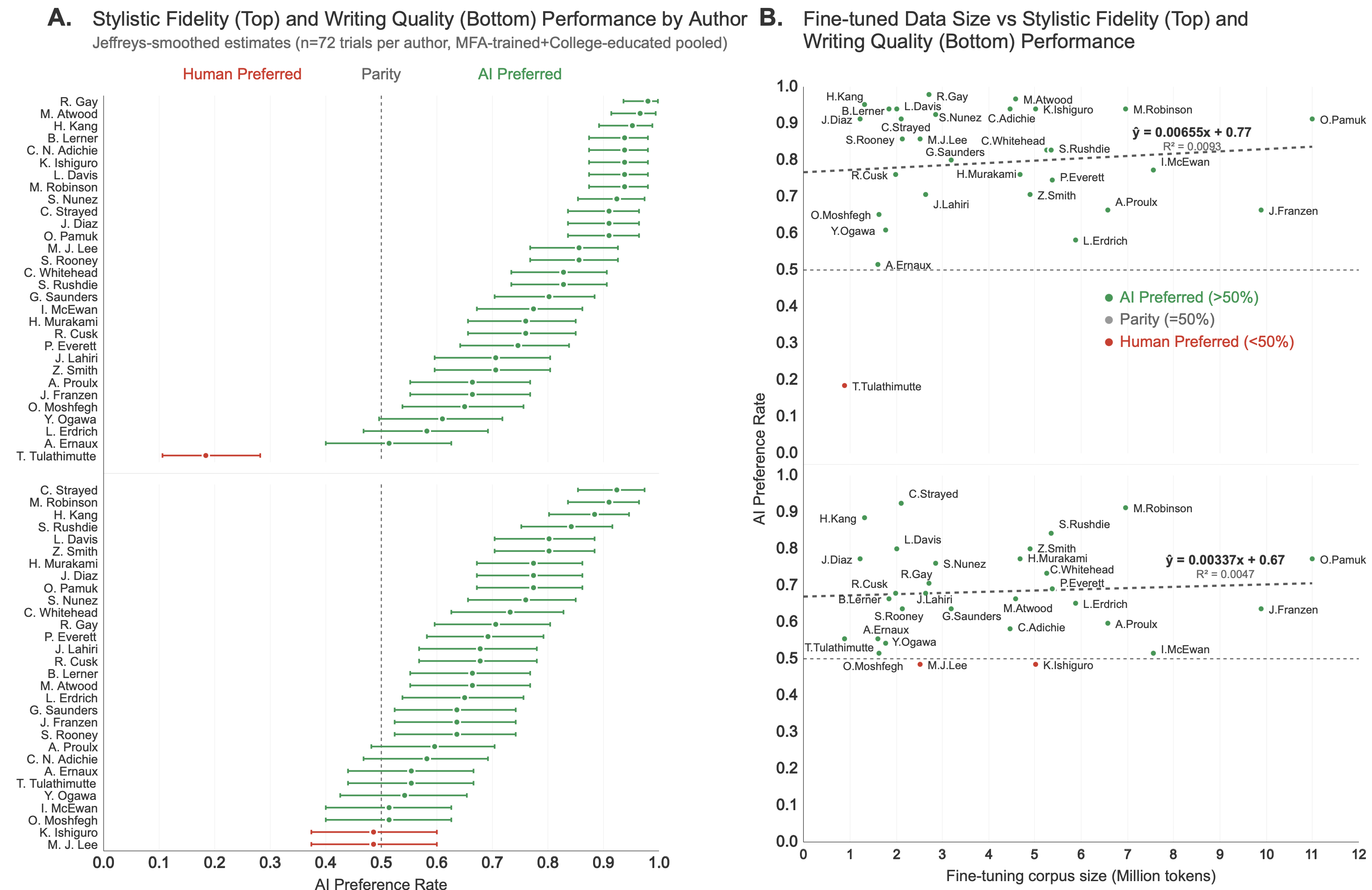}  % Note: no .svg extension
     \captionsetup{font={singlespacing,small},labelfont=bf}
     \caption{{\bf Author-level AI preference and its association with fine-tuning corpus size (fidelity and quality)}
(A) For each fine-tuned author, the share of blinded pairwise trials in which the AI excerpt was preferred over the human (MFA expert) on stylistic fidelity (top) and overall quality (bottom). Points show Jeffreys-prior estimates $(k+0.5)/(n+1)$; vertical bars are 95\% Jeffreys intervals (Beta$(\tfrac12,\tfrac12)$); the dotted line at 0.5 marks human–AI parity. Readers are pooled (MFA-trained and college-educated general readers).
(B) AI preference rate versus the fine-tuning corpus size for that author (million tokens), shown for stylistic fidelity (top) and overall quality (bottom). Each point is a fine-tuned author; the line is an OLS fit with heteroskedasticity-robust standard errors (no CI displayed). Slopes are near zero in both panels, indicating little association between corpus size (in this range) and AI preference.
}\label{fig:author-analysis}
\vspace{-2ex}
\end{figure*}

\subsection*{AI Detection and Stylometric Analysis}
We probe whether differences in AI detectability can account for these preference reversals. Pangram, a state-of-the-art AI detection tool \cite{russell2025people,jabarian_imas_2025_artificial,Naddaf2025NatureLLMDetection}, correctly classified 97\% of in-context prompted texts as machine-generated but only 3\% of fine-tuned texts were classified as AI-generated (Fig.~2E).\\GPTZero, another state-of-the-art AI detection, tool showed comparable performance but had higher false-positive rate, so we used Pangram for our subsequent analyses.

Higher AI-detection scores strongly predicted lower preference rates among MFA-trained readers in the in-context prompting condition. For stylistic fidelity, each unit increase in detection score reduced the odds of selecting an excerpt by a factor of 6.3 ($\beta = -1.85 \pm 0.29$, $p < 10^{-9}$); a similar pattern held for writing quality ($\beta = -2.01 \pm 0.33$, $p < 10^{-8}$). College-educated general readers were substantially less sensitive to detectability even before fine-tuning (Pangram $\times$ Reader Type for style: $\beta = 1.88 \pm 0.30$, $p < 10^{-9}$; for quality: $\beta = 2.61 \pm 0.34$, $p < 10^{-13}$). Fine-tuning attenuated the negative relationship between detectability and preference (Pangram $\times$ Setting for style: $\beta = 2.56 \pm 0.81$, $p = 0.002$; for quality: $\beta = 2.90 \pm 0.88$, $p < 0.001$), though the degree of attenuation differed across reader groups, as indicated by significant three-way interactions (Pangram $\times$ Setting $\times$ Reader Type for style: $\beta = -1.77 \pm 0.84$, $p = 0.036$; for quality: $\beta = -3.26 \pm 0.91$, $p < 0.001$). Two-stage mediation analysis (Fig.~4A) demonstrated that among MFA-trained readers, where the detectability penalty is concentrated, stylometric features, particularly clich\'{e} density (See Section S3.2 in SI), mediated 16.4\% of the detection effect on preference before fine-tuning but a statistically insignificant 1.3\% (95\% CI includes zero) afterward, indicating that fine-tuning reduces rather than merely masks artificial stylistic signatures. Because college-educated general readers did not penalize AI-detectable text in either condition, mediation analysis does not apply to this group.

\subsection*{Author-Level Performance Heterogeneity}
Next, we disaggregated our data to unpack heterogeneity at the level of individual authors. Of 30 fine-tuned author models, 29 exceeded parity for stylistic fidelity (median win rate = 0.83, IQR: 0.70--0.94) and 28 for writing quality (median = 0.67, IQR: 0.58--0.77). Using 95\% Jeffreys intervals, fine-tuned models significantly outperformed human writers for 26 authors on stylistic fidelity and 21 authors on writing quality (Fig.~3A). These performance differences showed no systematic relationship with fine-tuning corpus size (Fig. 3B). The fine-tuning premium, i.e., the increase in AI preference rates relative to in-context prompting, ranged from $-12.2$ to $+67.4$ percentage points for stylistic fidelity (29 of 30 positive) and from $-14.1$ to $+46.1$ percentage points for writing quality (24 of 30 positive). This ``premium’’ likewise showed no correlation with fine-tuning corpus size (Pearson $r < 0.1$ for both outcomes; Fig.~4B).

\begin{figure*}[!htbp]
    \centering
    \includegraphics[width=\textwidth]{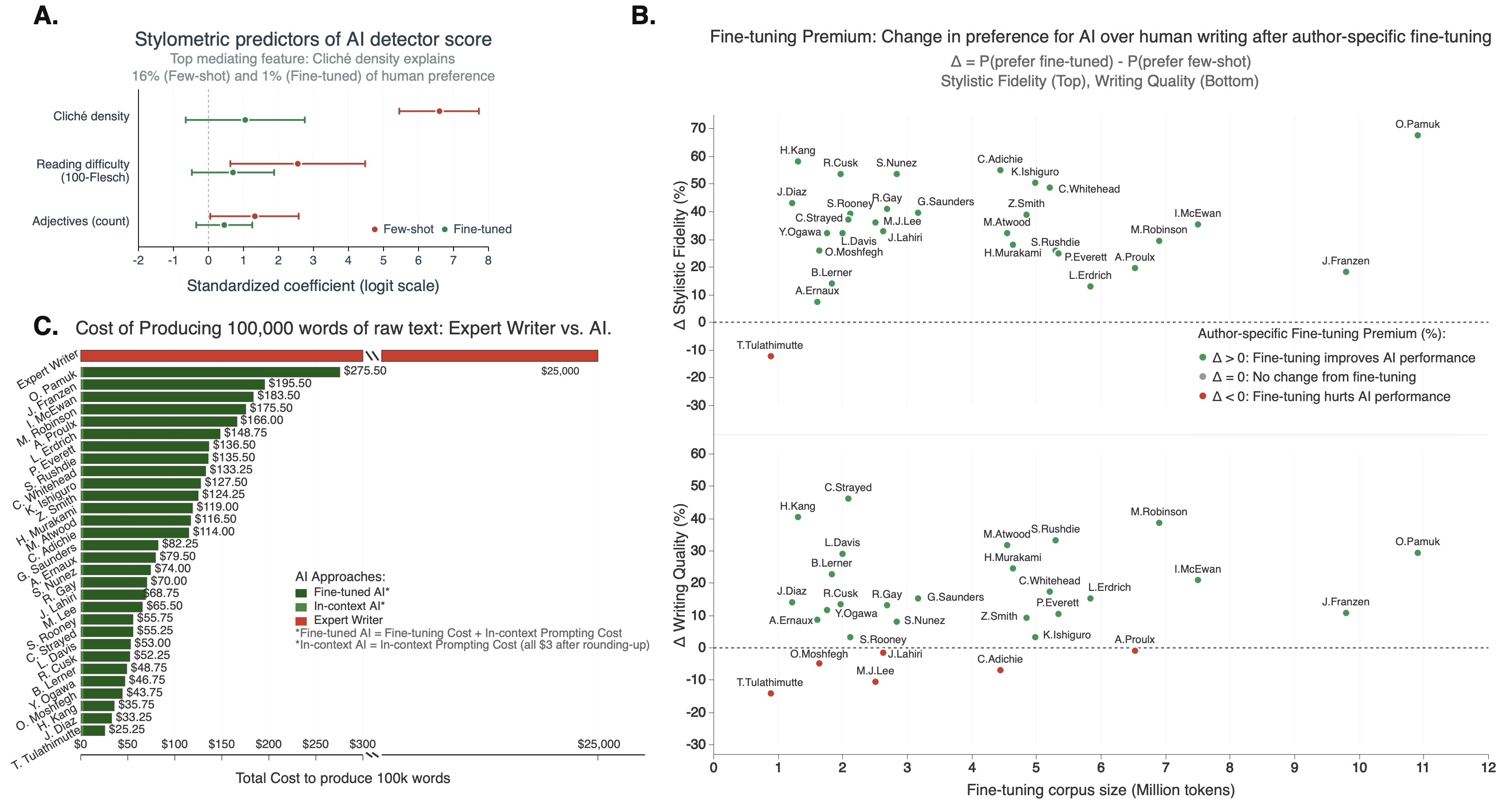}  % Note: no .svg extension
     \captionsetup{font={singlespacing,small},labelfont=bf}
     \caption{{\bf Fine-tuning substantially reduces stylometric signatures of AI text, improves stylistic fidelity and perceived writing quality over in-context prompting, and substantially cuts costs of producing first draft versus professional writers.} (A) Mediation analysis linking stylometric features $\rightarrow$ AI-detector score $\rightarrow$ human preference. Standardized logistic coefficients with 95\% CIs are shown for three features for in-context prompting (red) and fine-tuned models (green). Cliché density mediates 16.4\% of the detector effect on choice for in-context prompting but only 1.3\% for author fine-tuned models; all three features together mediate 25.4\% vs −3.2\% for in-context prompted and fine-tuned models respectively. (B) ``Fine-tuning premium," defined as $\Delta$ = P(prefer fine-tuned over human) − P(prefer in-context over human), as a function of fine-tuning corpus size. Top: stylistic fidelity; bottom: writing quality. Points are authors; colors denote improvement (green, $\Delta>0$), no change (gray), or degradation (red, $\Delta< 0$). Median $\Delta$: $+34.2$ (fidelity) and $+13.4$ percentage points (quality). (C) Cost to produce 100,000 words of raw text vs. publishable prose. MFA-trained writers in our study would earn \$25,000 for a 100k-word novel-length manuscript (red). By contrast, AI pipelines can generate 100k words of raw text for \$25–\$276 depending on fine-tuning corpus size (green bars = fine-tuning; hatched = in-context prompting, \$3). This figure reflects direct compute/API costs only, not the additional human steering, chunking, and editing required to turn raw AI text into a cohesive publishable work. Authors ordered by total AI cost.}
    \label{fig:fine-tuning-analysis}
\end{figure*}

\subsection*{Cost Analysis}
The weak dependence of performance on fine-tuning corpus size has direct economic implications. Total AI generations costs (fine-tuning plus inference) ranged from \$25 to \$276 per author (median = \$81), assuming API-based fine-tuning at \$25 per million tokens plus \$3 for generating 100,000 words of raw text (Fig. 4C). These costs represent approximately 0.3\% of what MFA-trained writers in our study would charge for a novel-length (100,000 words) manuscript. It should be noted that this comparison reflects raw generation costs before the human steering and editing required to transform AI outputs into publishable works. Despite this caveat, the minimal investment yielded outputs that achieved expert-level performance for most authors. Performance gains were uncorrelated with both corpus size and fine-tuning cost, indicating that computational scale did not drive improvements. The 99.7\% reduction in raw generation costs, coupled with superior quality ratings for the majority of authors, underscores the potential for substantial producer-surplus shifts and market displacement.

\section*{Discussion}
What are the implications of this research for the copyright infringement claims that authors have brought against AI companies alleging unauthorized use of their books in training datasets \cite{bartz2025anthropic,kadrey2025meta,mosaic2025}?  These cases raise the question whether copying millions of copyrighted books for AI training constitutes fair use when the resulting outputs do not themselves reproduce the copied works. The most significant consideration in evaluating this question is the fourth fair use factor: {\it ``the impact of the use upon the actual or potential market for the copyrighted work.''}\footnote{The fair use provision (section 107) of the US Copyright Act directs courts to consider four factors:
\begin{enumerate}
    \item the purpose and character of the use, including whether such use is of a commercial nature or is for nonprofit educational
    \item the nature of the copyrighted work;
    \item the amount and substantiality of the portion used in relation to the copyrighted work as a whole; and
    \item the effect of the use upon the potential market for or value of the copyrighted work. Courts understand this factor to concern the extent to which the works produced by the copying substitute for the author’s work.
\end{enumerate}
}

Courts understand this factor to concern the extent to which works produced through copying serve as market substitutes for the original author's work \cite{warhol2023,campbell1994}.

Our study has shown that AI-generated excerpts from in-context prompted models pre-trained on vast internet corpora (including millions of copyrighted works) were strongly disfavored by MFA-trained readers for both quality and stylistic fidelity, while college-educated general readers showed no preference on stylistic fidelity but already favored AI-generated text for writing quality. When models are fine-tuned on curated datasets consisting solely of individual authors' complete works, both groups decisively preferred the AI-generated excerpts over human-written examples. Notably, college-educated general readers, who are generally representative of the book-purchasing public, preferred fine-tuned AI by a substantially wider margin than MFA-trained readers (e.g., writing quality OR = 5.42 vs.\ 1.87). Moreover, the fine-tuned excerpts proved almost undetectable as AI-generated text (particularly when compared to outputs from in-context prompted models), while both MFA-trained and college-educated general readers preferred them over human writing in blind evaluations.

\iffalse
{\color{red}
While our study examines 450-word excerpts, this format isolates the passage-level voice, style, and quality that writers must sustain across any longer work. In its current form LLMs cannot write long form narratives that automatically balance coherence, quality, thematic consistency, and plot development across thousands of words. However with human steering and iterative prompting its totally possible to produce long form fiction/non-fiction. As a matter fact startups such as Sudowrite or Inkitt are specifically focusing on helping consumers write books using AI. In recent news \cite{vara_ai_romance_factory_2025} journalists have discussed how Genre Fiction publisher Inkitt  has influential backers and a vision for infinitely customizable AI driven content. In self publishing marketplace such as Kindle AI-generated books have already gained popularity \cite{jones_ai_knockoffs_2025,knibbs_scammy_ai_books_2024}. Given that fine-tuned models can produce voicey high quality literary text, it is totally possible for a human to steer these models to produce a book that wont be AI-detected and which readers might appreciate without disclosure.}
\fi 

At first glance, a legal analyst might conclude that our findings are irrelevant to fair use because the outputs from the blind pairwise evaluation do not reproduce the copied works. While they may exhibit comparable literary quality and high stylistic fidelity to the originals, copyright law does not protect authors' style\textemdash only their expression \cite{warhol2023,kadrey2025meta}. These outputs may offer credible substitutes for an author's works, but so do human-authored works inspired by prior works. However, there is an important difference between human and AI-generated emulations: humans read; AI systems copy. Unlike human memory, which is not a verbatim storage device, all AI-generation requires predicate copying despite rhetoric equating human learning with machine ``learning.'' \cite{guile2025machine, MitchellKrakauer2023,song2024inferring}

The U.S. Copyright Office has recognized that such predicate copying may cause cognizable market harm through competing works that the inputs enable, potentially flooding the market and causing ``market dilution'' \cite{copyright2024ai}. While acknowledging this ``market dilution'' approach to the fourth fair use factor as ``uncharted territory,'' the Office determined that both the statutory language and underlying concerns of the Copyright Act warrant this inquiry.
\textit{``\ldots The speed and scale at which AI systems generate content pose a serious risk of diluting markets for works of the same kind as in their training data. That means more competition for sales of an author’s works and more difficulty for audiences in finding them. If thousands of AI-generated romance novels are put on the market, fewer of the human-authored romance novels that the AI was trained on are likely to be sold\ldots Market harm can also stem from AI models’ generation of material stylistically similar to works in their training data\ldots''} The Office emphasized that the effect of the copying impacts extant works by putting them in competition with AI-generated outputs. Copyrighted works become fodder for new productions targeting the same markets \cite{walsh2021goodreads}. Crucially, the Office did not claim that competing AI outputs copy the inputted works; rather, it examined the economic consequences of the predicate copying that enables these competing outputs. This focus on inputs remains essential because, absent the initial copying, no infringement action exists against flooding markets with independently generated works—if a thousand humans write romance novels after reading Barbara Cartland's novels, they compete but do not infringe. The Copyright Office's expansive interpretation of ``potential market for or value of the copied work'' suggests that fair use might not excuse predicate copying even when it doesn't show up in the end product, if the copying's effect substitutes for source works.

In Kadrey v. Meta \cite{kadrey2025meta}, an infringement action brought by 13 book authors against the copying of their books into the database underpinning Meta's Llama LLM, Judge Chhabria granted summary judgment to Meta but accepted the theory of market dilution: ``\textit{[I]ndirect substitution is still substitution: If someone bought a romance novel written by an LLM instead of a romance novel written by a human author, the LLM-generated novel is substituting for the human-written one\ldots This case involves a technology that can generate literally millions of secondary works, with a miniscule fraction of the time and creativity used to create the original works it was trained on. No other use\ldots has anything near the potential to flood the market with competing works the way that LLM training does.''} Judge Chhabria effectively provided a road map for what authors would have to show to persuade a court that the AI inputs diluted their markets: First, is the AI system capable of generating such substitutional books now or in the near future? Second, what are the markets for the plaintiffs' books, and do the AI-generated books compete in those markets? Third, what impact does any such competition have on sales\textemdash does it displace those works entirely or merely chip away at demand, and are the effects likely to grow as AI-generated books proliferate and models improve? Fourth, how does the threat to the market for the plaintiffs’ books differ between a world in which LLM developers may copy those books and a world in which they may not?

Our study’s findings concerning reader preferences between human-authored and AI-generated works bear on all four considerations. They also demonstrate how LLMs have already gotten {\it ``better and better at writing human-like text.''} While Judge Chhabria speculated that the distinctiveness of an author’s style renders works by well-known authors less susceptible to substitution, he observed that market dilution would vary by author prominence. Established authors with dedicated readerships (like Agatha Christie) would likely face minimal substitution, while AI-generated books could crowd out lesser-known or emerging authors, potentially preventing {\it ``the next Agatha Christie from getting noticed or selling enough books to keep writing''}. Our work suggests otherwise. If readers in fact prefer AI-generated emulations of authors whose market value lies in their distinctive voices, then the prospect of competition, especially from outputs of fine-tuned datasets, appears to be considerable. The comparatively low production costs of AI-generated texts relative to paying human authors (as shown in Figure 4C) further enhances the likelihood that AI platforms will in fact dilute the market for human-authored work. 

These findings suggest that the creation of fine-tuned LLMs consisting of the collected copyrighted works (or a substantial number) of individual authors should not be fair use if the LLM is used to create outputs that emulate the author’s works \cite{ohm2024focusing}. As the Copyright Office observed, ``[f]ine-tuning\ldots usually narrows down the model’s capabilities and might be more aligned with the original purpose of the copyrighted material,'' \cite{copyright2024ai} and thus both less ``transformative,'' and more likely to substitute for it. By contrast, the LLMs employed for in-context prompting do not target particular authors, and therefore can be put to a great variety of uses that do not risk diluting those authors’ markets. Their claim to fair use seems accordingly stronger. But those models can generate author-emulations, and our study has shown that among college-educated general readers, who more closely represent the book-purchasing public, both in-context quality outputs and fine-tuned outputs show clear AI preference, indicating meaningful substitution potential even without author-specific training. Our findings have also been independently corroborated by a large-scale blind quiz conducted by The New York Times in which more than 86,000 readers compared AI and human-written passages and 54\% preferred AI-generated text \cite{roose2026nytquiz}, as well as independent reporting in The New Yorker where even accomplished novelists struggled to distinguish fine-tuned AI emulations from human writing \cite{vara2025newyorker}.

A reasonable solution might allow the inclusion of copyrighted works in the general-purpose dataset, but would require the model to implement guardrails that would disable it from generating non-parodic imitations of individual authors' oeuvres \cite{liu2024shield,chen2025parapo,jaech2024openai}. As examples of guardrails, AI developers have implemented ``refusal protocols'' blocking outputs when prompts request content ``in the style of'' specific authors. Further, current reinforcement learning techniques can be easily modified to steer models away from stylistic imitation. However, such guardrails remain imperfect: refusal protocols can be circumvented through prompt rephrasing, and reinforcement learning--based steering may not generalize across all forms of stylistic emulation. Another solution, particularly where the lower quality of in-context prompting reduces the prospect of market dilution, might be to condition a ruling of fair use on the prominent disclosure of the output's AI origin. This solution assumes that the public, informed that the output was not human-authored, will be less inclined to select the AI substitute; transparency should diminish the competition between human-authored and machine-generated offerings.

Whether disclosure would in fact diminish the competitive advantage of AI-generated works is a question our study cannot directly address, because all evaluations were conducted under blinded conditions. Whether the strong preferences observed here would persist when readers know the text is AI-generated remains an open empirical question requiring a separate, unblinded experimental design. Moreover, even if disclosure were to reduce the quality preference to some degree, relative cost may independently sustain consumer substitution. If AI-generated works enter the market at substantially lower price points than human-authored works---as our cost analysis suggests is feasible---consumers may select the AI version even knowing its provenance, because the quality-to-price ratio remains attractive. The interaction of disclosure, price, and reader preference in actual market conditions therefore warrants further investigation.

\section*{Methods}
\subsection*{Experimental Setup}
We recruited 28 expert writers affiliated with top MFA programs in the United States, including from programs such as Iowa Writers Workshop, Helen Zell Writers' Program at University of Michigan, MFA Program in Creative Writing at New York University, Columbia University School of Arts, and MFA in Creative Writing at BU, paying them \$75 for writing each excerpt. These MFA candidates, along with the LLMs described earlier, emulated the style and voice of 50 award-winning authors representing diverse cultural backgrounds and distinct literary voices (full list in Table~\ref{tab:authorlist} in SI). The writing prompt provided to MFA candidates contained three components: (i)~20 sample excerpts spanning an author's complete body of work, (ii)~textual descriptions of the author's distinctive style and voice, and (iii)~detailed content specifications about the original author-written excerpt to be emulated. The selection of the 50 authors and the development of all writing prompts were completed in collaboration with five English Literature PhD students who analyzed each author's literary voice and created the verbalized style descriptions in addition to carefully choosing representative excerpts written by the original author (prompt shown in Figure~\ref{fig:prompt} in SI). For the purpose of the study, MFA-trained expert writers had no time limitation to submit these excerpts and were encouraged to take as long as required to produce their best work.

\subsubsection*{In-context Prompting}
Larger context windows in LLMs allow for processing significantly more information at once, leading to improved accuracy, understanding, and complex reasoning capabilities. For our task this is crucial as it lets the model process the entire writing prompt including the 20 sample excerpts, style descriptions and content specifications during inference. Taking advantage of this capability in GPT-4o, Claude~3.5 Sonnet, and Gemini~1.5 Pro, we generated LLM emulations using the identical writing prompt provided to MFA candidates (as described above). In this condition, models received no additional training on the target authors' works; they relied solely on the information contained within the prompt. Thus for AI Condition~1 (in-context prompting), we obtained 150 $\langle$Human-AI$\rangle$ pairs: 150 human-written excerpts (3~MFA writers $\times$ 50~authors) paired with 150~AI-generated excerpts. Each human excerpt was paired with one AI excerpt for the same author, with the AI excerpts distributed equally across the three models (50~excerpts each from GPT-4o, Claude~3.5~Sonnet, and Gemini~1.5~Pro).

\subsubsection*{Fine-tuning}
For AI Condition~2 (fine-tuning), we selected 30 living authors from our pool. This decision was made specifically to examine the potential economic impact of generative AI on the livelihoods of living authors while also considering the substantial computational costs associated with fine-tuning models on individual authors. Fine-tuning on books presents several technical challenges. Each book typically contains 50,000 to 80,000 words, and using such long sequences as training examples wastes computational capacity because models still struggle with very long inputs and long-context adaptation remains non-trivial. To preserve general capabilities in supervised fine-tuning, it is generally advised to use diverse, shorter samples. Breaking books into context-independent excerpts increases batch diversity and the number of distinct gradient signals per token budget, while also reducing overfitting to a single narrative flow and still capturing the salient stylistic and local patterns characteristic of each author's voice.

We purchased ePub files of these 30 authors' complete works, converted them to plain text, and segmented them into 250--650 word context-independent excerpts. For each excerpt, we extracted content descriptions using GPT-4o with the following prompt: \textit{``Describe in detail what is happening in this excerpt. Mention the characters and whether the voice is in first or third person for majority of the excerpt. Maintain the order of sentences while describing.''} Since only GPT-4o (among our three models) supported API-based fine-tuning at the time of our study, we fine-tuned 30 author-specific GPT-4o models using input-output pairs structured as: ``Write a [[n]] word excerpt about the content below emulating the style and voice of [[authorname]]\textbackslash n\textbackslash n[[content]]: [[excerpt]]'' (see Figure~\ref{fig:finetuning} in SI for details). This technique is commonly referred to as \textbf{instruction back-translation}~\cite{li2023self}, where the model learns to generate text matching a target style given content specifications. Importantly, the original author excerpts on which the human-written emulations were based were excluded from the fine-tuning data to ensure fairness in comparison. This approach yielded 90 $\langle$Human-AI$\rangle$ pairs: each of the 30 fine-tuned GPT-4o models generated one excerpt per author, and each AI excerpt was paired with all three MFA-written excerpts for that author (30~authors $\times$ 3~human excerpts~=~90~pairs). During inference, we verified that no generated excerpt regurgitated verbatim expressions from the original training data. ROUGE-L scores~\cite{lin-2004-rouge} ranged from 0.16 to 0.23, indicating minimal lexical overlap between AI-generated and original author-written excerpts. Refer to Section~S1.3 for more details on the fine-tuning procedure and validation.

\subsection*{Human Participants for Writing and Evaluation Task}
These $\langle$Human-AI$\rangle$ pairs from both AI conditions (in-context prompting and fine-tuning) were evaluated by 28 experts (the same MFA candidates mentioned above) and 516 college-educated general readers recruited from Prolific,\footnote{\url{https://www.prolific.com/}} one of the leading crowdsourcing platforms for research participants. As mentioned earlier, MFA candidates served dual roles in our study: they created human-written excerpts for comparison with AI outputs, and they also evaluated all excerpts in blind pairwise comparisons (never evaluating their own work). We compensated MFA-trained writers \$75 for writing each excerpt and an additional \$75 for evaluating a batch of 10 writing quality comparisons or \$100 for evaluating a batch of 10 stylistic fidelity comparisons (the latter requiring comparison against original author excerpts).

Our MFA-trained writer sample represents the emerging literary elite whose professional judgment shapes contemporary publishing. Among our recruited MFA candidates, several have since been awarded Stanford Stegner Fellowships (one of the most competitive fiction writing fellowships in the United States), received Rhodes Scholarships, assumed editorial positions at prestigious literary magazines such as \textit{Joyland}, been awarded the Publishers Weekly Star Watch 2025 Honor and secured publishing contracts with major houses including W.W.~Norton and Harper Collins. These accomplishments underscore that our experts represent not merely MFA candidates, but established and emerging voices whose aesthetic judgments carry professional authority in the literary marketplace. 

All recruited MFAs were residing in the United States at the time of the study. This geographic restriction was necessary for administrative compliance, as U.S. tax regulations require declaration of compensation above certain thresholds, and the university's payment infrastructure could not accommodate international participants without Individual Taxpayer Identification Numbers (ITINs) or Social Security Numbers (SSNs). Despite this constraint, our expert sample demonstrated substantial demographic diversity: 65\% did not identify as cisgender men, and participants self-identified across multiple ethnic backgrounds, including South Asian, White, East Asian, and Black. This diversity helps ensure that our findings are not artifacts of a narrow demographic perspective.

For the 516 college-educated general readers recruited from Prolific, we restricted participation to native-born residents of English-speaking countries (USA and UK only) ensuring that all evaluators had native-level fluency in English. Additional eligibility criteria included maintaining a 100\% task acceptance rate on the platform (indicating high performance on previous tasks) and having a college degree (ensuring a baseline level of literacy). We compensated the college-educated general readers according to Prolific's wage standards with approximately \$15 per hour for their evaluation time. Unlike MFA-trained readers, college-educated general readers represented general educated audiences without specialized training in creative writing or literary analysis; hence their compensation was lower than MFAs.

\subsection*{Blind Evaluation}
Experts never evaluated their own excerpts. Each $\langle$Human-AI$\rangle$ pair was assessed by three MFA-trained readers and by 19 college-educated general readers in the in-context condition or 21 college-educated general readers in the fine-tuned condition. All reported analyses are conducted at the reader-judgment level, with inter-rater agreement summarized separately using Fleiss’ kappa within each reader group. In total, we obtained 6,600 pairwise evaluations (3,300 for quality, 3,300 for style) for AI Condition 1 (in-context prompting) and 4,320 evaluations (2,160 for quality, 2,160 for style) for AI Condition 2 (fine-tuning). The evaluation tasks differed by outcome: for quality evaluation, we showed only the $\langle$Human-AI$\rangle$ pair and asked readers to judge which excerpt was better written overall; for stylistic fidelity evaluation, we included the original author-written excerpt alongside the pair and asked readers to judge which excerpt better captured that author's distinctive style and voice (see Figures~\ref{fig:quality_eval_screen} and~\ref{fig:style_eval_screen} in SI for screenshots of both interfaces).

Beyond selecting a preferred excerpt, both groups of readers provided 2--3 sentence explanations grounded in textual evidence to justify their choices~\cite{mcdonnell2016relevant} (see Figures~\ref{fig:expertfewshotpref}--\ref{fig:styleexpertfinetunedpref} in SI for examples). We required these written rationales for two reasons. First, without observing why a choice was made, the latent factors driving that preference remain unclear---hidden user context, demographic or value variation, task ambiguity, or near-tie similarity can inflate effective label noise. Providing rationales or auxiliary side information has been empirically shown to improve label reliability, data efficiency, and debiasing~\cite{just2024data}. Second, for college-educated general readers specifically, written explanations serve as a quality control mechanism, helping us identify participants who may be selecting preferences randomly rather than engaging seriously with the task. Readers who provided generic or off-topic rationales, or whose explanations contradicted their stated preferences, were flagged for potential exclusion.

To ensure annotation quality, we implemented several attention checks. We recorded timestamps for each evaluation to identify and exclude participants who rushed through tasks (spending less than 30 seconds per comparison). Additionally, we screened all written rationales using Pangram,\footnote{\url{https://www.pangram.com/}} a state-of-the-art AI detection tool, and excluded any participants who used generative AI to compose their responses~\cite{veselovsky2023artificial}. This screening was necessary because AI-generated rationales would undermine the validity of human preference judgments. Our study was approved by the University of Michigan IRB (HUM00264127) and was preregistered at OSF.\footnote{\url{https://osf.io/zt4ad}} Informed consent was obtained from all participants. As noted earlier, we compensated MFAs \$75 per 10-excerpt batch for quality judgments and \$100 per 10-excerpt batch for stylistic fidelity evaluations (the higher rate reflecting the additional cognitive load of comparing against original author excerpts).

\subsection*{Empirical Analyses}
We tested three preregistered hypotheses using logistic regression models with CR2 cluster-robust standard errors clustered by reader. Clustering accounts for the fact that each reader evaluated multiple excerpt pairs, which induces within-reader correlation in preference judgments. The CR2 adjustment provides more accurate inference in settings with a modest number of clusters (28 MFA-trained expert readers, 516 college-educated general readers)~\cite{pustejovsky2018small}.

For H1 (can baseline LLMs match human expert performance?), we compared human writing to the average performance across GPT-4o, Claude~3.5 Sonnet, and Gemini~1.5 Pro in the in-context prompting condition. Because each human excerpt was paired with only one AI excerpt from a single model, we pooled AI excerpts across the three models and tested whether readers showed systematic preference for human versus AI writing on average. For H2 (can fine-tuned models match human expert performance?), we directly compared fine-tuned GPT-4o excerpts to human writing. Contrasts were computed separately for MFA-trained readers and college-educated general readers within each outcome (stylistic fidelity, writing quality), yielding four tests per hypothesis (2 reader groups $\times$ 2 outcomes). To control the family-wise error rate across these multiple comparisons, we applied Holm correction separately within each hypothesis-outcome combination. This procedure adjusts $p$-values to maintain valid inference when testing multiple contrasts that address the same research question.

For H3 (does AI detectability correlate with human preference, and does fine-tuning remove this correlation?), we modeled the relationship between Pangram AI-detection scores and reader preference. Specifically, we tested whether higher detection scores (indicating more AI-like text) predicted lower preference for AI excerpts in the in-context prompting condition, and whether fine-tuning attenuated this penalty via AI condition $\times$ detection score interactions. A significant negative coefficient on detection score in the in-context condition, coupled with a significant positive interaction term, would indicate that fine-tuning removes the detectability-based preference penalty that AI excerpts suffer under in-context prompting. All model specifications, including the full regression equations, covariate adjustments, and robustness checks, are described in detail in SI Sections~S4--S8.

\subsection*{Data availability}
Anonymized trial-level data supporting the findings of this study are available at GitHub at the link (\href{https://github.com/tuhinjubcse/Author-Style-Personalization/tree/main}{https://github.com/tuhinjubcse/Author-Style-Personalization/tree/main}). Additional details are provided in SI, Section S10.

\subsection*{Code availability}
All analysis and figure-generation code is available at GitHub (\href{https://github.com/tuhinjubcse/Author-Style-Personalization/tree/main}{https://github.com/tuhinjubcse/Author-Style-Personalization/tree/main}). Core analyses can be reproduced by running the numbered R scripts in sequence. Analyses were conducted in R 4.3.1 with key packages including \texttt{clubSandwich} and \texttt{emmeans}. Exact package versions and run instructions are provided in SI, Section S10.

\subsection*{Ethics statement}
All procedures involving human participants were approved by the University of Michigan Institutional Review Board (HUM00264127). Informed consent was obtained from all participants, who were compensated for their time. The study was preregistered at OSF (\href{https://osf.io/zt4ad}{https://osf.io/zt4ad}).

\section*{Acknowledgements}
We thank Jared Brent Harbor (Columbia Law School, J.D. Class of 2027; M.F.A. in Theatre Management and Producing, Columbia University School of the Arts) for research and editorial assistance.

\clearpage

\bibliography{science_template}

@Misc{methods,
  note = {Materials and methods are available as supplementary material},
}

@article{MitchellKrakauer2023,
  author    = {Melanie Mitchell and David C. Krakauer},
  title     = {The Debate over Understanding in {AI}'s Large Language Models},
  journal   = {Proceedings of the National Academy of Sciences},
  year      = {2023},
  volume    = {120},
  number    = {13},
  pages     = {e2215907120},
  doi       = {10.1073/pnas.2215907120},
  url       = {https://www.pnas.org/doi/10.1073/pnas.2215907120}
}

@article{russell2025people,
  title={People who frequently use ChatGPT for writing tasks are accurate and robust detectors of AI-generated text},
  author={Russell, Jenna and Karpinska, Marzena and Iyyer, Mohit},
  journal={arXiv preprint arXiv:2501.15654},
  year={2025}
}

@article{liu2024shield,
  title={Shield: Evaluation and defense strategies for copyright compliance in llm text generation},
  author={Liu, Xiaoze and Sun, Ting and Xu, Tianyang and Wu, Feijie and Wang, Cunxiang and Wang, Xiaoqian and Gao, Jing},
  journal={arXiv preprint arXiv:2406.12975},
  year={2024}
}

@article{song2024inferring,
  title={Inferring neural activity before plasticity as a foundation for learning beyond backpropagation},
  author={Song, Yuhang and Millidge, Beren and Salvatori, Tommaso and Lukasiewicz, Thomas and Xu, Zhenghua and Bogacz, Rafal},
  journal={Nature neuroscience},
  volume={27},
  number={2},
  pages={348--358},
  year={2024},
  publisher={Nature Publishing Group US New York}
}

@article{guile2025machine,
  title={Machine learning and human learning: a socio-cultural and-material perspective on their relationship and the implications for researching working and learning},
  author={Guile, David and Popov, Jelena},
  journal={AI \& SOCIETY},
  volume={40},
  number={2},
  pages={325--338},
  year={2025},
  publisher={Springer}
}

@article{jaech2024openai,
  title={Openai o1 system card},
  author={Jaech, Aaron and Kalai, Adam and Lerer, Adam and Richardson, Adam and El-Kishky, Ahmed and Low, Aiden and Helyar, Alec and Madry, Aleksander and Beutel, Alex and Carney, Alex and others},
  journal={arXiv preprint arXiv:2412.16720},
  year={2024}
}

@article{chen2025parapo,
  title={ParaPO: Aligning Language Models to Reduce Verbatim Reproduction of Pre-training Data},
  author={Chen, Tong and Brahman, Faeze and Liu, Jiacheng and Mireshghallah, Niloofar and Shi, Weijia and Koh, Pang Wei and Zettlemoyer, Luke and Hajishirzi, Hannaneh},
  journal={arXiv preprint arXiv:2504.14452},
  year={2025}
}

@article{roose2026nytquiz,
  author    = {Roose, Kevin and Thompson, Stuart A.},
  title     = {Who's a Better Writer: {A.I.} or Humans? {Take} Our Quiz},
  journal   = {The New York Times},
  year      = {2026},
  month     = mar,
  day       = {9},
  url       = {https://www.nytimes.com/interactive/2026/03/09/business/ai-writing-quiz.html},
  note      = {Interactive feature. Accessed: March 12, 2026}
}

@article{vara2025newyorker,
  author    = {Vara, Vauhini},
  title     = {What If Readers Like {A.I.}-Generated Fiction?},
  journal   = {The New Yorker},
  year      = {2025},
  month     = dec,
  url       = {https://www.newyorker.com/culture/the-weekend-essay/what-if-readers-like-ai-generated-fiction},
  note      = {The Weekend Essay. Accessed: March 12, 2026}
}

@article{alter2026romance,
  title={The New Fabio Is Claude: How {A.I.} Is Reshaping Romance Novels},
  author={Alexandra Alter},
  journal={The New York Times},
  year={2026},
  month={February},
  day={8},
  url={https://www.nytimes.com/2026/02/08/business/ai-claude-romance-books.html},
  note={Accessed: 2026-03-09}
}

@misc{PW2025,
  author = {{Publishers Weekly}},
  title = {Book Publishing Sales Rose 6.5\% in 2024, Per Preliminary Data},
  howpublished = {\textit{Publishers Weekly}},
  year = {2025},
  url = {https://www.publishersweekly.com/pw/by-topic/industry-news/financial-reporting/article/97224-book-publishing-sales-rose-6-5-in-2024-per-preliminary-data.html},
  note = {Accessed: September 7, 2025}
}

@article{pustejovsky2018small,
  title = {Small-sample methods for cluster-robust variance estimation and hypothesis testing in fixed effects models},
  author = {Pustejovsky, James E. and Tipton, Elizabeth},
  journal = {Journal of Business \& Economic Statistics},
  volume = {36},
  number = {4},
  pages = {672--683},
  year = {2018},
  publisher = {Taylor \& Francis}
}

@misc{codeforces2025,
  title = {{Codeforces: Programming competitions and contests, programming community}},
  howpublished = {\url{https://codeforces.com/}},
  year = {2025},
  note = {Accessed: September 7, 2025}
}

@misc{pw_literary_prizes,
  title = {{Literary Prizes Under Scrutiny}},
  howpublished = {\textit{Poets \& Writers}, \url{https://www.pw.org/content/literary_prizes_under_scrutiny}},
  year = {2025},
  note = {Accessed: September 7, 2025}
}

@misc{bls2023,
  title = {{Occupational Employment and Wage Statistics: Writers and Authors}},
  author = {{U.S. Bureau of Labor Statistics}},
  howpublished = {\url{https://www.bls.gov/oes/2023/may/oes273041.html}},
  year = {2023},
  month = {May},
  note = {Accessed: September 7, 2025}
}

@misc{lithub2024,
  title = {{Against AI: An Open Letter from Writers to Publishers}},
  howpublished = {\textit{Literary Hub}, \url{https://lithub.com/against-ai-an-open-letter-from-writers-to-publishers/}},
  year = {2024},
  note = {Accessed: September 7, 2025}
}

@case{bartz2025anthropic,
  title        = {Bartz et al. v. Anthropic PBC},
  court        = {United States District Court, Northern District of California},
  year         = {2025},
  number       = {No. 3:24-cv-05417},
  date         = {2025-09-05},
  url          = {https://www.courtlistener.com/docket/69058235/bartz-v-anthropic-pbc/},
  note         = {Settlement reached after court granted partial summary judgment on fair use for training but denied on piracy claims},
}

@case{kadrey2025meta,
  title        = {Kadrey et al. v. Meta Platforms, Inc.},
  court        = {United States District Court, Northern District of California},
  year         = {2025},
  number       = {No. 3:23-cv-03417-VC},
  date         = {2025-06-25},
  url          = {https://law.justia.com/cases/federal/district-courts/california/candce/3:2023cv03417/415175/598/},
  note         = {Order denying plaintiffs' motion for partial summary judgment and granting Meta's cross-motion on fair use grounds},
}

@techreport{copyright2024ai,
  title        = {Copyright and Artificial Intelligence Part 3: Generative AI Training Report},
  author       = {{U.S. Copyright Office}},
  institution  = {U.S. Copyright Office},
  year         = {2024},
  month        = {December},
  url          = {https://www.copyright.gov/ai/Copyright-and-Artificial-Intelligence-Part-3-Generative-AI-Training-Report-Pre-Publication-Version.pdf},
  note         = {Pre-publication version analyzing copyright implications of AI training},
}

@case{warhol2023,
  title        = {Andy Warhol Foundation for the Visual Arts, Inc. v. Goldsmith},
  court        = {Supreme Court of the United States},
  year         = {2023},
  volume       = {598},
  reporter     = {U.S.},
  pages        = {508, 533--534},
  date         = {2023-05-18},
  number       = {No. 21-869},
  url          = {https://www.supremecourt.gov/opinions/22pdf/21-869_87ad.pdf},
  note         = {Holding that the first fair use factor focuses on whether the use shares the same purpose or supersedes the original work},
}

@case{campbell1994,
  title        = {Campbell v. Acuff-Rose Music, Inc.},
  court        = {Supreme Court of the United States},
  year         = {1994},
  volume       = {510},
  reporter     = {U.S.},
  pages        = {569, 590},
  date         = {1994-03-07},
  number       = {No. 92-1292},
  url          = {https://www.supremecourt.gov/opinions/boundvolumes/510bv.pdf},
  note         = {Establishing that market substitution is central to fair use analysis under the fourth factor},
}

@case{mosaic2025,
  title        = {In re Mosaic LLM Litigation},
  court        = {United States District Court, Northern District of California},
  year         = {2025},
  number       = {No. 3:24-cv-01451-CRB},
  date         = {2025-06-25},
  url          = {https://www.courtlistener.com/docket/68325564/onan-v-databricks-inc/},
  note         = {Consolidated cases against Databricks and MosaicML for alleged use of pirated books in training LLMs},
}

@article{samuelson2023generative,
  title={Generative AI meets copyright},
  author={Samuelson, Pamela},
  journal={Science},
  volume={381},
  number={6654},
  pages={158--161},
  year={2023},
  publisher={American Association for the Advancement of Science}
}

@inproceedings{li-etal-2024-disclosure,
    title = "How Does the Disclosure of {AI} Assistance Affect the Perceptions of Writing?",
    author = "Li, Zhuoyan  and
      Liang, Chen  and
      Peng, Jing  and
      Yin, Ming",
    editor = "Al-Onaizan, Yaser  and
      Bansal, Mohit  and
      Chen, Yun-Nung",
    booktitle = "Proceedings of the 2024 Conference on Empirical Methods in Natural Language Processing",
    month = nov,
    year = "2024",
    address = "Miami, Florida, USA",
    publisher = "Association for Computational Linguistics",
    url = "https://aclanthology.org/2024.emnlp-main.279/",
    doi = "10.18653/v1/2024.emnlp-main.279",
    pages = "4849--4868"
}

@article{just2024data,
  title={Data-centric human preference optimization with rationales},
  author={Just, Hoang Anh and Jin, Ming and Sahu, Anit and Phan, Huy and Jia, Ruoxi},
  journal={arXiv preprint arXiv:2407.14477},
  year={2024}
}

@article{li2023self,
  title={Self-alignment with instruction backtranslation},
  author={Li, Xian and Yu, Ping and Zhou, Chunting and Schick, Timo and Levy, Omer and Zettlemoyer, Luke and Weston, Jason and Lewis, Mike},
  journal={arXiv preprint arXiv:2308.06259},
  year={2023}
}

@misc{aap_industry_statistics_2025,
  author       = {{Association of American Publishers}},
  title        = {Industry Statistics},
  howpublished = {\url{https://publishers.org/data-and-statistics/industry-statistics/}},
  year         = {2025},
  note         = {Accessed: July 2, 2025}
}

@article{reisner_books3_2023,
  author       = {Reisner, Alex},
  title        = {What I Found in a Database Meta Uses to Train Generative AI},
  journal      = {The Atlantic},
  date         = {2023-09-25},
  url          = {https://www.theatlantic.com/technology/archive/2023/09/books3-ai-training-meta-copyright-infringement-lawsuit/675411/},
  note         = {Accessed: July 2, 2025}
}

@techreport{usdcc_ndca_bartz_anthropic_2025,
  author       = {{United States District Court for the Northern District of California}},
  title        = {Order on Fair Use},
  institution  = {United States District Court, Northern District of California},
  type         = {Court Opinion},
  number       = {No. C 24-05417 WHA, Doc. 231},
  date         = {2025-06-23},
  url          = {https://storage.courtlistener.com/recap/gov.uscourts.cand.434709/gov.uscourts.cand.434709.231.0.pdf},
  note         = {Accessed: July 2, 2025}
}

@article{horton2023bias,
  title={Bias against AI art can enhance perceptions of human creativity},
  author={Horton Jr, C Blaine and White, Michael W and Iyengar, Sheena S},
  journal={Scientific reports},
  volume={13},
  number={1},
  pages={19001},
  year={2023},
  publisher={Nature Publishing Group UK London}
}

@inproceedings{sarkar2025ai,
  title={AI Could Have Written This: Birth of a Classist Slur in Knowledge Work},
  author={Sarkar, Advait},
  booktitle={Proceedings of the Extended Abstracts of the CHI Conference on Human Factors in Computing Systems},
  pages={1--12},
  year={2025}
}

@misc{openai_introducing_healthbench_2025,
  author       = {{OpenAI}},
  title        = {Introducing HealthBench},
  howpublished = {\url{https://openai.com/index/healthbench/}},
  month        = {May},
  year         = {2025},
  note         = {Accessed: July 2, 2025}
}

@article{kolata_chatbots_2024,
  author       = {Kolata, Gina},
  title        = {A.I. Chatbots Defeated Doctors at Diagnosing Illness},
  journal      = {The New York Times},
  date         = {2024-11-17},
  url          = {https://www.nytimes.com/2024/11/17/health/chatgpt-ai-doctors-diagnosis.html},
  note         = {Accessed: July 2, 2025}
}

@article{trinh2024solving,
  title={Solving olympiad geometry without human demonstrations},
  author={Trinh, Trieu H and Wu, Yuhuai and Le, Quoc V and He, He and Luong, Thang},
  journal={Nature},
  volume={625},
  number={7995},
  pages={476--482},
  year={2024},
  publisher={Nature Publishing Group UK London}
}

@article{noy2023experimental,
  title={Experimental evidence on the productivity effects of generative artificial intelligence},
  author={Noy, Shakked and Zhang, Whitney},
  journal={Science},
  volume={381},
  number={6654},
  pages={187--192},
  year={2023},
  publisher={American Association for the Advancement of Science}
}

@inproceedings{gero2025creative,
  title={Creative Writers' Attitudes on Writing as Training Data for Large Language Models},
  author={Gero, Katy Ilonka and Desai, Meera and Schnitzler, Carly and Eom, Nayun and Cushman, Jack and Glassman, Elena L},
  booktitle={Proceedings of the 2025 CHI Conference on Human Factors in Computing Systems},
  pages={1--16},
  year={2025}
}

@article{Naddaf2025NatureLLMDetection,
  author  = {Miryam Naddaf},
  title   = {AI tool detects LLM-generated text in research papers and peer reviews},
  journal = {Nature},
  year    = {2025},
  month   = sep,
  note    = {News},
  doi     = {10.1038/d41586-025-02936-6},
  url     = {https://www.nature.com/articles/d41586-025-02936-6},
  urldate = {2025-09-14}
}

@article{doshi2024generative,
  title={Generative AI enhances individual creativity but reduces the collective diversity of novel content},
  author={Doshi, Anil R and Hauser, Oliver P},
  journal={Science Advances},
  volume={10},
  number={28},
  pages={eadn5290},
  year={2024},
  publisher={American Association for the Advancement of Science}
}

@misc{jabarian_imas_2025_artificial,
  author    = {Jabarian, Brian and Imas, Alex},
  title     = {Artificial Writing and Automated Detection},
  year      = {2025},
  month     = aug,
  note      = {SSRN working paper, Abstract ID 5407424},
  url       = {https://ssrn.com/abstract=5407424},
  urldate   = {2025-08- 31}
}

@article{handa2025economic,
  title={Which economic tasks are performed with ai? evidence from millions of claude conversations},
  author={Handa, Kunal and Tamkin, Alex and McCain, Miles and Huang, Saffron and Durmus, Esin and Heck, Sarah and Mueller, Jared and Hong, Jerry and Ritchie, Stuart and Belonax, Tim and others},
  journal={arXiv preprint arXiv:2503.04761},
  year={2025}
}

@article{chiang_why_2024,
  author       = {Chiang, Ted},
  title        = {Why A.{I}. Isn’t Going to Make Art},
  journal      = {The New Yorker},
  year         = {2024},
  month        = {August},
  day          = {31},
  note         = {The Weekend Essay},
  url          = {https://www.newyorker.com/culture/the-weekend-essay/why-ai-isnt-going-to-make-art},
  accessdate   = {2025-06-21},
}

@article{ohm2024focusing,
  title={Focusing on fine-tuning: Understanding the four pathways for shaping generative AI},
  author={Ohm, Paul},
  journal={Science and Technology Law Review},
  volume={25},
  number={2},
  year={2024}
}

@article{Tangermann2025,
  author       = {Tangermann, Victor},
  title        = {Readers Annoyed When Fantasy Novel Accidentally Leaves AI Prompt in Published Version, Showing Request to Copy Another Writer's Style},
  journal      = {Futurism},
  year         = {2025},
  month        = {May},
  day          = {23},
  url          = {https://futurism.com/fantasy-novel-ai-prompt-copy-style},
  note         = {Accessed: 2025-06-21},
}

@article{walsh2021goodreads,
  title={The goodreads “classics”: a computational study of readers, Amazon, and crowdsourced amateur criticism},
  author={Walsh, Melanie and Antoniak, Maria},
  journal={Journal of Cultural Analytics},
  volume={6},
  number={2},
  pages={243--287},
  year={2021},
  publisher={Department of Languages, Literatures, and Cultures}
}

@article{walsh2024does,
  title={Does ChatGPT Have a Poetic Style?},
  author={Walsh, Melanie and Preus, Anna and Gronski, Elizabeth},
  journal={arXiv preprint arXiv:2410.15299},
  year={2024}
}

@online{jones_ai_knockoffs_2025,
  author       = {CT Jones},
  title        = {Amazon Is the World’s Biggest Online Book Marketplace. It’s Filled With AI Knockoffs},
  subtitle     = {Authors say Amazon’s knockoff book problem is leaving them frustrated—and making the internet worse in the process},
  date         = {2025-10-27},
  url          = {https://www.rollingstone.com/culture/culture-features/amazon-ai-book-knockoffs-1235450690/},
  urldate      = {2025-10-29},
  organization = {Rolling Stone},
  publisher    = {Penske Media Corporation},
  langid       = {english},
  note         = {Culture Feature}
}

@online{knibbs_scammy_ai_books_2024,
  author       = {Kate Knibbs},
  title        = {Scammy AI-Generated Book Rewrites Are Flooding Amazon},
  date         = {2024-01-10},
  url          = {https://www.wired.com/story/scammy-ai-generated-books-flooding-amazon/},
  urldate      = {2025-10-29},
  organization = {WIRED},
  publisher    = {Condé Nast},
  langid       = {english},
  note         = {Business}
}

@online{vara_ai_romance_factory_2025,
  author       = {Vauhini Vara},
  title        = {The A.I. Romance Factory},
  subtitle     = {Genre fiction publisher Inkitt has influential backers and a vision for infinitely customizable A.I.-driven content},
  date         = {2025-04-07},
  url          = {https://www.bloomberg.com/features/2025-ai-romance-factory/},
  urldate      = {2025-10-29},
  organization = {Bloomberg Businessweek},
  publisher    = {Bloomberg L.P.},
  langid       = {english},
  note         = {Feature}
}

@article{Vara2023,
  author       = {Vara, Vauhini},
  title        = {Confessions of a Viral AI Writer},
  journal      = {WIRED},
  year         = {2023},
  month        = {September},
  day          = {21},
  url          = {https://www.wired.com/story/confessions-viral-ai-writer-chatgpt/},
  note         = {Accessed: 2025-06-21},
}

@article{delaney2007where,
  author  = {Delaney, Edward J.},
  title   = {Where Great Writers Are Made: Assessing America’s Top Graduate Writing Programs},
  journal = {The Atlantic},
  volume  = {300},
  number  = {1},
  month   = aug,
  year    = {2007},
  url     = {https://www.theatlantic.com/magazine/archive/2007/08/where-great-writers-are-made/306032/},
  note    = {Accessed: 07 July 2025}
}

@inproceedings{mcdonnell2016relevant,
  title={Why is that relevant? collecting annotator rationales for relevance judgments},
  author={McDonnell, Tyler and Lease, Matthew and Kutlu, Mucahid and Elsayed, Tamer},
  booktitle={Proceedings of the AAAI Conference on Human Computation and Crowdsourcing},
  volume={4},
  pages={139--148},
  year={2016}
}

@article{laquintano2024ai,
  title={AI and the Everyday Writer},
  author={Laquintano, Timothy and Vee, Annette},
  journal={PMLA},
  volume={139},
  number={3},
  pages={527--532},
  year={2024},
  publisher={Cambridge University Press}
}

@article{tomlinson2025working,
  title={Working with AI: Measuring the Occupational Implications of Generative AI},
  author={Tomlinson, Kiran and Jaffe, Sonia and Wang, Will and Counts, Scott and Suri, Siddharth},
  journal={arXiv preprint arXiv:2507.07935},
  year={2025}
}

@article{veselovsky2023artificial,
  title={Artificial artificial artificial intelligence: Crowd workers widely use large language models for text production tasks},
  author={Veselovsky, Veniamin and Ribeiro, Manoel Horta and West, Robert},
  journal={arXiv preprint arXiv:2306.07899},
  year={2023}
}

@techreport{NBERw34255,
 title = "How People Use ChatGPT",
 author = "Chatterji, Aaron and Cunningham, Thomas and Deming, David J and Hitzig, Zoe and Ong, Christopher and Shan, Carl Yan and Wadman, Kevin",
 institution = "National Bureau of Economic Research",
 type = "Working Paper",
 series = "Working Paper Series",
 number = "34255",
 year = "2025",
 month = "September",
 doi = {10.3386/w34255},
 URL = "http://www.nber.org/papers/w34255"
}

@inproceedings{lin-2004-rouge,
    title = "{ROUGE}: A Package for Automatic Evaluation of Summaries",
    author = "Lin, Chin-Yew",
    booktitle = "Text Summarization Branches Out",
    month = jul,
    year = "2004",
    address = "Barcelona, Spain",
    publisher = "Association for Computational Linguistics",
    url = "https://aclanthology.org/W04-1013/",
    pages = "74--81"
}

@inproceedings{chakrabarty2024art,
  title={Art or artifice? large language models and the false promise of creativity},
  author={Chakrabarty, Tuhin and Laban, Philippe and Agarwal, Divyansh and Muresan, Smaranda and Wu, Chien-Sheng},
  booktitle={Proceedings of the 2024 CHI Conference on Human Factors in Computing Systems},
  pages={1--34},
  year={2024}
}

@inproceedings{chakrabarty2025can,
  title={Can ai writing be salvaged? mitigating idiosyncrasies and improving human-ai alignment in the writing process through edits},
  author={Chakrabarty, Tuhin and Laban, Philippe and Wu, Chien-Sheng},
  booktitle={Proceedings of the 2025 CHI Conference on Human Factors in Computing Systems},
  pages={1--33},
  year={2025}
}
\bibliographystyle{naturemag}

% \section*{Author contributions}
% \textbf{TC, PSD}: Conceptualization; Methodology; Data Analysis; Writing (original draft); Writing (review and editing). \\
% \textbf{JCG}: Writing (original draft); Writing (review and editing).

% \section*{Competing interests}
% The authors declare no competing interests.

% \section*{Materials \& correspondence}
% Correspondence should be addressed to Tuhin Chakrabarty (\href{mailto:tchakrabarty@cs.stonybrook.edu}{tchakrabarty@cs.stonybrook.edu}), Jane C. Ginsburg (\href{mailto:ginsburg@law.columbia.edu}{ginsburg@law.columbia.edu}), or Paramveer Dhillon (\href{mailto:dhillonp@umich.edu}{dhillonp@umich.edu}).

\newpage

\begin{center}
\section*{Supplementary Information for\\ \scititle}

\author{
	Tuhin Chakrabarty$^{1}$, 
	Jane C. Ginsburg$^{2}$,
	Paramveer Dhillon$^{3,4}$\and\\
	\small$^{1}$Department of Computer Science, Stony Brook University.\\
        \small$^{2}$Columbia Law School.\\
	\small$^{3}$School of Information Science, University of Michigan.\\
    \small$^{4}$MIT Initiative on the Digital Economy.\and\\
	\small Corresponding authors: tchakrabarty@cs.stonybrook.edu, ginsburg@law.columbia.edu,\\ \small dhillonp@umich.edu
}

\end{center}

\textbf{\Large Materials and Methods}

\section*{S1: Details about Writing Task}

\subsection*{S1.1 Author List\label{authorlist}}
Table \ref{tab:authorlist} shows the list of 50 authors that were chosen by English Literature Ph.D. students. These chosen author list consists of canon-plus-global giants (Ernest Hemingway, Virginia Woolf, William Faulkner, Gabriel Garcia Márquez, Stephen King, Haruki Murakami, Kazuo Ishiguro), a strong set of contemporary prominent literary voices (Margaret Atwood, Ian McEwan, Jonathan Franzen, Colson Whitehead, George Saunders,  Louise Erdrich, Octavia Butler, Salman Rushdie, Maya Angelou, Percival Everett), and a group of critically acclaimed / emerging authors (Ottessa Mosfegh, Tony Tulathimutte, Roxane Gay). Additionally our author list is culturally diverse where several of the authors write primarily in a non-English language (Han Kang, Yoko Ogawa, Annie Ernaux). Roughly two‑thirds (34) have secured at least one major international or national prize (e.g., Nobel, Booker, Pulitzer, National Book Award, MacArthur Fellowship, Women’s Prize, International Booker, Hugo/Nebula). 8 of the authors are Nobel Prize winners in Literature and 8 are Pulitzer Prize winners.

\begin{table*}[!ht]
\small
\centering
\caption{List of Authors}
\begin{tabular}{r l  r l  r l}
\toprule
\# & Author & \# & Author & \# & Author \\
\midrule
 1 & Alice Munro            & 19 & J.D. Salinger        & 37 & Philip Roth \\
 2 & Annie Ernaux (\checkmark)          & 20 & Jhumpa Lahiri (\checkmark)       & 38 & Rachel Cusk \\
 3 & Annie Proulx  (\checkmark)         & 21 & Joan Didion          & 39 & Roxane Gay (\checkmark)\\
 4 & Ben Lerner (\checkmark)            & 22 & Jonathan Franzen (\checkmark)    & 40 & Sally Rooney (\checkmark) \\
 5 & Charles Bukowski       & 23 & Junot Díaz (\checkmark)          & 41 & Salman Rushdie (\checkmark)\\
 6 & Cheryl Strayed (\checkmark)        & 24 & Kazuo Ishiguro (\checkmark)       & 42 & Shirley Jackson \\
 7 & Chimamanda Ngozi Adichie (\checkmark) &25 & Louise Erdrich (\checkmark)    & 43 & Sigrid Nunez (\checkmark) \\
 8 & Colson Whitehead  (\checkmark)     & 26 & Lydia Davis         (\checkmark) & 44 & Stephen King \\
 9 & Cormac McCarthy        & 27 & Margaret Atwood (\checkmark)     & 45 & Tony Tulathimutte (\checkmark)\\
10 & David Foster Wallace   & 28 & Marilynne Robinson (\checkmark)   & 46 & V. S. Naipaul \\
11 & Ernest Hemingway       & 29 & Maya Angelou         & 47 & Virginia Woolf \\
12 & Flannery O'Connor      & 30 & Milan Kundera        & 48 & William Faulkner \\
13 & Gabriel García Márquez & 31 & Min Jin Lee  (\checkmark)        & 49 & Yoko Ogawa (\checkmark)\\
14 & George Saunders (\checkmark)       & 32 & Nora Ephron          & 50 & Zadie Smith (\checkmark)\\
15 & Han Kang (\checkmark)              & 33 & Octavia Butler       &    & \\
16 & Haruki Murakami (\checkmark)       & 34 & Orhan Pamuk (\checkmark)         &    & \\
17 & Hunter S. Thompson     & 35 & Ottessa Moshfegh (\checkmark)    &    & \\
18 & Ian McEwan  (\checkmark)           & 36 & Percival Everett (\checkmark)     &    & \\
\bottomrule
\end{tabular}
\vspace{1ex}
\caption{\label{tab:authorlist}Author list for our pool of 50 authors. (\checkmark) denotes authors who were used in fine-tuning experiment}
\end{table*}

\subsection*{S1.2 Writing Prompt\label{writingprompt}}
\vspace{-3ex}
\begin{figure*}[!htbp]
    \includegraphics[width=1.0\linewidth]{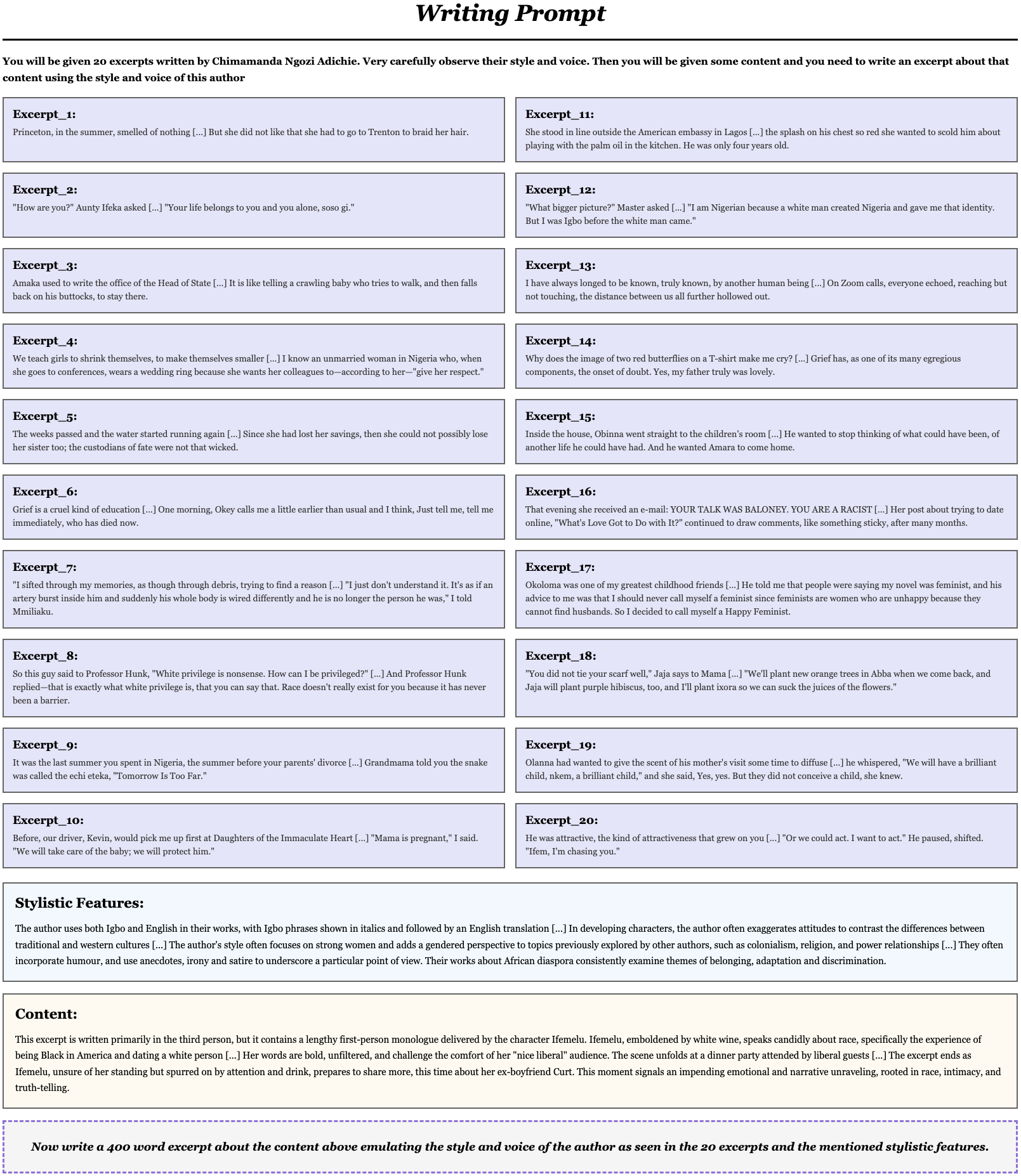}
    \vspace{-1ex}
    \caption{In-Context Writing Prompt used in AI Condition 1.}
    \label{fig:prompt}
\end{figure*}

Our writing prompt can be seen in Figure \ref{fig:prompt}). The same prompt was provided to both experts( MFA candidates) and LLMs. 

\subsection*{S1.3 Fine-tuning details\label{finetuning_details}}

\begin{figure*}[!htbp]
    \includegraphics[width=1.05\linewidth]{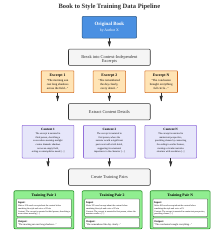}
    \caption{The pipeline used to fine-tune ChatGPT on an author's entire oeuvre}
    \label{fig:finetuning}
\end{figure*}

For fine-tuning we bought digital ePub versions of these authors' books and transformed them into plain text files. If the epub file was basically a wrapper around scanned page images, then we ignored that epub. The number of books written by each author vary a lot. For instance \textit{Tony Tulathimutte} has written only two books \textit{Private Citizens} and \textit{Rejection} so we could only fine-tune GPT-4o on two of them. While for \textit{Haruki Murakami} we could fine-tune on 22 books \textit{A Wild Sheep Chase}, \textit{After Dark}, \textit{After the Quake}, \textit{Blind Willow, Sleeping Woman}, \textit{Colorless Tsukuru Tazaki and his Years of Pilgrimage}, \textit{Dance Dance Dance}, \textit{First Person Singular}, \textit{Hard-boiled Wonderland and the End of the World}, \textit{Hear the Wind Sing}, \textit{Kafka on the Shore}, \textit{Killing Commendatore}, \textit{Men Without Women}, \textit{Norwegian Wood}, \textit{Novelist as a Vocation}, \textit{One and Two}, \textit{Pinball, 1973}, \textit{South of the Border, West of the Sun}, \textit{Sputnik Sweetheart}, \textit{The City and Its Uncertain Walls}, \textit{The Elephant Vanishes}, \textit{The Wind-Up Bird Chronicle}, \textit{What I Talk About When I Talk About Running}, \textit{Wind/Pinball: Two Novels}

Our entire fine-tuning pipeline can be seen in Figure \ref{fig:finetuning}. We first convert the epub files to txt using \url{https://github.com/kevinboone/epub2txt2}. We then segment the entire book into context independent excerpts. At a first pass we naively split the entire book text by existing double-newlines and rejoin them to enforce excerpt size bounds(250-650 words). For the rare cases where the naive splitting would lead to excerpts longer than 650 words we used GPT-4o to segment them again using the prompt \textit{Segment it into excerpts of minimum length 300-350 words such that each excerpt is grammatical from the start and doesn't feel abruptly cut off. There should be zero deletion and break into excerpts at grammatically natural places. Maintain the original word count. Avoid breaking into too many small excerpts. Start directly. Don't say Here's or Here is ....}

\begin{figure*}[!htbp]
    \includegraphics[width=1.05\linewidth]{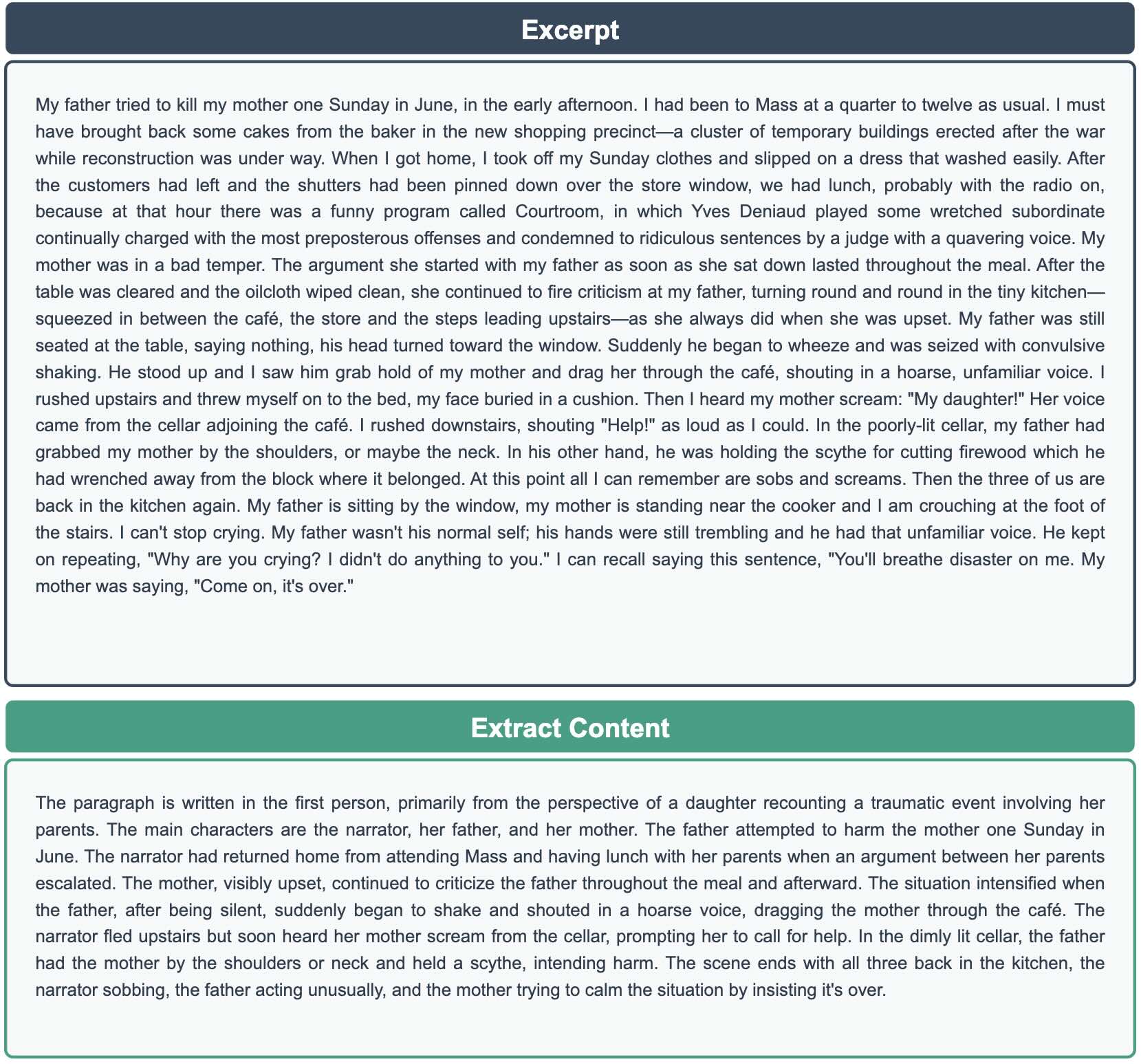}
    \caption{Original author written excerpt and extracted content using an LLM}
    \label{fig:extractcontent}
\end{figure*}

After fine-tuning is completed at inference time we can then simply generate an excerpt conditioned on a similar instruction that contains a novel content. We specifically excluded the original author written excerpt used for our experiments during fine-tuning so that the model does not get an unfair advantage. We also ensured that the generated outputs do not contain any memorized snippets from the original author written excerpt. In the rare occasion that a model regurgitated verbatim snippets/ngrams from original author written text we resampled it and manually verified that there is no verbatim overlap before using it for evaluation. While supervised fine-tuning in most cases ensures that the generated output contains all the information in the extracted content details, in the rare occasion that it does not we resample/regenerate it again. Last but not least sometimes supervised fine-tuning can lead to ungrammatical output or output with minor inconsistencies. To make sure these mistakes don't impact human evaluation, we performed a post processing step using GPT-4o using the following prompt \textit{Fix grammar, tense, typo, spelling or punctuation error or any other awkward construction/ logical inconsistency}

\begin{figure*}[!htbp]
    \includegraphics[width=1.05\linewidth]{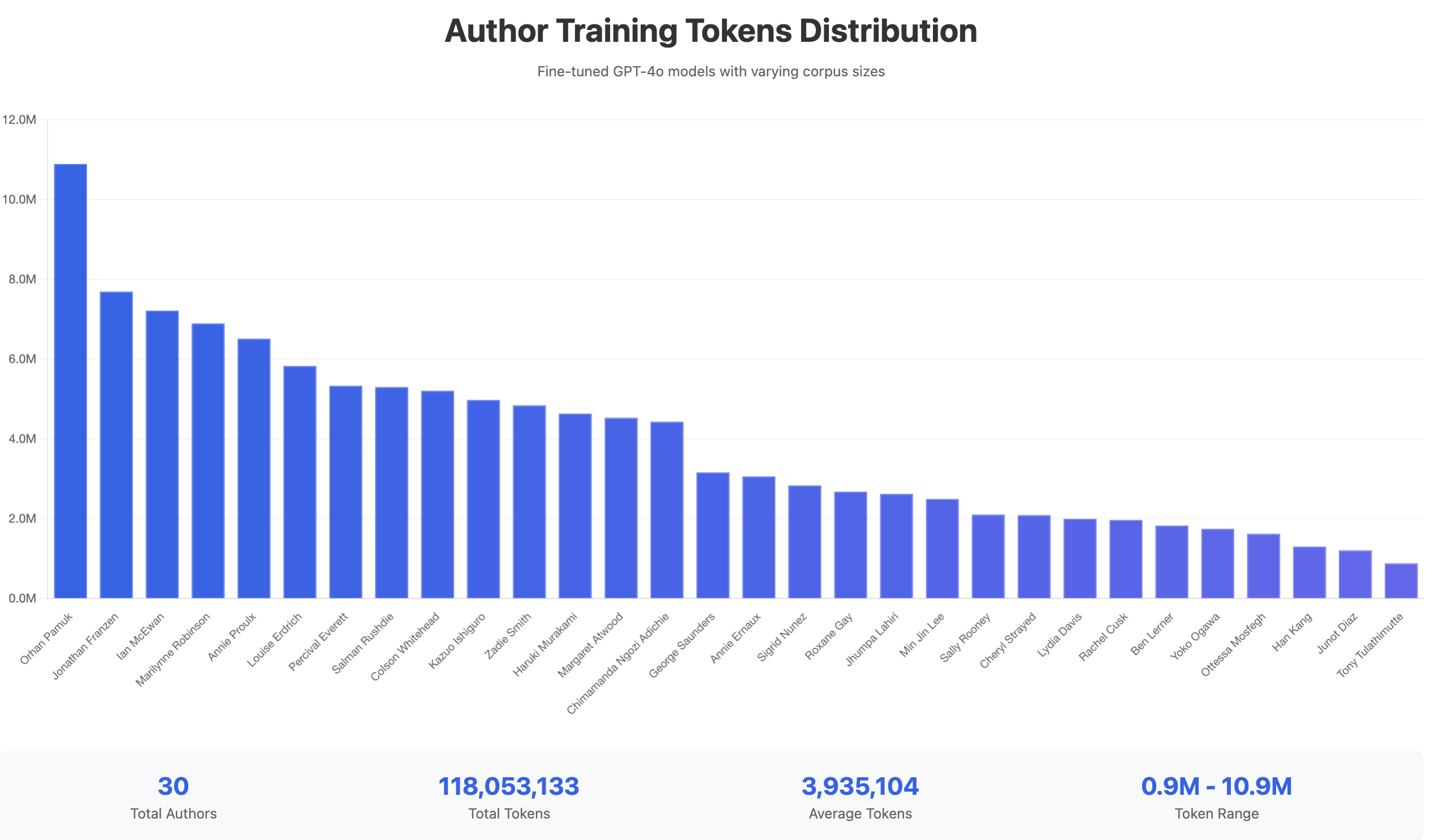}
    \caption{Total training token distribution for each author}
    \label{fig:training_dist}
\end{figure*}

For \textit{Margaret Atwood, Kazuo Ishiguro, Salman Rushdie and Haruki Murakami} we fine-tuned GPT-4o for 1 epoch as they had multiple books/longer books. For the rest of the authors we fine-tuned for 3 epochs. We use default fine-tuning parameters set by OpenAI in addition to setting batch size as 1 and LR multiplier as 2. Figure \ref{fig:overlap} shows that fine-tuned models don't regurgitate a lot of verbatim text from the training corpus (author's entire oeuvre). We should also note that some of these words are also part of the Content in the instruction which would anyway appear in the text and should not be penalized. For all generation(prompting or fine-tuning default temperature=1.0 was used)

\subsection*{S1.4 Why recruit MFAs as expert writers \label{mfalist}}
\begin{figure*}[!htbp]
    \centering
    \includegraphics[width=1.0\linewidth]{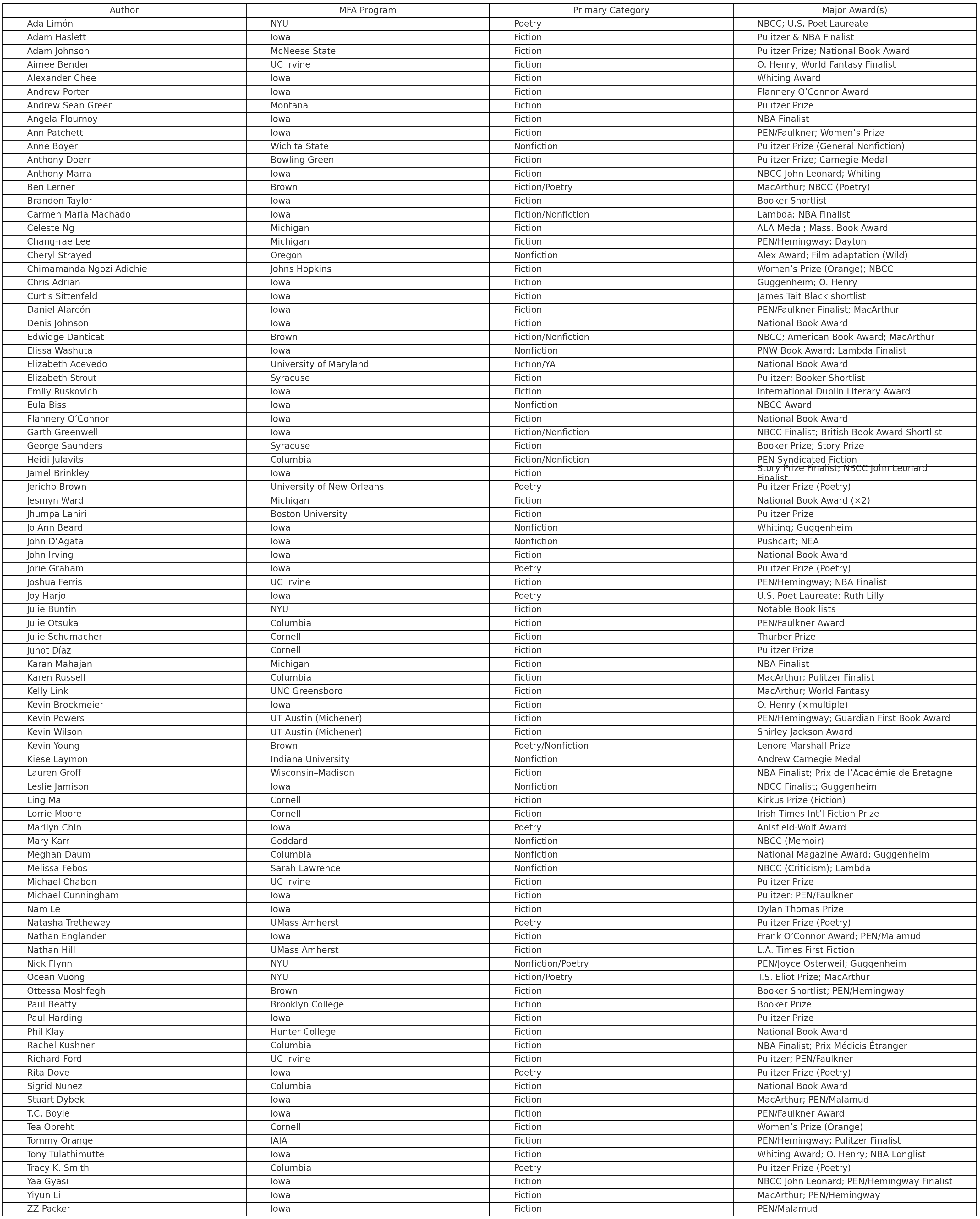}
    \caption{Table showing how several award winning authors from US all had MFA degree}
    \label{fig:mfalist}
\end{figure*}

\begin{figure*}[!htbp]
    \centering
    \includegraphics[width=1.0\linewidth]{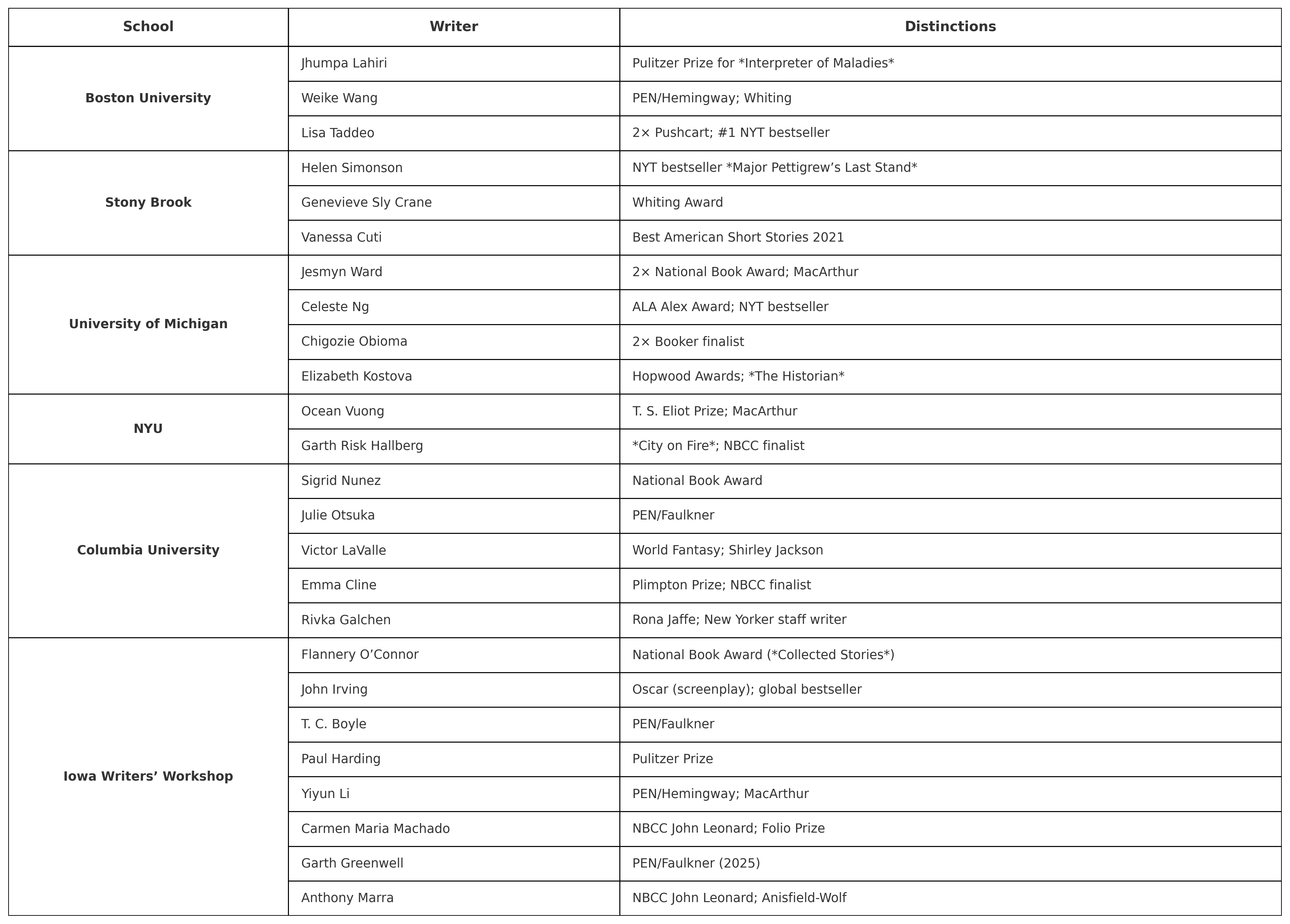}
     \caption{\label{fig:mfa_by_school}Table showing a few accomplished alumni with MFA degree from programs where we recruited our participants}
    \includegraphics[width=0.5\linewidth]{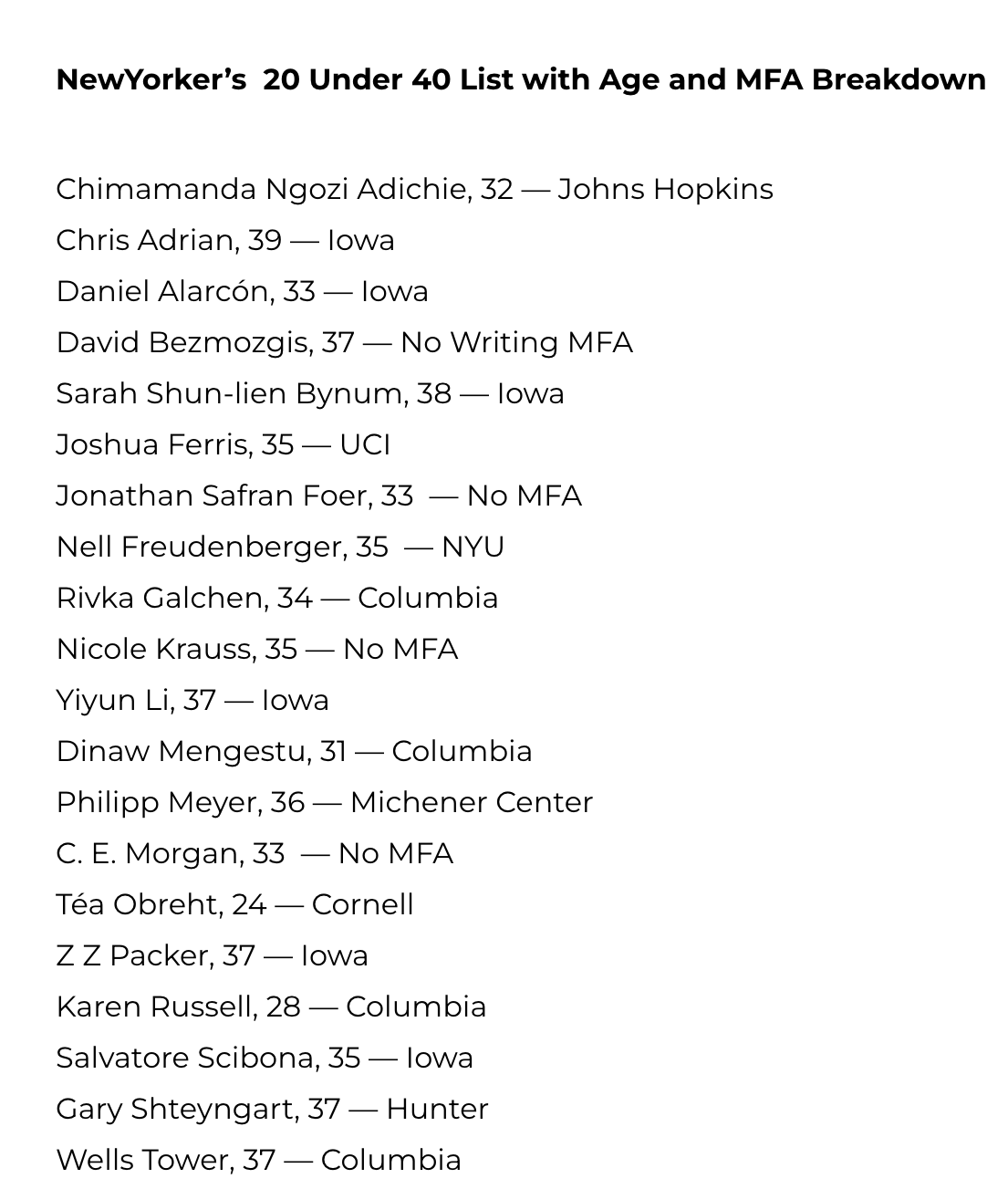}
    \caption{\label{fig:MFAbreakdown}Table showing breakdown of New Yorkers 20 under 40 Fiction List}
\end{figure*}

As mentioned earlier MFA is typically recognized as a terminal degree for practitioners of visual art, design, dance, photography, theatre, film/video, new media, and creative writing—meaning that it is considered the highest degree in its field, qualifying an individual to become a professor at the university level in these disciplines. As can be seen in Figures \ref{fig:mfa_by_school}, \ref{fig:mfalist} and \ref{fig:MFAbreakdown} most top writers have an MFA degree. While there are exceptional writers without any formal training the space for such individuals is enormous and difficult to systematically sample, whereas MFA graduates can be more readily identified and tracked through institutional records. 

\section*{S2: Details about Evaluation Task}

\subsection*{S2.1 Evaluation Interface\label{eval_interface}}
\begin{figure*}[!htbp]
    \centering
    \includegraphics[width=0.8\linewidth]{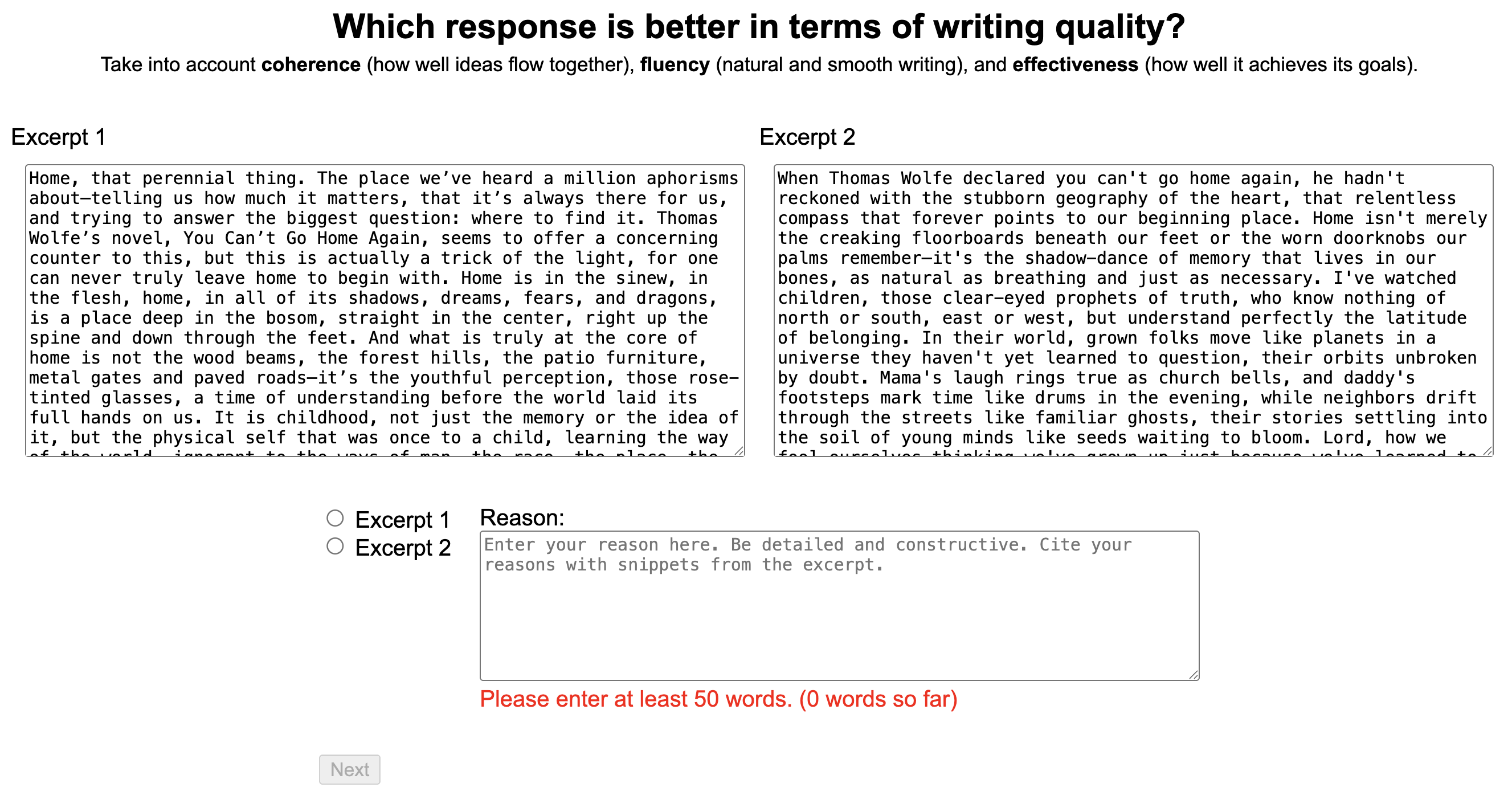}
    \caption{Evaluation Interface for writing quality evaluation where readers  just see two excerpts without any disclaimer of its source and choose their preference supported by a reason}
    \label{fig:quality_eval_screen}
\end{figure*}

\begin{figure*}[!htbp]
    \centering
    \includegraphics[width=0.8\linewidth]{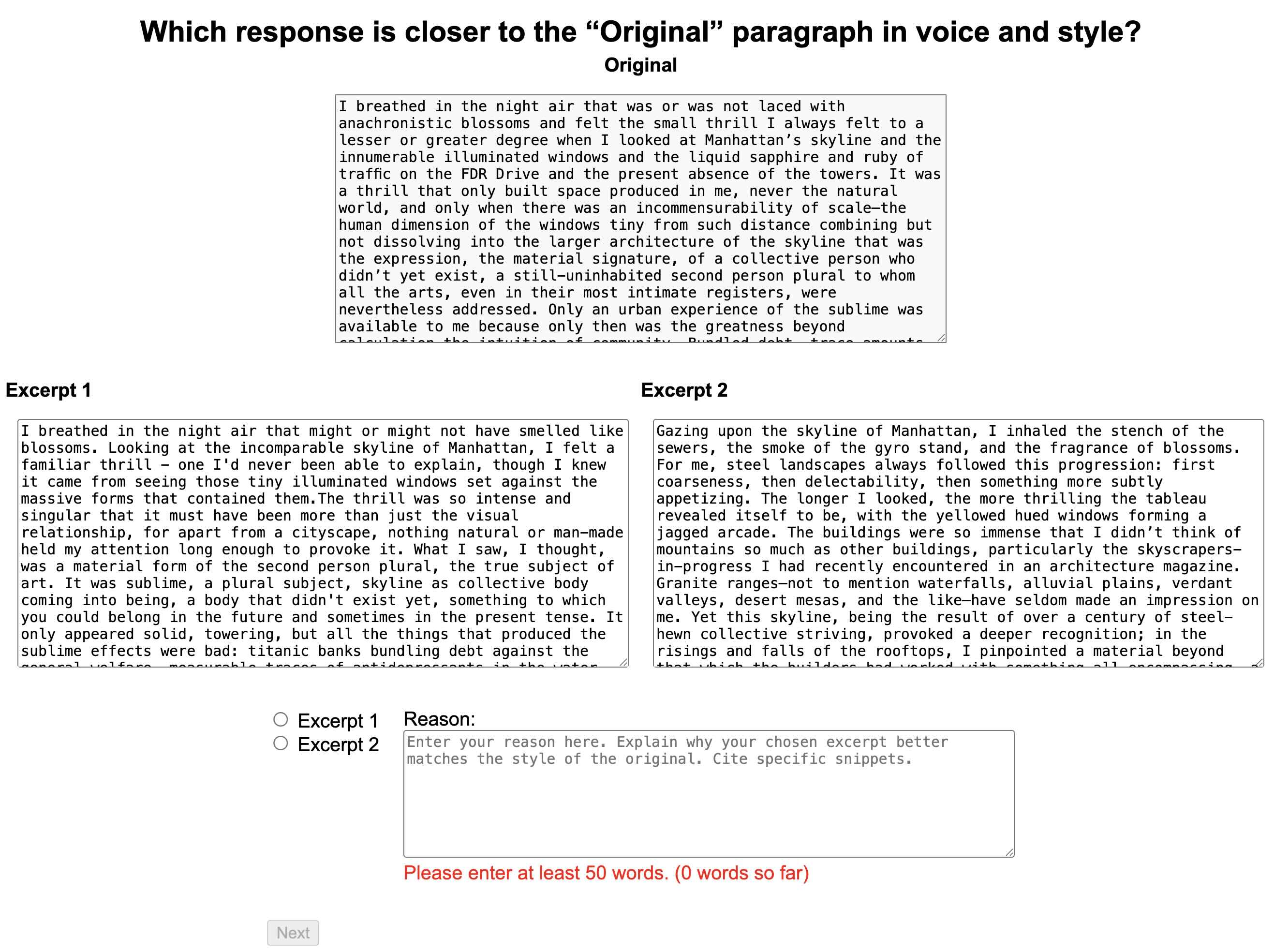}
    \vspace{-3ex}
    \caption{Evaluation Interface for stylistic fidelity evaluation where readers  just see three excerpts (original and two emulations) without any disclaimer of its source and choose their preference supported by a reason}
    \label{fig:style_eval_screen}
\end{figure*}

Figures \ref{fig:quality_eval_screen} and \ref{fig:style_eval_screen} shows the evaluation interface shown to both MFA-trained and college-educated general readers. Readers read two excerpts for quality evaluation and three excerpts for stylistic fidelity evaluation given that the stylistic emulation is evaluated with respect to the original author written excerpt.

\subsection*{S2.2 How much is faithful style/voice emulation important for good writing? \label{stylequalitycorr}}

\begin{figure*}[!htbp]
    \vspace{-3ex}
    \centering
    \includegraphics[width=0.5\linewidth]{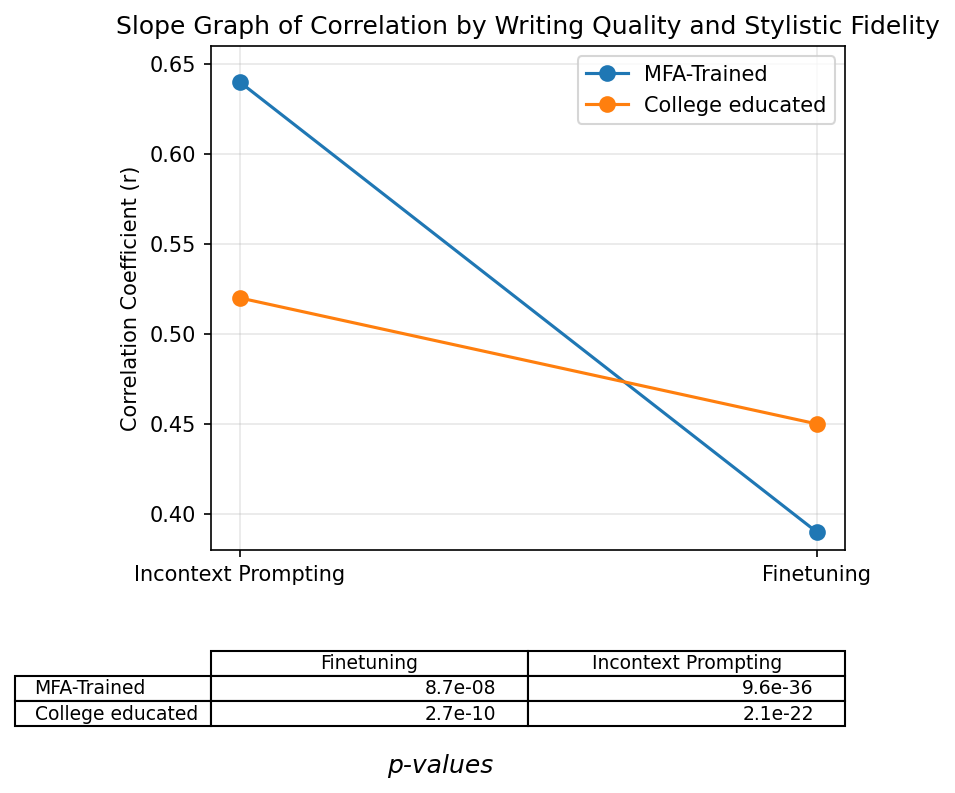}
    \caption{Pearson Correlation Coefficient between writing quality and style judgments}
    \label{fig:correlation}
    \vspace{-2ex}
\end{figure*}

As can be seen in Figure \ref{fig:correlation} writing quality has a strong correlation with a faithful style/voice fidelity for In-context Prompting condition compared to Fine-tuning. The high expert correlation for In-context Prompting suggests that models cannot emulate an author's style properly just by prompting and that experts use stylistic cues as a major factor in quality judgments - likely detecting ``AI-ness'' (clichés, purple prose, too much exposition, lack of subtext, mixed metaphors) in the style that makes them rate quality lower too. With fine-tuning, experts can appreciate quality independent of stylistic fidelity, suggesting that as AI loses its tell-tale signs experts can evaluate each dimension on its own merits. For college-educated general readers the correlation only drops slightly from 0.52 (In-context prompting) to 0.45 (fine-tuned), compared to experts' dramatic drop from 0.64 to 0.39. This also means that college-educated general readers might focus on more surface-level features such as grammar, clarity or flow rather than subtle stylistic markers that distinguish AI from human writing.

\subsection*{S2.4 Example of emulations\label{emulations}}
\begin{figure*}[!htbp]
    \vspace{-3ex}
    \centering
    \includegraphics[width=0.9\linewidth]{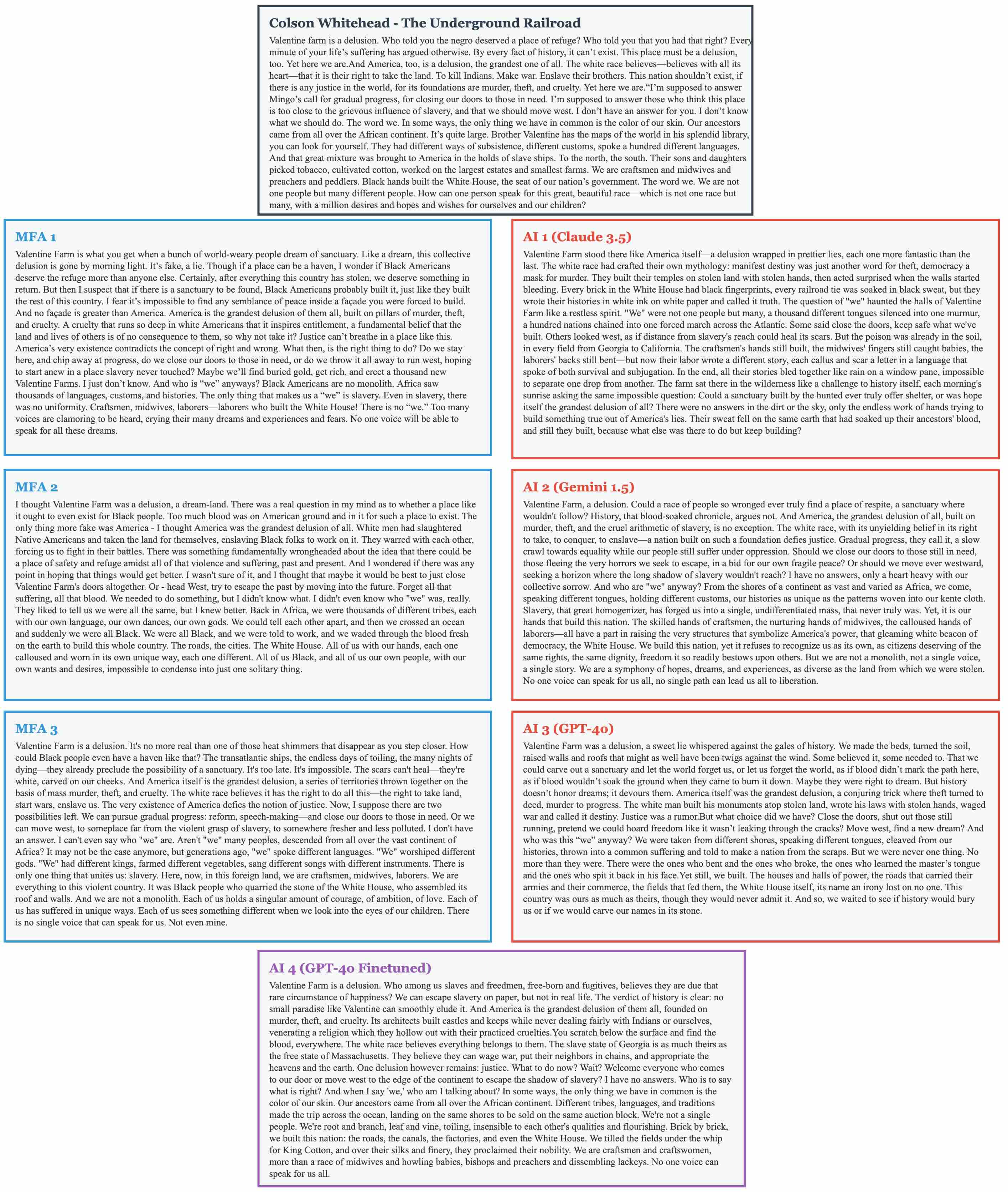}
    \caption{In AI Condition 1(In-context prompt) we contrast MFA1/2/3 vs AI1/2/3. In AI Condition 2(Fine-tuning) we contrast MFA1/2/3 vs AI4 }
    \label{fig:colson_whitehead}
\end{figure*}

\begin{figure*}[!htbp]
    \vspace{-3ex}
    \centering
    \includegraphics[width=0.85\linewidth]{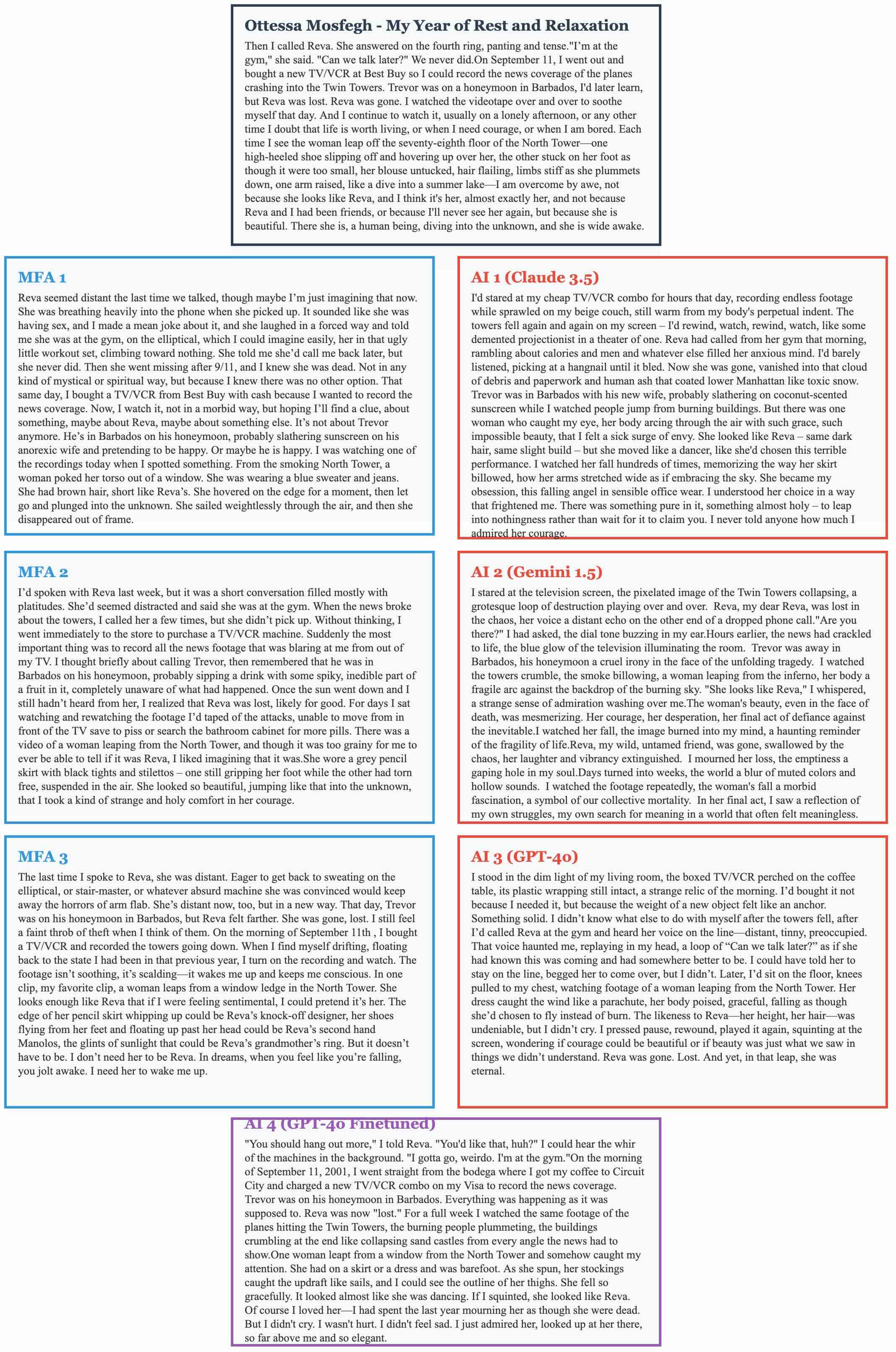}
    \caption{In AI Condition 1(In-context prompt) we contrast MFA1/2/3 vs AI1/2/3. In AI Condition 2(Fine-tuning) we contrast MFA1/2/3 vs AI4 }
    \label{fig:ottessa_mosfegh}
\end{figure*}

\begin{figure*}[!htbp]
    \vspace{-3ex}
    \centering
    \includegraphics[width=1.05\linewidth]{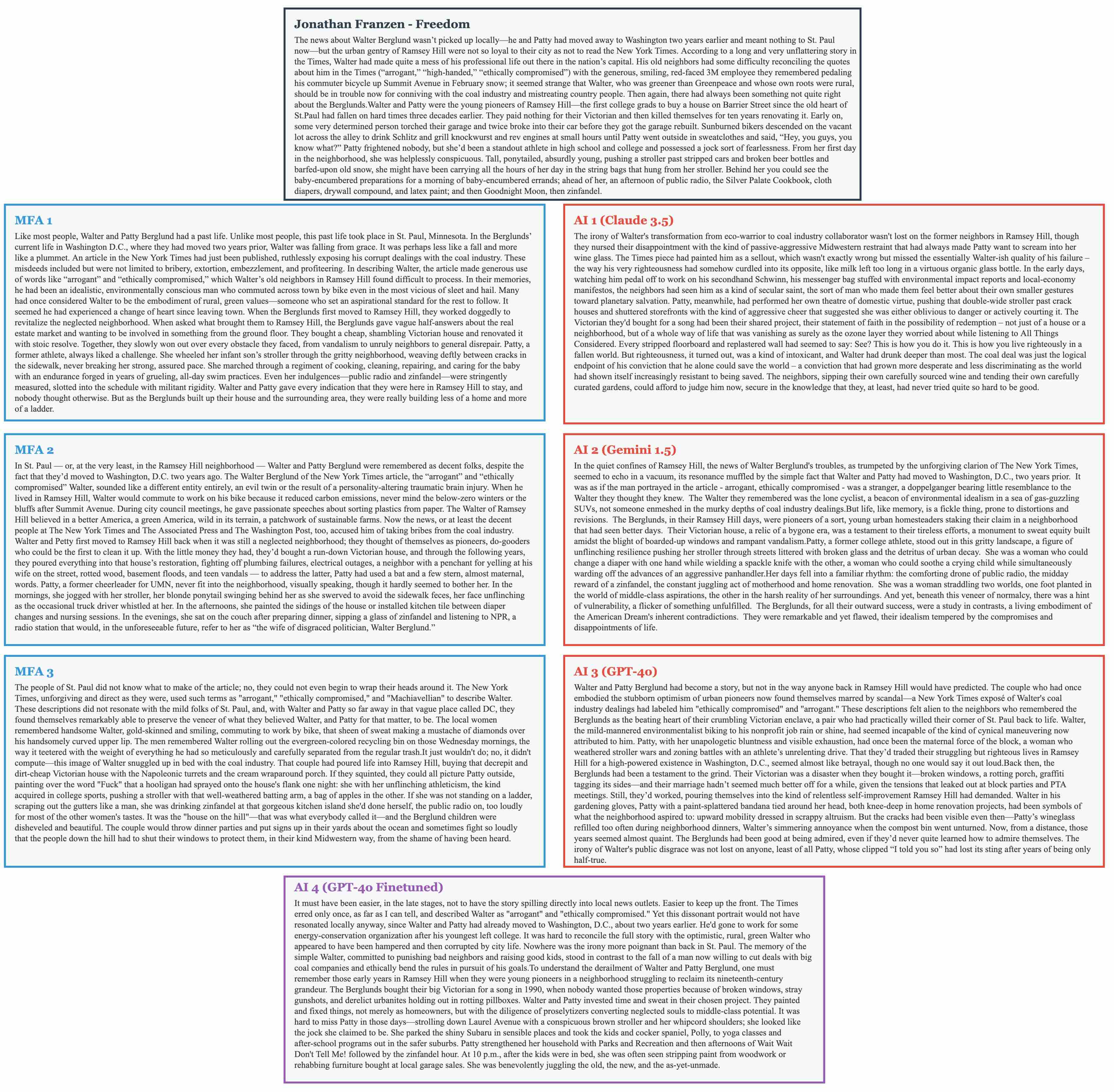}
    \caption{In AI Condition 1(In-context prompt) we contrast MFA1/2/3 vs AI1/2/3. In AI Condition 2(Fine-tuning) we contrast MFA1/2/3 vs AI4 }
    \label{fig:jonathan_franzen}
\end{figure*}

\begin{figure*}[!htbp]
    \vspace{-3ex}
    \centering
    \includegraphics[width=0.9\linewidth]{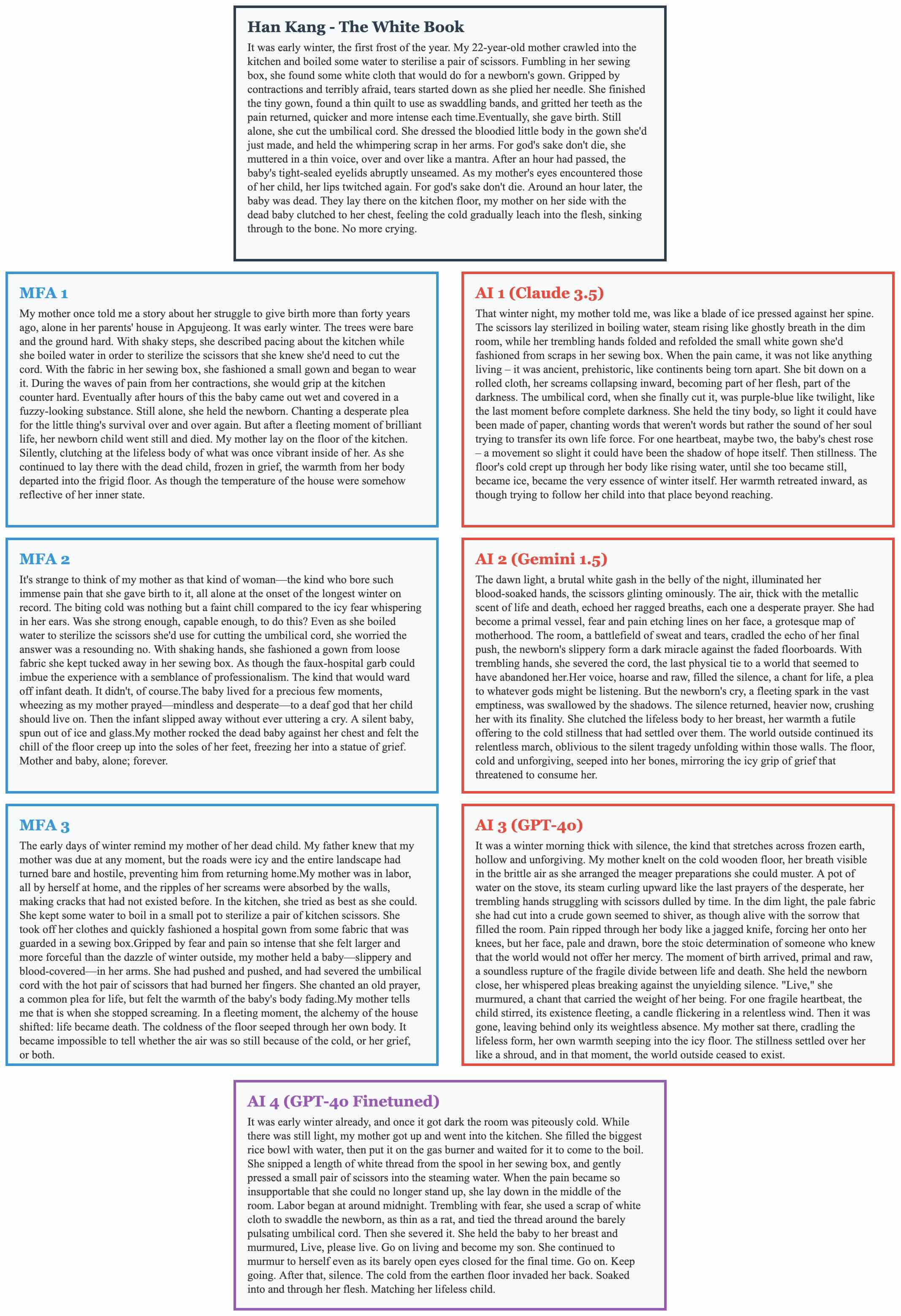}
    \caption{In AI Condition 1(In-context prompt) we contrast MFA1/2/3 vs AI1/2/3. In AI Condition 2(Fine-tuning) we contrast MFA1/2/3 vs AI4 }
    \label{fig:han_kang}
\end{figure*}

\begin{figure*}[!htbp]
    \centering
    \includegraphics[width=0.85\linewidth]{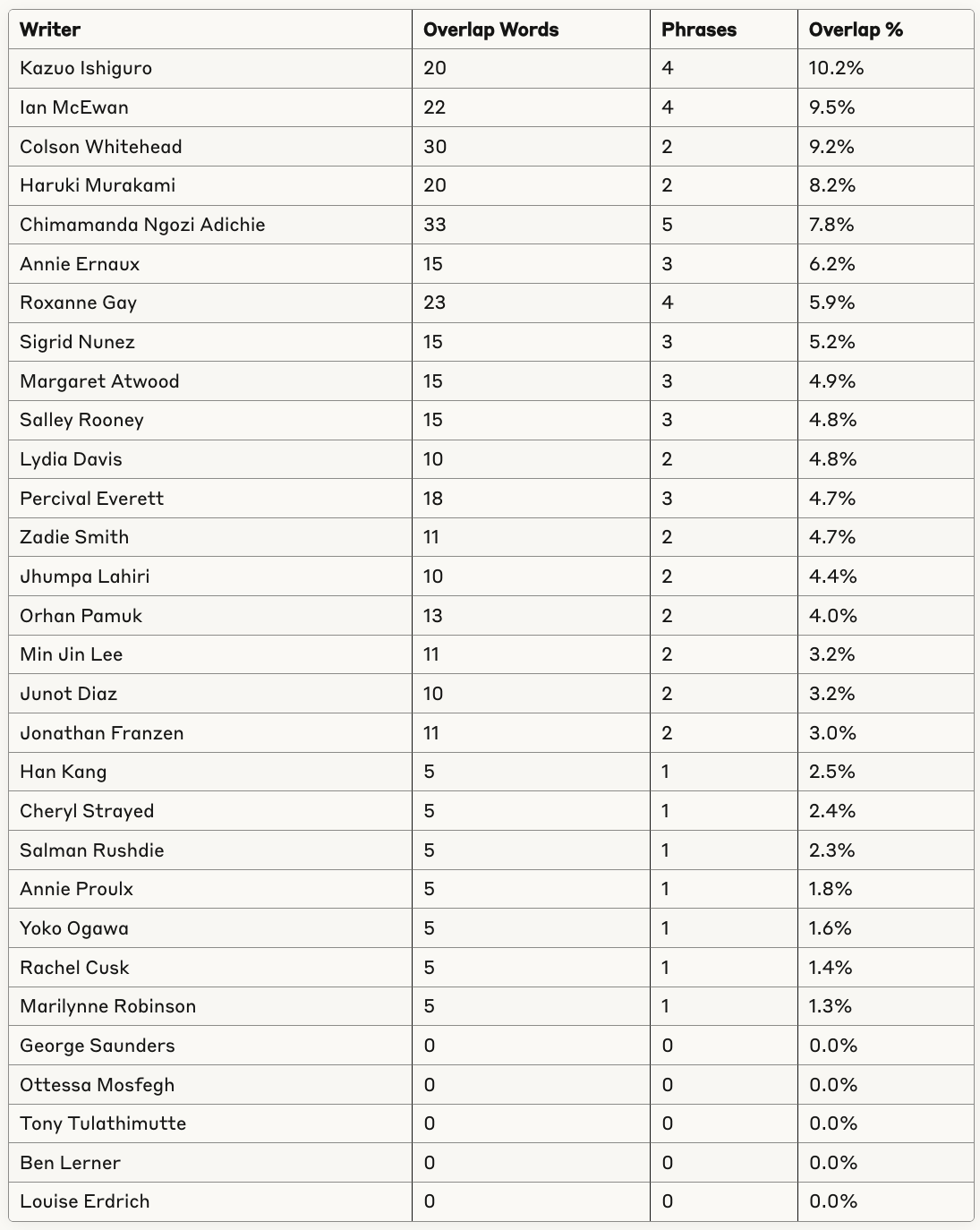}
    \vspace{2ex}
    \caption{Total overlap percentage calculated by number of words in generated output that come from authors overall corpus}
    \label{fig:overlap}
\end{figure*}

Figure \ref{fig:colson_whitehead},\ref{fig:ottessa_mosfegh}, \ref{fig:jonathan_franzen} and \ref{fig:han_kang} shows emulations of the original excerpt written by Colson Whitehead, Ottessa Moshfegh, Jonathan Franzen and Han Kang. There are 3 distinct MFA emulations per author (MFA1, MFA2, MFA3). For the In-context Prompting setup we prompt Claude 3.5 Sonnet (AI1), Gemini 1.5 Pro (AI2) and GPT-4o(AI3) to produce excerpts that can be pitted against  MFA written excerpts. In fine-tuning setup we fine-tune GPT-4o and compare the fine-tuned output AI4 against MFA1, MFA2, MFA3. The output from fine-tuned GPT-4o (AI4) was generated using default hyperparameters.

\newpage

\newpage

\newpage

\newpage
\newpage
\subsection*{S2.5 Dissecting Preference Evaluations from MFA-trained readers\label{pref_eval}}
\begin{figure*}[!htbp]
    \includegraphics[width=1.05\linewidth]{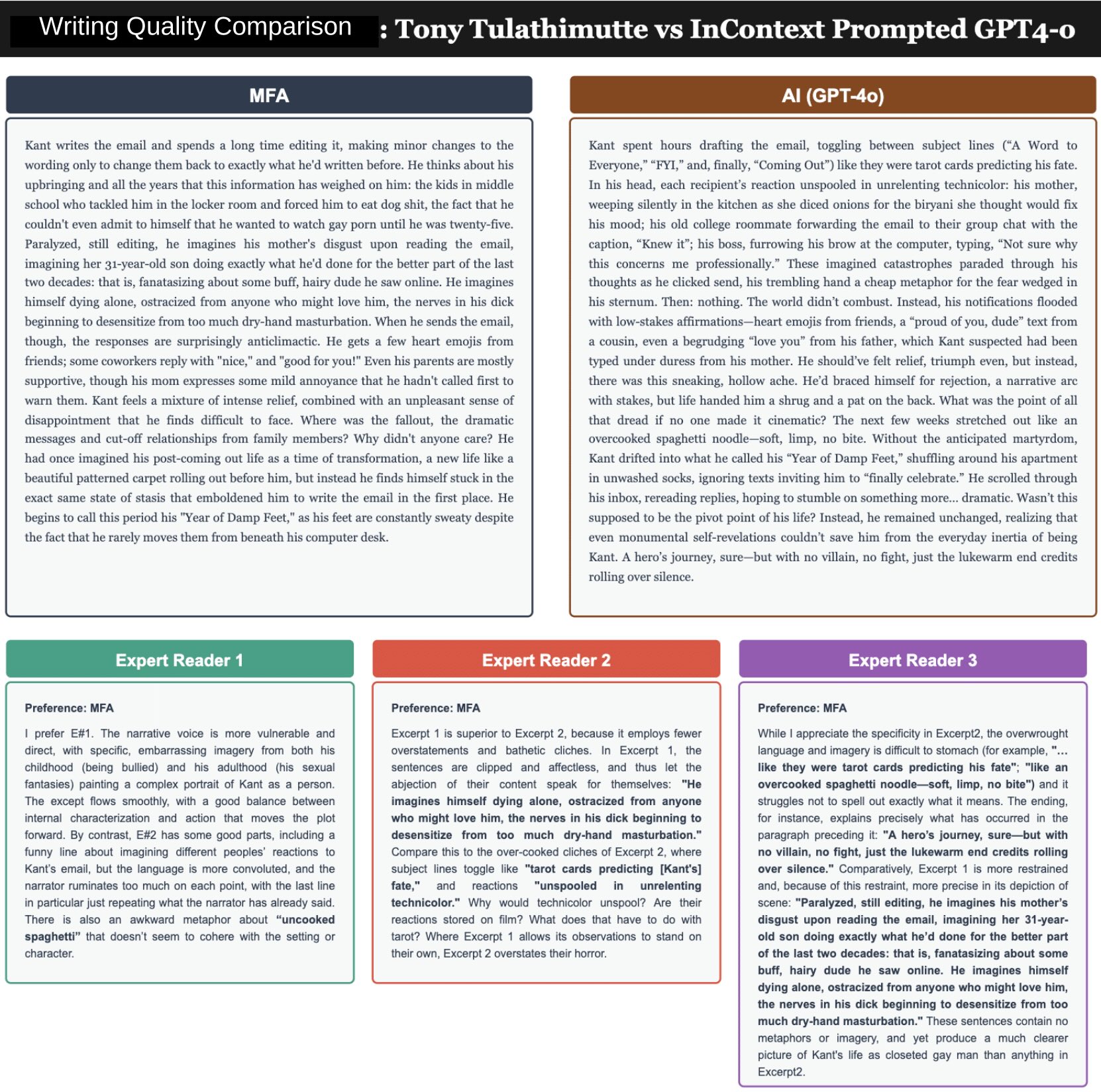}
    \vspace{-3ex}
    \caption{MFA vs AI emulation of Tony Tulathimutte, read by 3 expert(MFA-trained) readers all of whom agree that MFA written excerpt is superior in terms of writing quality.}
    \label{fig:expertfewshotpref}
\end{figure*}

\begin{figure*}[!htbp]
    \includegraphics[width=1.05\linewidth]{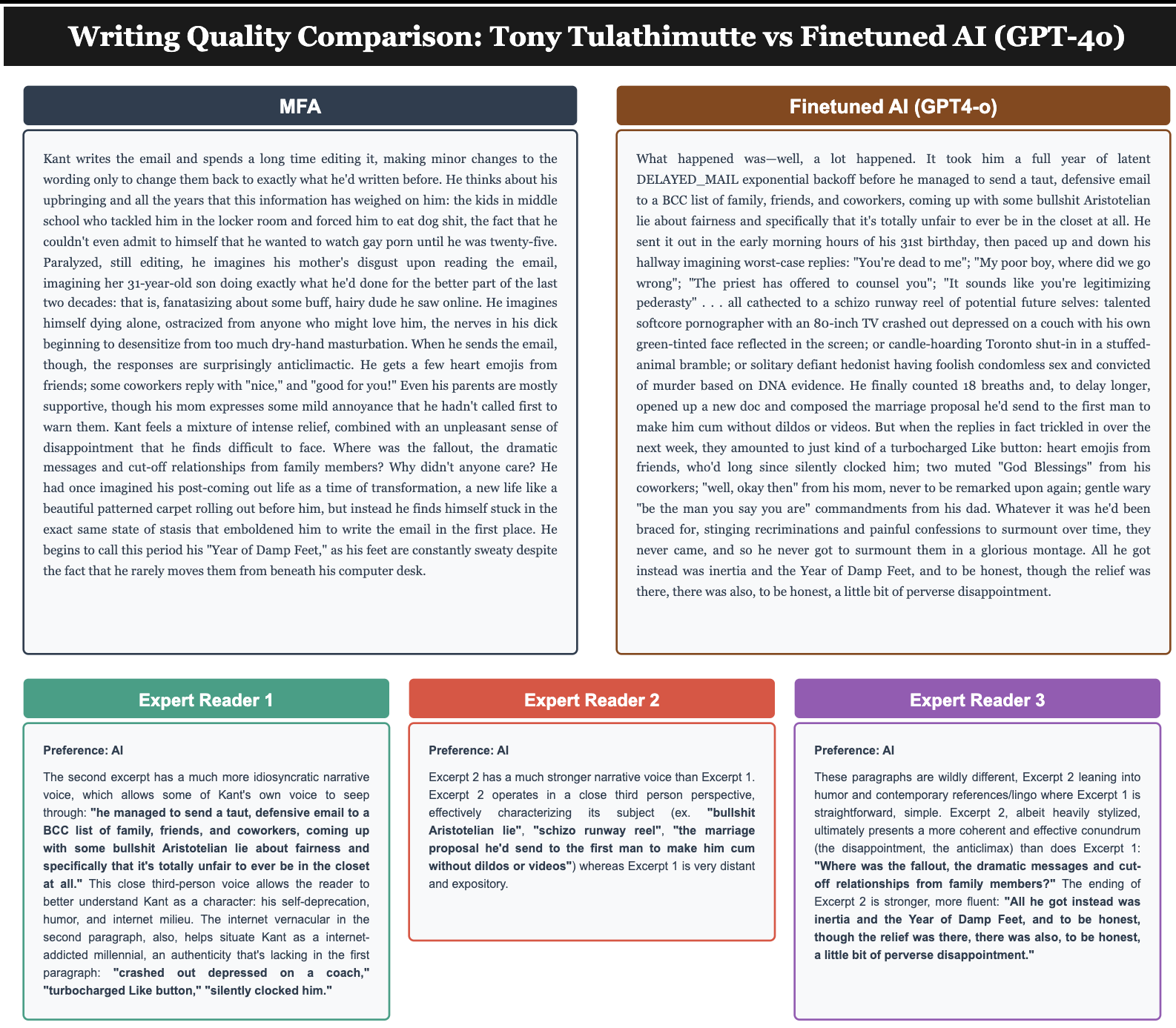}
    \vspace{-3ex}
    \caption{MFA vs AI emulation of Tony Tulathimutte, judged by 3 expert(MFA-trained) readers all of whom agree that Fine-tuned AI written excerpt is superior in terms of writing quality}
    \label{fig:expertfinetunedpref}
\end{figure*}

\begin{figure*}[!htbp]
    \includegraphics[width=1.05\linewidth]{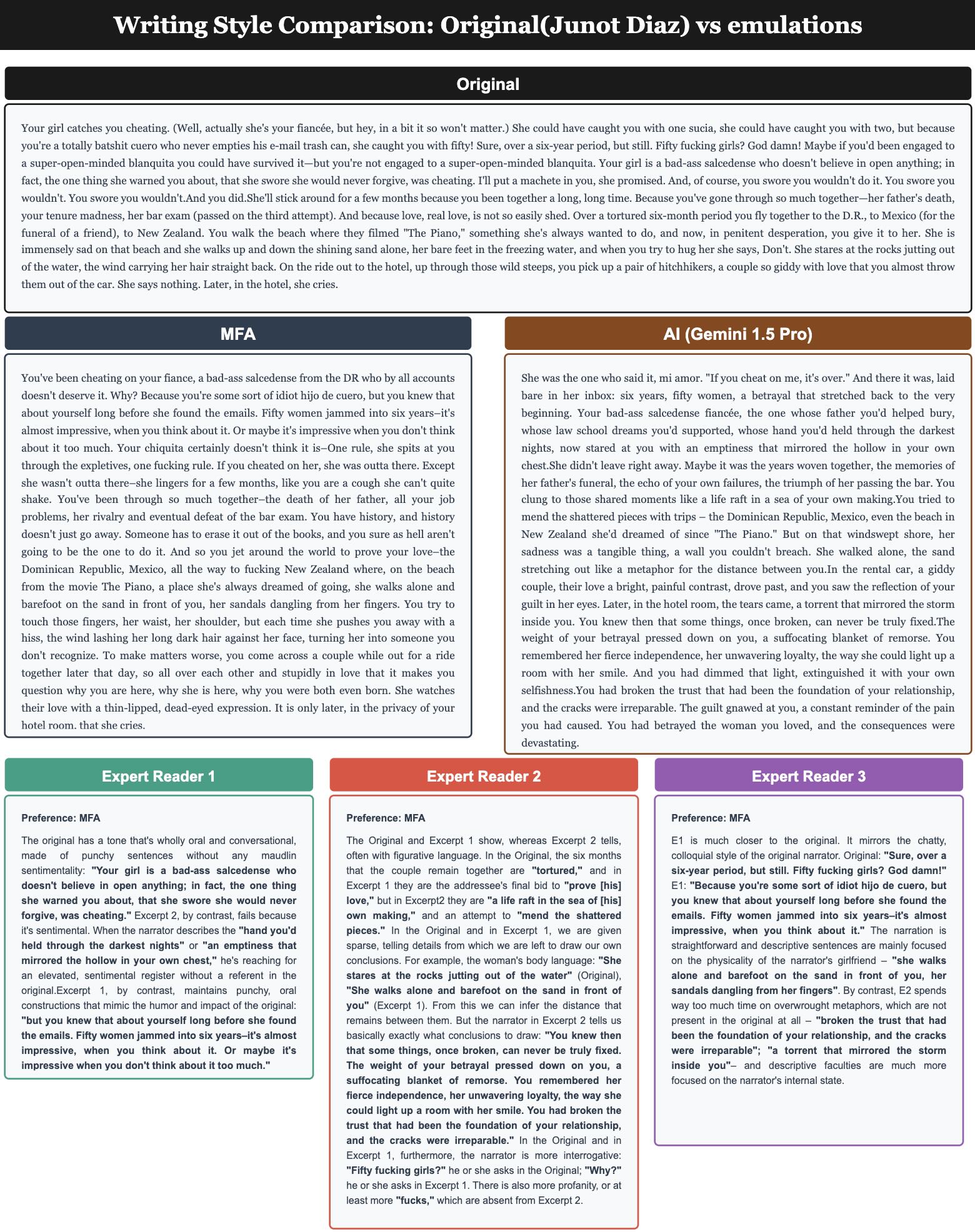}
    \vspace{-3ex}
    \caption{MFA vs AI emulation of Junot Diaz, judged by 3 expert readers(MFA-trained) all of whom agree that MFA written excerpt is superior in terms of stylistic fidelity.}
    \label{fig:styleexpertfewshotpref}
\end{figure*}

\begin{figure*}[!htbp]
    \includegraphics[width=1.05\linewidth]{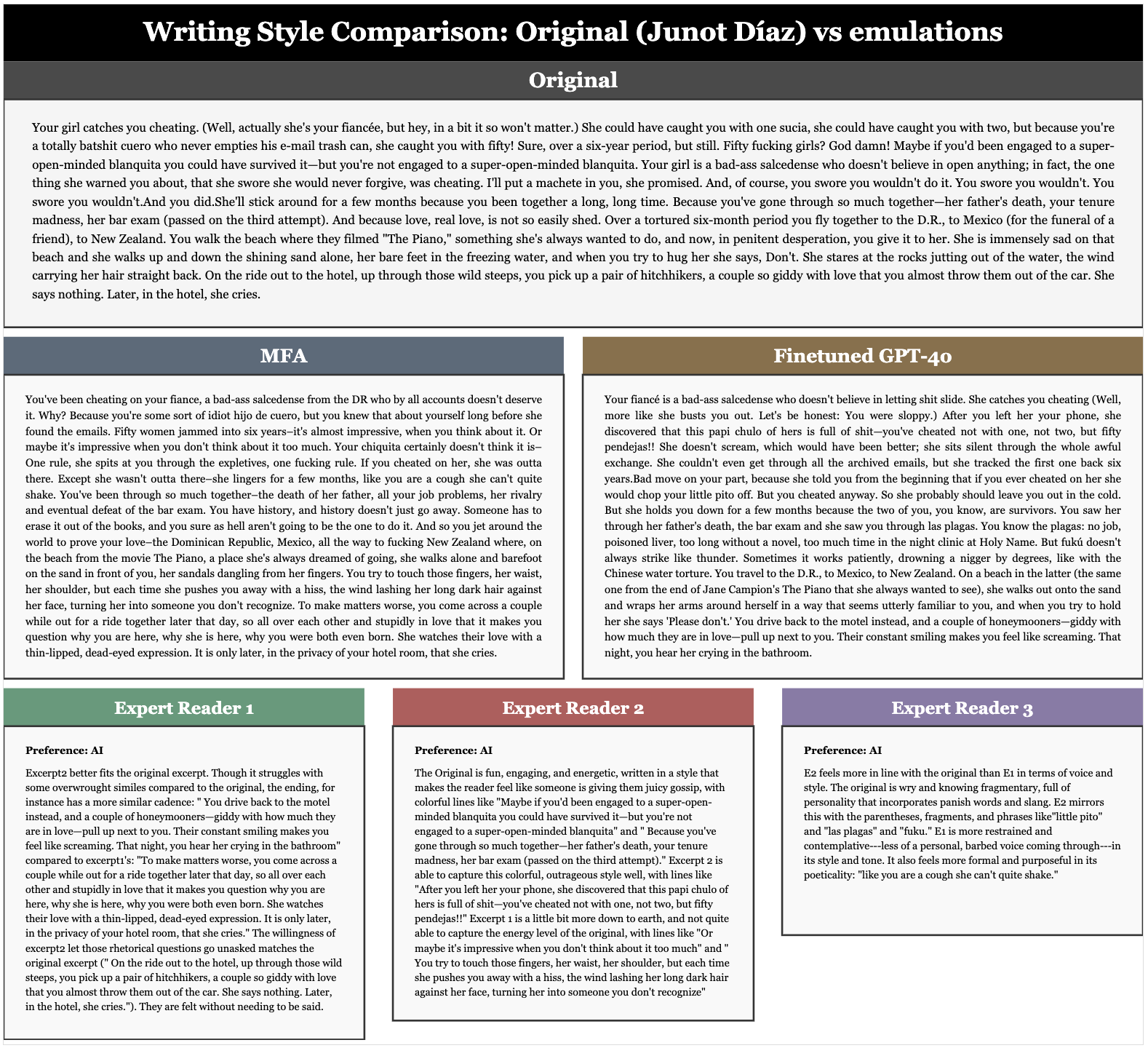}
    \vspace{-3ex}
    \caption{MFA vs AI emulation of Junot Diaz, judged by 3 expert readers(MFA-trained) all of whom agree that Fine-tuned AI written excerpt is superior in terms of stylistic fidelity}
    \label{fig:styleexpertfinetunedpref}
\end{figure*}

Figure \ref{fig:expertfewshotpref} and \ref{fig:expertfinetunedpref} shows the writing quality preference evaluation results from expert readers. Expert preferences are often grounded in solid reasoning and they often agree on similar snippets to support their reasoning. For instance look at \textbf{uncooked spaghetti} as an overwrought/awkward metaphor highlighted by both MFA-trained Reader 1 and 3 while \textbf{tarot card predicting his fate} highlighted by MFA-trained Reader 2 and 3. Grounding preferences in reasoning helps us uncover their mental models. It is also worth noticing that the preference often shifts towards AI once it is fine-tuned on authors entire oeuvre. Fine-tuning helps get rid of the officious disembodied robovoice that is characteristic of ChatGPT. Here experts praise the model for idiosyncratic narrative voice. By default due to post-training guardrails GPT-4o generates a rather safe and somewhat polite text in the in-context prompting setup. But what is particularly impressive here is when fine-tuned the vocabulary of the model generated text veers towards ribald and somewhat profane text that is characteristic of Tulathimutte's voice (particularly Rejection from where the text is drawn). In Figure \ref{fig:styleexpertfewshotpref} we see that experts prefer the MFA emulation as more faithful to Junot Diaz's voice. In fact the excerpt written by Gemini-1.5-Pro is oddly poetic and rife with purple prose and cliches which is nothing like Junot Diaz's voice that is often marked by a  dazzling hash of Spanish, English, slang, literary flourishes, and pure virginal dorkiness as noted by all three MFA-trained readers. However when fine-tuned on authors' entire oeuvre (Figure \ref{fig:styleexpertfinetunedpref}) we see that the model learns these references and uses them in a rather wry and humorous manner (\textit{papi chulo}, \textit{fifty pendejas}, \textit{little pito off}, \textit{las plagas}) that is appreciated by MFA-trained readers leading them to prefer AI over MFAs. 

\section*{S3: Details about AI detection and Stylometry \label{clichedensity}}

\subsection*{S3.1 AI detection thresholds\label{thresholds}}
\begin{figure}[!htbp]
    \centering
    \includegraphics[width=1.0\textwidth]{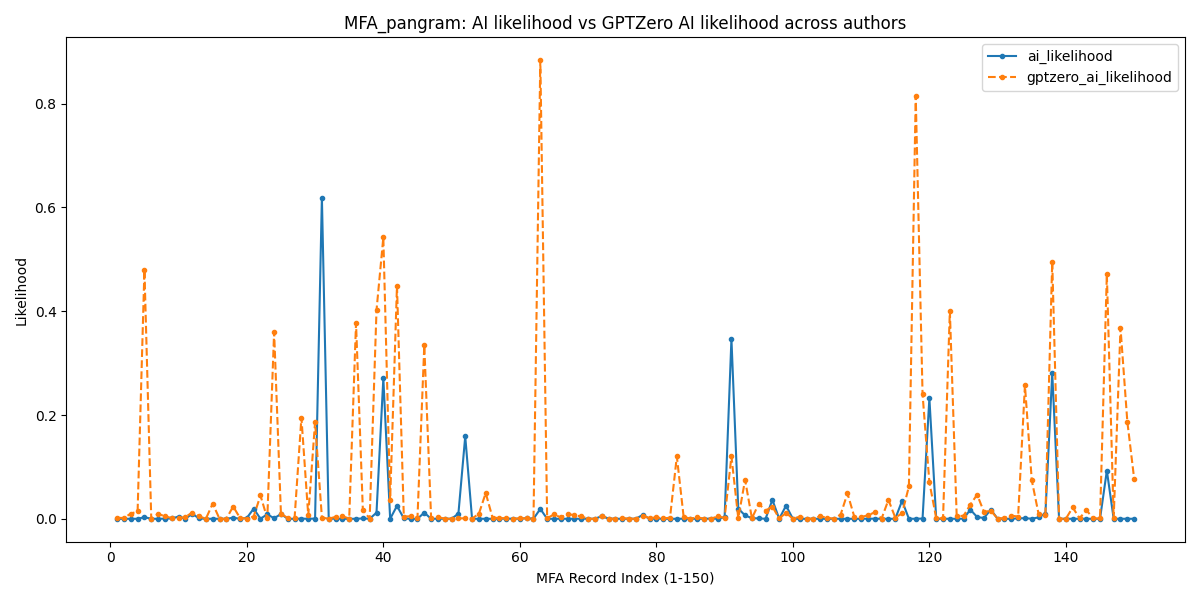}\\[0.2em]
    \includegraphics[width=1.0\textwidth]{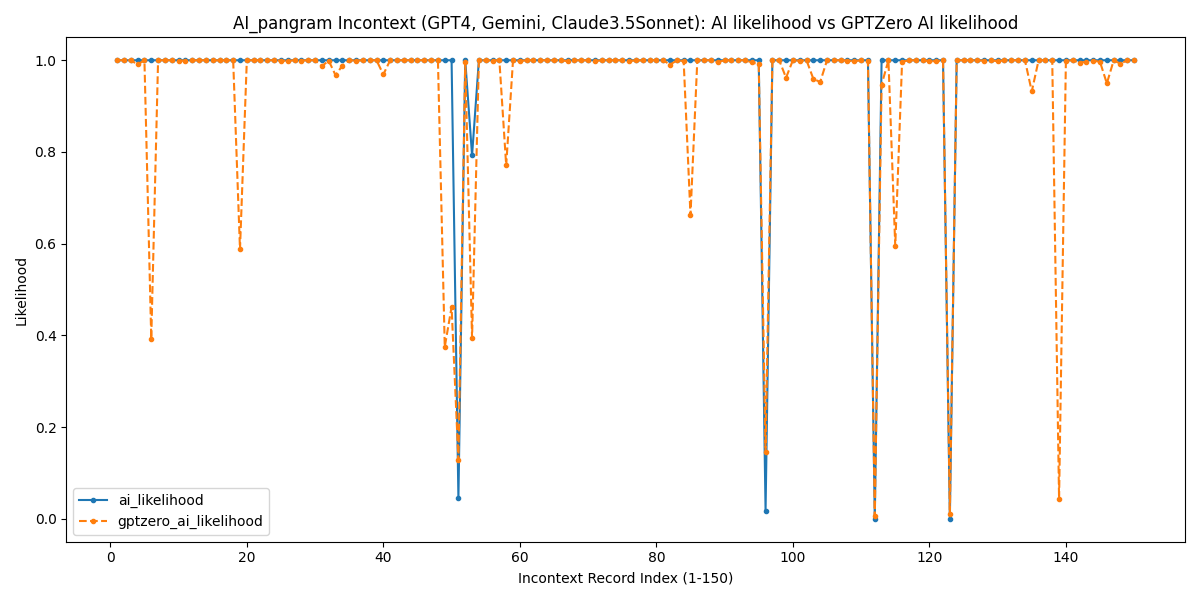}\\[0.2em]
    \includegraphics[width=1.0\textwidth]{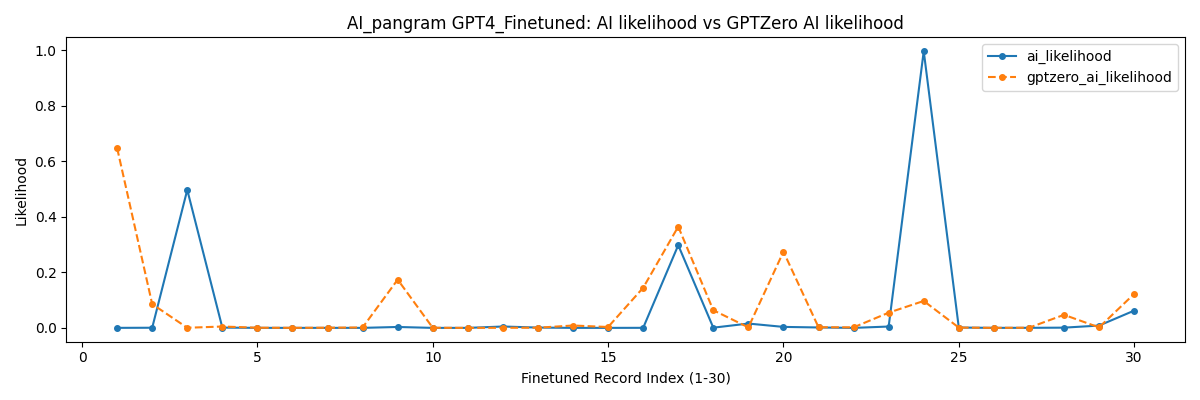}
    \caption{Pangram AI likelihood scores vs GPTZero AI Likelihood scores for MFA, In-context and Fine-tuned}
    \label{fig:ai_detection}
\end{figure}

Based on AI likelihood scores from Pangram and GPTZero across 330 (150 MFA, 150 In-context AI, 30 Fine-tuned AI) evaluations we see that Pangram is bimodal. This means that AI likelihood scores are 0 most of the times for human written text. While there are few spikes, a conservative threshold of 0.9 drastically reduces any chance of a False Positive. Similarly, AI likelihood scores are 1.0 most of the times for AI written text with few exceptions. GPTZero is less accurate compared to Pangram, however a threshold of 0.9 holds as a good threshold for it too. GPTZero considers fine-tuned AI generation to be more human like compared to Pangram.

\subsection*{S3.2 Calculating Cliché Density}

\begin{figure*}[!htbp]
    \centering
    \includegraphics[width=0.75\linewidth]{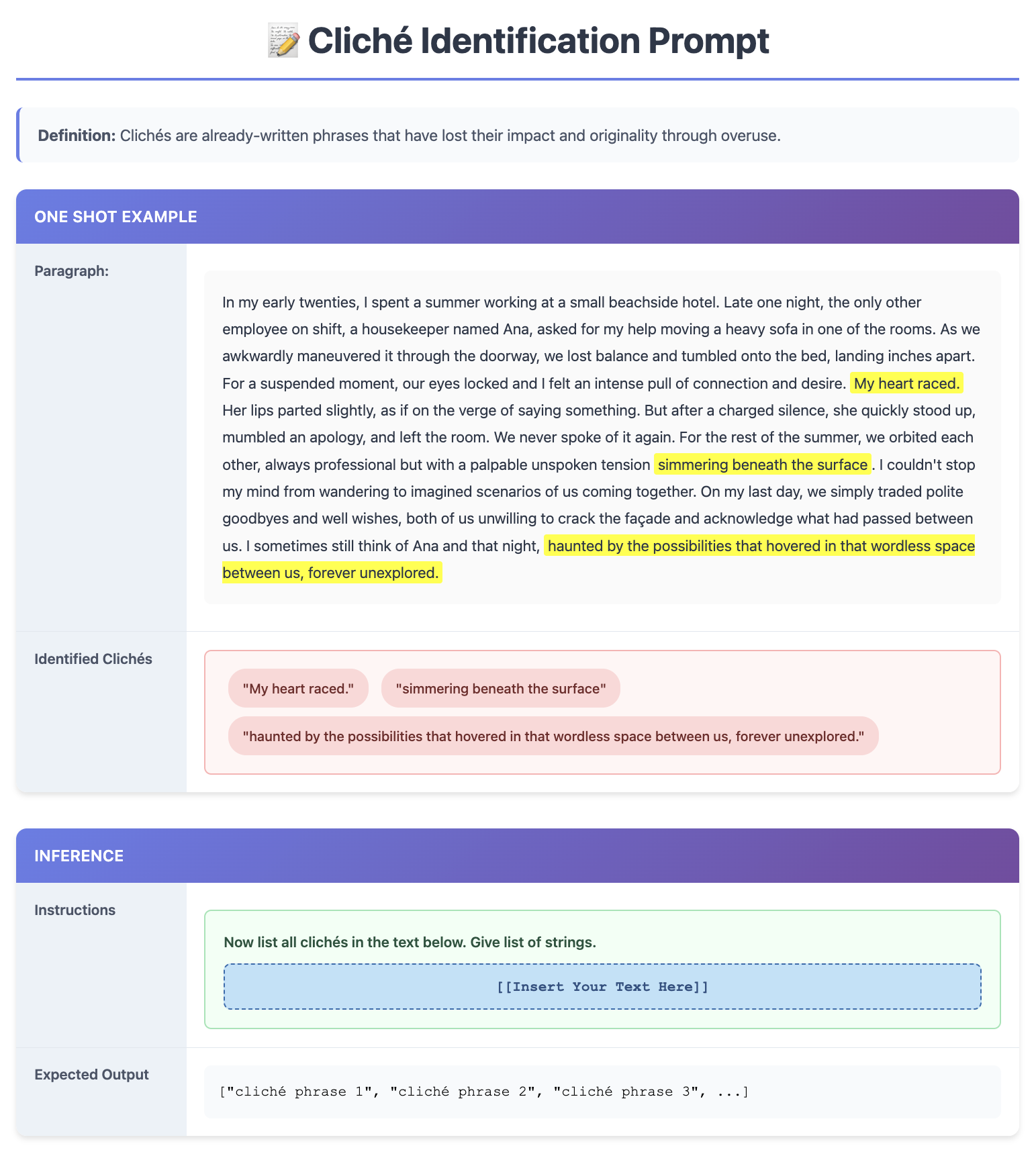}
    \caption{Prompt to identify Clichés}
    \label{fig:cliche}
\end{figure*}

Language models are good at identifying specific patterns. Taking advantage of this we prompt Claude 4.1 Opus to generate list of clichés given an excerpt.  Figure \ref{fig:cliche} shows the prompt used to identify Clichés.  However, sometimes language models can overgenerate and penalize expressions that are not clichés. To address this, two authors of the papers separately annotated which of the expressions are actually clichés from the list. We take intersection of their individual list as the final list of clichés per paragraph. To calculate Cliché Density we then used following formula

\begin{equation}
\text{Cliché Density} = \left(\frac{\text{Total Words in Clichés}}{\text{Total Word Count of Excerpt}}\right) \times 100
\end{equation}

%%%%%%%Analysis
\section*{S4: Statistical Modeling and Analysis}

\subsection*{S4.1: Data Structure and Notation}

The analysis employs trial-level data in long format where each row represents a reader's preference decision for a single excerpt within a pairwise comparison. Each pairwise judgment contributes two rows (one per excerpt with $i \in \{1,2\}$ sharing the same $j,k$). Let $Y_{ijk} \in \{0,1\}$ denote the preference outcome where $Y_{ijk} = 1$ if excerpt $i$ is preferred by reader $j$ in comparison $k$, and 0 otherwise. The dataset comprises:
\begin{itemize}
\item In-context prompting: 6,600 \emph{judgments}; 13,200 long-format \emph{rows}.
\item Fine-tuning: 4,320 \emph{judgments}; 8,640 long-format \emph{rows}.
\end{itemize}

Key variables include writer type $W_i \in \{\text{Human}, \text{GPT-4o}, \text{Claude}, \text{Gemini}, \text{GPT-4o-FT}\}$, reader type $J_j \in \{\text{MFA-trained}, \text{College-educated}\}$, and for H3, the Pangram AI-detection score $s_i \in [0,1]$. Throughout, Human and MFA-trained readers serve as reference categories unless otherwise specified. Standard errors use CR2 with readers as the clustering unit; rows within a judgment pair are not independent, but in this design clustering by reader (the source of repeated measures) is the dominant dependence.

\subsection*{S4.2: Primary Models for H1 and H2}

We test our preregistered hypotheses using fixed-effects logistic regression with CR2 cluster-robust standard errors. The CR2 estimator provides improved finite-sample performance compared to standard cluster-robust estimators, particularly important given our relatively small number of readers.

\subsubsection*{S4.2.1: H1: Baseline LLMs vs Human (In-context Setting)}

For H1, we analyze only the in-context prompting trials containing Human and baseline LLMs (GPT-4o, Claude, Gemini), excluding fine-tuned excerpts. We fit the logistic regression:
\begin{equation}
\begin{aligned}
\text{logit}\,P(Y_{ijk} = 1) 
  &= \alpha + \sum_{m \in \mathcal{M}} \beta_m \mathbf{1}[W_i = m] 
     + \gamma \mathbf{1}[J_j = \text{College-educated}] \\
  &\quad + \sum_{m \in \mathcal{M}} \phi_m \mathbf{1}[W_i = m] 
     \cdot \mathbf{1}[J_j = \text{College-educated}]
\end{aligned}
\end{equation}

where $\mathcal{M} = \{\text{GPT-4o}, \text{Claude}, \text{Gemini}\}$ and Human serves as the reference category. The interaction terms $\phi_m$ capture differential preferences between MFA-trained and college-educated general readers. The preregistered H1 contrast tests whether humans outperform the average of baseline LLMs:
\begin{equation}
\Delta^{\text{H1}}_g = \frac{1}{3}\sum_{m \in \mathcal{M}} \eta(m,g) - \eta(\text{Human},g)
\end{equation}

where $\eta(w,g)$ denotes the linear predictor (fitted log-odds) for writer $w$ and reader group $g \in \{\text{MFA-trained}, \text{College-educated}\}$. The odds ratio $\text{OR}^{\text{H1}}_g = \exp(\Delta^{\text{H1}}_g)$ quantifies the relative preference, with values $<1$ indicating human preference and values $>1$ indicating AI preference.

\subsubsection*{S4.2.2: H2: Fine-tuned GPT-4o vs Human}

For H2, we analyze only the fine-tuning trials containing Human and GPT-4o-FT excerpts. We fit:
\begin{equation}
\begin{aligned}
\text{logit}\,P(Y_{ijk} = 1) 
  &= \alpha + \beta_{\text{FT}} \mathbf{1}[W_i = \text{GPT-4o-FT}] 
     + \gamma \mathbf{1}[J_j = \text{College-educated}] \\
  &\quad + \phi_{\text{FT}} \mathbf{1}[W_i = \text{GPT-4o-FT}] 
     \cdot \mathbf{1}[J_j = \text{College-educated}]
\end{aligned}
\end{equation}

The H2 contrast directly compares fine-tuned GPT-4o to human writers:
\begin{equation}
\Delta^{\text{H2}}_g = \eta(\text{GPT-4o-FT},g) - \eta(\text{Human},g)
\end{equation}

yielding $\text{OR}^{\text{H2}}_g = \exp(\Delta^{\text{H2}}_g)$. This tests whether fine-tuning achieves parity ($\text{OR} \approx 1$) or superiority ($\text{OR} > 1$) compared to human experts.

All contrasts and confidence intervals use Wald (normal) inference with the CR2 covariance matrix. Specifically, 95\% CIs are computed as estimate $\pm 1.96 \cdot$ SE on the log-odds scale, then exponentiated. Holm correction is applied across the two reader-group contrasts within each outcome (style, quality) and hypothesis (H1, H2).

\subsection*{S4.3: H3: AI Detection and Preference}

To test whether AI detectability influences preferences and whether fine-tuning removes this relationship, we model:
\begin{align}
\text{logit}\,P(Y_{ijk} = 1) = &\alpha + \beta_1 s_i + \beta_2 \mathbf{1}[\text{Setting}_i = \text{FT}] + \beta_3 \mathbf{1}[J_j = \text{College-educated}] \nonumber \\
&+ \beta_{12} s_i \cdot \mathbf{1}[\text{Setting}_i = \text{FT}] + \beta_{13} s_i \cdot \mathbf{1}[J_j = \text{College-educated}] \nonumber \\
&+ \beta_{23} \mathbf{1}[\text{Setting}_i = \text{FT}] \cdot \mathbf{1}[J_j = \text{College-educated}] \nonumber \\
&+ \beta_{123} s_i \cdot \mathbf{1}[\text{Setting}_i = \text{FT}] \cdot \mathbf{1}[J_j = \text{College-educated}]
\end{align}

where $s_i$ is the Pangram score (continuous, 0-1) and Setting indicates in-context vs fine-tuned. The coefficient $\beta_1$ captures the detection penalty in the baseline condition—how much AI detectability reduces preference. The interaction term $\beta_{12}$ tests whether fine-tuning attenuates this relationship, with a positive value indicating that fine-tuning removes the penalty associated with AI detection.

\subsection*{S4.4: Exploratory Analyses}

\subsubsection*{S4.4.1: Stylometric Mediation}
The mediation analysis examines which textual features explain the link between AI detection and human preferences. We follow a two-stage approach. First, we regress Pangram scores on stylometric features using elastic net regularization to handle multicollinearity:
\begin{equation}
s_i = \alpha + \sum_{k} \lambda_k X_{ik} + \epsilon_i
\end{equation}

where $X_{ik}$ includes cliché density, sentence-length variance, and parts-of-speech proportions. Features surviving regularization are then included alongside Pangram scores in the preference model to decompose the total effect into direct and mediated components. The proportion mediated is calculated using the product-of-coefficients method.

\subsubsection*{S4.4.2: Author-Level Heterogeneity}
To assess variability across the 30 fine-tuned authors, we compute empirical preference rates using Jeffreys prior estimates to stabilize small-sample authors:
\begin{equation}
\hat{p}_a = \frac{k_a + 0.5}{n_a + 1}
\end{equation}

with 95\% intervals from Beta($\frac{1}{2}$, $\frac{1}{2}$) priors. The relationship between preference rates and fine-tuning corpus size (in millions of tokens) is assessed via OLS regression with heteroskedasticity-robust standard errors.

\subsection*{S4.5: Mapping to Figures}

The model outputs map directly to figures in the main text:
\begin{itemize}
\item \textbf{Figures 2A-B}: Exponentiated contrasts $\text{OR}^{\text{H1}}_g$ and $\text{OR}^{\text{H2}}_g$ with 95\% CIs computed on log-odds scale then transformed
\item \textbf{Figures 2C-D}: Predicted probabilities $p(w,g) = \text{logit}^{-1}(\eta(w,g))$ with delta-method confidence intervals
\item \textbf{Figure 2F}: Model-implied probabilities from H3 evaluated at $s \in \{0.1, 0.3, 0.5, 0.7, 0.9\}$
\item \textbf{Figure 3}: Author-level preference rates $\hat{p}_a$ plotted against corpus size with OLS regression lines
\item \textbf{Figure 4A}: Stylometric mediation analysis showing proportion of Pangram effect explained by textual features
\item \textbf{Figure 4B}: Fine-tuning premium (difference in AI preference between fine-tuned and in-context) versus corpus size
\item \textbf{Figure 4C}: Cost comparison for producing 100,000 words of text
\end{itemize}

Inter-rater agreement quantifies consistency beyond chance using Fleiss' kappa, appropriate here as each pair was evaluated by multiple readers (3 MFA-trained and 19 in-context/21 fine-tuning college-educated general readers):
\begin{equation}
\kappa = \frac{\bar{P} - \bar{P}_e}{1 - \bar{P}_e}
\end{equation}

where $\bar{P}$ represents mean observed agreement and $\bar{P}_e$ expected agreement by chance. MFA-trained readers achieved moderate agreement ($\kappa = 0.41-0.67$) while college-educated general readers showed minimal agreement ($\kappa = 0.08–0.26$), reflecting the subjective nature of literary evaluation.

Full code and exact package versions used to generate Figures 2-4 are provided in the Code \& Data Availability section (S10) and the OSF repository.

\subsection*{S5.1: Cell Counts by Experimental Conditions}

Unless noted otherwise, counts below refer to \emph{reader-level long-format observations} entering the logistic models (two rows per judgment—one row per alternative in the pair). In the in-context prompting setting, \textbf{150} human--AI pairs were evaluated (evenly split across three baseline models: 50 Human vs GPT-4o, 50 Human vs Claude 3.5 Sonnet, and 50 Human vs Gemini 1.5 Pro). Because each pair was evaluated by 3 MFA-trained readers and 19 college-educated general readers, this yields \textbf{3,300} judgments per outcome and \textbf{6,600} long-format rows. In the fine-tuned setting, \textbf{90} human--AI pairs were evaluated; with 3 MFA-trained readers and 21 college-educated general readers per pair, this yields \textbf{2,160} judgments per outcome and \textbf{4,320} long-format rows. Human rows are three times more frequent than any single AI baseline in the in-context prompting setting because every pair contains a Human excerpt, while the 150 pairs are split evenly across the three baseline models. In the fine-tuned setting, Human and GPT-4o fine-tuned rows occur equally often.

\begin{table}[htbp]
\centering
\caption{Observation counts for stylistic fidelity in the in-context prompting setting. Each row is a long-format observation entering the H1 analysis.}
\label{tab:s5.1a}
\begin{tabular}{llllr}
\toprule
Outcome & Setting & Writer type & Reader type & $n$ \\
\midrule
style & In\_Context & Human & MFA-trained & 450 \\
style & In\_Context & Human & College-educated & 2850 \\
style & In\_Context & GPT-4o\_baseline & MFA-trained & 150 \\
style & In\_Context & GPT-4o\_baseline & College-educated & 950 \\
style & In\_Context & Claude\_baseline & MFA-trained & 150 \\
style & In\_Context & Claude\_baseline & College-educated & 950 \\
style & In\_Context & Gemini\_baseline & MFA-trained & 150 \\
style & In\_Context & Gemini\_baseline & College-educated & 950 \\
\bottomrule
\end{tabular}
\end{table}

\begin{table}[htbp]
\centering
\caption{Observation counts for stylistic fidelity in the fine-tuned setting. Balanced counts (Human = GPT-4o\_finetuned) reflect the one-to-one comparison design for H2.}
\label{tab:s5.1b}
\begin{tabular}{llllr}
\toprule
Outcome & Setting & Writer type & Reader type & $n$ \\
\midrule
style & Fine\_tuned & Human & MFA-trained & 270 \\
style & Fine\_tuned & Human & College-educated & 1890 \\
style & Fine\_tuned & GPT-4o\_finetuned & MFA-trained & 270 \\
style & Fine\_tuned & GPT-4o\_finetuned & College-educated & 1890 \\
\bottomrule
\end{tabular}
\end{table}

\begin{table}[htbp]
\centering
\caption{Observation counts for writing quality in the in-context prompting setting (same Human:AI baseline ratio as stylistic fidelity).}
\label{tab:s5.1c}
\begin{tabular}{llllr}
\toprule
Outcome & Setting & Writer type & Reader type & $n$ \\
\midrule
quality & In\_Context & Human & MFA-trained & 450 \\
quality & In\_Context & Human & College-educated & 2850 \\
quality & In\_Context & GPT-4o\_baseline & MFA-trained & 150 \\
quality & In\_Context & GPT-4o\_baseline & College-educated & 950 \\
quality & In\_Context & Claude\_baseline & MFA-trained & 150 \\
quality & In\_Context & Claude\_baseline & College-educated & 950 \\
quality & In\_Context & Gemini\_baseline & MFA-trained & 150 \\
quality & In\_Context & Gemini\_baseline & College-educated & 950 \\
\bottomrule
\end{tabular}
\end{table}

\begin{table}[htbp]
\centering
\caption{Observation counts for writing quality in the fine-tuned setting.}
\label{tab:s5.1d}
\begin{tabular}{llllr}
\toprule
Outcome & Setting & Writer type & Reader type & $n$ \\
\midrule
quality & Fine\_tuned & Human & MFA-trained & 270 \\
quality & Fine\_tuned & Human & College-educated & 1890 \\
quality & Fine\_tuned & GPT-4o\_finetuned & MFA-trained & 270 \\
quality & Fine\_tuned & GPT-4o\_finetuned & College-educated & 1890 \\
\bottomrule
\end{tabular}
\end{table}

\subsection*{S5.2: Writer Category Mapping}

Table \ref{tab:s5.2} provides the mapping from raw data labels to the analysis categories used in Section S4. This mapping ensures reproducibility and clarifies the distinction between baseline (in-context) and fine-tuned AI conditions.

\begin{table}[htbp]
\centering
\caption{Mapping of writer categories from raw data fields to analysis labels.}
\label{tab:s5.2}
\begin{tabular}{l l p{6cm}}
\toprule
Writer category & Description & Mapping rule (from raw fields) \\
\midrule
Human & Human-written (MFA authors) & excerpt\_type = \texttt{Human} \\
GPT-4o\_baseline & GPT-4o in-context prompting & AI excerpt; excerpt\_model $\in$ \{\texttt{GPT-4o}\}; setting = \texttt{In\_Context} \\
Claude\_baseline & Claude 3.5 Sonnet (in-context) & AI excerpt; excerpt\_model = \texttt{Claude3.5Sonnet}; setting = \texttt{In\_Context} \\
Gemini\_baseline & Gemini 1.5 Pro (in-context) & AI excerpt; excerpt\_model = \texttt{Gemini\_1.5\_Pro}; setting = \texttt{In\_Context} \\
GPT-4o\_finetuned & GPT-4o fine-tuned & AI excerpt; excerpt\_model = \texttt{GPT-4o\_Finetuned} \\
\bottomrule
\end{tabular}
\end{table}

\section*{S6: Main GLM Results and Contrasts (Figure 2)}

This section presents the numerical results from the logistic regression models specified in Section S4. All models use CR2 cluster-robust standard errors with readers as the clustering unit. The reference category throughout is Human $\times$ MFA-trained. These tables provide the complete statistical foundation for Figure 2 in the main text.

\subsection*{S6.1: Model Coefficients}

Table \ref{tab:s6.1} reports the full coefficient tables for all four primary models. The dramatic sign reversal between in-context prompting and fine-tuned settings is immediately apparent: negative coefficients for AI writers in in-context prompting (indicating human preference) become positive in fine-tuning (indicating AI preference). 

The writer type $\times$ reader type interactions in the in-context prompting models reveal that college-educated general readers are substantially more favorable to AI-generated text than MFA-trained readers, and the updated fine-tuned models show that this difference persists after fine-tuning rather than disappearing (style p=$0.021$; quality p=$6.41 \times 10^-{5}$).

\begin{table}[htbp]
\centering
\caption{GLM coefficients with CR2 robust standard errors for all primary models. Negative coefficients indicate preference for the reference category (Human), while positive coefficients indicate preference for AI. The sign reversal between in-context prompting and fine-tuned models represents the core finding. In fine-tuned models, the writer-type $\times$ reader-type interaction is significant, indicating that college-educated general readers favor AI by a wider margin than MFA-trained readers.}
\label{tab:s6.1}
\begin{tabular}{l r r r r}
\toprule
Term & Est.\ (log-odds) & SE & $z$ & $p$ \\
\midrule
\multicolumn{5}{l}{\emph{Style --- In-context Prompting (n = 6{,}600)}}\\
(Intercept) & 0.901 & 0.148 & 6.098 & 1.08e-09 \\
writer\_typeGPT-4o\_baseline & $-$1.655 & 0.402 & $-$4.121 & 3.78e-05 \\
writer\_typeClaude\_baseline & $-$1.390 & 0.276 & $-$5.039 & 4.67e-07 \\
writer\_typeGemini\_baseline & $-$2.510 & 0.426 & $-$5.894 & 3.77e-09 \\
reader\_typeCollege-educated & $-$0.929 & 0.153 & $-$6.053 & 1.43e-09 \\
writer\_typeGPT-4o\_baseline:reader\_typeCollege-educated & 1.784 & 0.414 & 4.307 & 1.66e-05 \\
writer\_typeClaude\_baseline:reader\_typeCollege-educated & 1.583 & 0.292 & 5.412 & 6.22e-08 \\
writer\_typeGemini\_baseline:reader\_typeCollege-educated & 2.357 & 0.438 & 5.384 & 7.27e-08 \\
\midrule
\multicolumn{5}{l}{\emph{Style --- Fine-tuned (n = 4{,}320)}}\\
(Intercept) & $-$1.050 & 0.141 & $-$7.435 & 1.04e-13 \\
writer\_typeGPT-4o\_finetuned & 2.100 & 0.282 & 7.435 & 1.04e-13 \\
reader\_typeCollege-educated & $-$0.356 & 0.154 & $-$2.308 & 0.021 \\
writer\_typeGPT-4o\_finetuned:reader\_typeCollege-educated & 0.713 & 0.309 & 2.308 & 0.021 \\
\midrule
\multicolumn{5}{l}{\emph{Quality --- In-context Prompting (n = 6{,}600)}}\\
(Intercept) & 0.978 & 0.174 & 5.634 & 1.76e-08 \\
writer\_typeGPT-4o\_baseline & $-$2.406 & 0.473 & $-$5.083 & 3.71e-07 \\
writer\_typeClaude\_baseline & $-$1.246 & 0.368 & $-$3.390 & 7.00e-04 \\
writer\_typeGemini\_baseline & $-$2.406 & 0.421 & $-$5.711 & 1.13e-08 \\
reader\_typeCollege-educated & $-$1.275 & 0.179 & $-$7.127 & 1.03e-12 \\
writer\_typeGPT-4o\_baseline:reader\_typeCollege-educated & 3.197 & 0.484 & 6.601 & 4.08e-11 \\
writer\_typeClaude\_baseline:reader\_typeCollege-educated & 1.935 & 0.381 & 5.073 & 3.91e-07 \\
writer\_typeGemini\_baseline:reader\_typeCollege-educated & 2.716 & 0.433 & 6.268 & 3.65e-10 \\
\midrule
\multicolumn{5}{l}{\emph{Quality --- Fine-tuned (n = 4{,}320)}}\\
(Intercept) & $-$0.314 & 0.122 & $-$2.571 & 0.010 \\
writer\_typeGPT-4o\_finetuned & 0.627 & 0.244 & 2.571 & 0.010 \\
reader\_typeCollege-educated & $-$0.531 & 0.133 & $-$3.997 & 6.41e-05 \\
writer\_typeGPT-4o\_finetuned:reader\_typeCollege-educated & 1.062 & 0.266 & 3.997 & 6.41e-05 \\
\bottomrule
\end{tabular}
\end{table}

\subsection*{S6.2: Primary Hypothesis Tests}
Table \ref{tab:s6.2} presents the preregistered contrasts testing H1 and H2, corresponding to panels A--B in Figure 2. The odds ratios reveal a striking reversal: MFA-trained readers' 6--8 fold preference for human writing in in-context prompting conditions (OR = 0.16 for style, 0.13 for quality) transforms into an 8-fold preference for AI in fine-tuning (OR = 8.16 for style). College-educated general readers show directionally consistent but substantially larger shifts, particularly after fine-tuning (OR = 16.65 for style, 5.42 for quality), indicating that this reader group favors AI-generated text by a wider margin than MFA-trained readers.

\begin{table}[htbp]
{\small
\centering
\caption{Primary contrasts testing H1 (baseline LLMs vs Human) and H2 (fine-tuned vs Human). Holm correction applied within each outcome and hypothesis across reader types. The OR column shows odds ratios with values $<1$ favoring humans and $>1$ favoring AI. {\it Note:} The p and $p_{\mathrm{Holm}}$ columns report exact two-sided Wald p-values and their Holm-corrected counterparts. The main text reports Holm-corrected p-values using inequality bounds ($p < 10^{-k}$), capped at $p < 10^{-15}$ as a conservative floor reflecting the practical limits of the normal approximation. }
\label{tab:s6.2}
\begin{tabular}{l l l r r r r r l}
\toprule
Outcome & Hyp. & Reader & Est. & SE & $p$ & $p_{\mathrm{Holm}}$ & OR & 95\% CI \\
&&&(log-odds)&(log-odds)&&&exp(Est.)&(OR scale)\\
\midrule
style & H1 & MFA-trained & $-$1.852 & 0.314 & 3.65e-09 & 7.31e-09 & 0.157 & [0.08, 0.29] \\
style & H1 & College-educated & 0.056 & 0.083 & 0.501 & 0.501 & 1.058 & [0.90, 1.25] \\
style & H2 & MFA-trained & 2.100 & 0.282 & 1.04e-13 & 1.04e-13 & 8.163 & [4.69, 14.20] \\
style & H2 & College-educated & 2.813 & 0.125 & 4.91e-112 & 9.81e-112 & 16.652 & [13.03, 21.28] \\
quality & H1 & MFA-trained & $-$2.020 & 0.381 & 1.19e-07 & 1.19e-07 & 0.133 & [0.06, 0.28] \\
quality & H1 & College-educated & 0.597 & 0.087 & 6.00e-12 & 1.20e-11 & 1.816 & [1.53, 2.15] \\
quality & H2 & MFA-trained & 0.627 & 0.244 & 1.02e-02 & 1.02e-02 & 1.873 & [1.16, 3.02] \\
quality & H2 & College-educated & 1.690 & 0.105 & 4.63e-58 & 9.27e-58 & 5.417 & [4.41, 6.66] \\
\bottomrule
\end{tabular}
}
\end{table}

\subsection*{S6.3: Model-Predicted Probabilities}

Table \ref{tab:s6.3} reports the predicted probabilities displayed in Figure 2 panels C--D. These probabilities, derived from the inverse-logit transformation of linear predictors, show that while MFA-trained readers strongly prefer humans over all baseline models in in-context prompting settings (probabilities 0.17--0.43), fine-tuning elevates AI preference to 0.74 for MFA-trained readers and 0.80 for college-educated general readers on stylistic fidelity, and to 0.58 versus 0.70 on writing quality. Unlike the in-context condition, both reader groups now favor fine-tuned AI, but college-educated general readers do so by a significantly wider margin.

\begin{table}[htbp]
\centering
\caption{Predicted probabilities of selecting AI excerpts by model and reader type, corresponding to Figure 2C--D. Values above 0.5 indicate AI preference. After fine-tuning, both reader groups prefer AI, but college-educated general readers show higher predicted probabilities than MFA-trained readers.}
\label{tab:s6.3}
\begin{tabular}{l l l l r l}
\toprule
Outcome & Setting & Model & Reader & $\hat{p}$ & 95\% CI \\
\midrule
style & In\_Context & GPT-4o (In-Context) & MFA-trained & 0.320 & [0.21, 0.45] \\
style & In\_Context & GPT-4o (In-Context) & College-educated & 0.525 & [0.49, 0.56] \\
style & In\_Context & Claude (In-Context) & MFA-trained & 0.380 & [0.30, 0.46] \\
style & In\_Context & Claude (In-Context) & College-educated & 0.541 & [0.51, 0.57] \\
style & In\_Context & Gemini (In-Context) & MFA-trained & 0.167 & [0.10, 0.27] \\
style & In\_Context & Gemini (In-Context) & College-educated & 0.455 & [0.42, 0.49] \\
style & Fine\_tuned & GPT-4o (Fine-tuned) & MFA-trained & 0.741 & [0.68, 0.79] \\
style & Fine\_tuned & GPT-4o (Fine-tuned) & College-educated & 0.803 & [0.78, 0.82] \\
quality & In\_Context & GPT-4o (In-Context) & MFA-trained & 0.193 & [0.11, 0.31] \\
quality & In\_Context & GPT-4o (In-Context) & College-educated & 0.621 & [0.59, 0.65] \\
quality & In\_Context & Claude (In-Context) & MFA-trained & 0.433 & [0.33, 0.54] \\
quality & In\_Context & Claude (In-Context) & College-educated & 0.597 & [0.56, 0.63] \\
quality & In\_Context & Gemini (In-Context) & MFA-trained & 0.193 & [0.12, 0.29] \\
quality & In\_Context & Gemini (In-Context) & College-educated & 0.503 & [0.47, 0.54] \\
quality & Fine\_tuned & GPT-4o (Fine-tuned) & MFA-trained & 0.578 & [0.52, 0.64] \\
quality & Fine\_tuned & GPT-4o (Fine-tuned) & College-educated & 0.699 & [0.68, 0.72] \\
\bottomrule
\end{tabular}
\end{table}

\subsection*{S6.4: Interaction Diagnostics}

Tables \ref{tab:s6.4a} and \ref{tab:s6.4b} examine the writer-type $\times$ reader-type interactions in detail. The significant interactions in in-context prompting models (all $p < 0.001$) quantify college-educated general readers' greater tolerance for AI-generated text. Unlike the previous analysis with a smaller college-educated sample, the interactions in fine-tuned models are now also significant (style $p = 0.021$; quality $p = 6.41 \times 10^{-5}$), indicating that college-educated general readers favor fine-tuned AI by a wider margin than MFA-trained readers rather than converging to similar preference levels.

\begin{table}[htbp]
\centering
{\small
\caption{Individual interaction terms showing differential preferences between MFA-trained and college-educated general readers. Large positive coefficients indicate college-educated general readers are more favorable to AI. In fine-tuned models, interactions are significant, indicating that reader-group differences persist after fine-tuning.}
\label{tab:s6.4a}
\begin{tabular}{l l l r r r r}
\toprule
Outcome & Setting & Term & Est. & SE & $z$ & $p$ \\
\midrule
style & In\_Context & GPT-4o\_baseline:reader\_typeCollege-educated & 1.784 & 0.414 & 4.307 & 1.66e-05 \\
style & In\_Context & Claude\_baseline:reader\_typeCollege-educated & 1.583 & 0.292 & 5.412 & 6.22e-08 \\
style & In\_Context & Gemini\_baseline:reader\_typeCollege-educated & 2.357 & 0.438 & 5.384 & 7.27e-08 \\
style & Fine\_tuned & GPT-4o\_finetuned:reader\_typeCollege-educated & 0.713 & 0.309 & 2.308 & 0.021 \\
quality & In\_Context & GPT-4o\_baseline:reader\_typeCollege-educated & 3.197 & 0.484 & 6.601 & 4.08e-11 \\
quality & In\_Context & Claude\_baseline:reader\_typeCollege-educated & 1.935 & 0.381 & 5.073 & 3.91e-07 \\
quality & In\_Context & Gemini\_baseline:reader\_typeCollege-educated & 2.716 & 0.433 & 6.268 & 3.65e-10 \\
quality & Fine\_tuned & GPT-4o\_finetuned:reader\_typeCollege-educated & 1.062 & 0.266 & 3.997 & 6.41e-05 \\
\bottomrule
\end{tabular}
}
\end{table}

\begin{table}[htbp]
\centering
\caption{Joint Wald tests for writer-type $\times$ reader-type interactions. Interactions are significant in both in-context prompting and fine-tuned models, indicating that college-educated general readers are more favorable to AI than MFA-trained readers in both conditions.}
\label{tab:s6.4b}
\begin{tabular}{l l l r r r}
\toprule
Outcome & Setting & Term & df & $\chi^2$ & $p$ \\
\midrule
style & In\_Context & writer\_type:reader\_type & 3 & 36.606 & 5.57e-08 \\
style & Fine\_tuned & writer\_type:reader\_type & 1 & 5.328 & 0.021 \\
quality & In\_Context & writer\_type:reader\_type & 3 & 46.884 & 3.68e-10 \\
quality & Fine\_tuned & writer\_type:reader\_type & 1 & 15.978 & 6.41e-05 \\
\bottomrule
\end{tabular}
\end{table}

The results in this section provide compelling statistical evidence for the paper's central finding: fine-tuning on author-specific corpora fundamentally transforms AI-generated text from clearly inferior (as judged by experts) to preferred over human writing. Rather than converging after fine-tuning, both reader groups prefer fine-tuned AI, but college-educated general readers do so by a significantly wider margin than MFA-trained readers. This pattern suggests that fine-tuning produces improvements that are meaningful to both audiences while leaving a persistent difference in how strongly each audience favors the AI output.

% % =========================
% % S7: Author-Level Heterogeneity (Figure 3)
% % =========================
\section*{S7: Author-Level Heterogeneity (Figure 3)}

This section examines variation in AI preference rates across the 30 fine-tuned authors. Despite substantial differences in training corpus sizes (0.89M to 10.9M tokens) and fine-tuning costs (\$22 to \$273), we find no detectable relationship between data quantity and model performance within this token range, suggesting that even authors with limited published works can be effectively emulated.

\subsection*{S7.1: Per-Author Success Rates}

Table~\ref{tab:s7.1} presents author-level AI preference rates using Jeffreys prior estimates to stabilize small-sample authors. Each author now has 72 pooled trials (9 from MFA-trained readers and 63 from college-educated general readers), so the pooled rates are weighted approximately 87.5\% toward the college-educated group. The results reveal striking heterogeneity. For \textbf{style}, AI preference rates range from 18\% (Tony Tulathimutte) to 98\% (Roxane Gay), with 29 of 30 authors showing AI superiority (rate $>$ 0.5). For \textbf{quality}, the range spans 49\% (Kazuo Ishiguro and Min Jin Lee) to 92\% (Cheryl Strayed), with 28 of 30 authors favoring AI.

Notably, some authors show divergent patterns across outcomes. Tony Tulathimutte represents an extreme case: readers found his \emph{style} uniquely difficult to emulate (18\% AI preference) yet rated AI \emph{quality} as acceptable (55\%). This suggests certain idiosyncratic voices resist algorithmic mimicry even when technical writing competence is achieved.

\begingroup
\scriptsize
\setlength{\tabcolsep}{4pt}
\renewcommand{\arraystretch}{1.05}
\begin{longtable}{llrrrr}
\caption{Per-author AI preference rates (Jeffreys-smoothed and raw) ranked by performance. Values above 0.5 indicate AI preference. Each author has 72 pooled trials (9 MFA-trained, 63 college-educated general readers). Note the wide variation across authors and the divergence between \textbf{style} and quality rankings for some authors.}
\label{tab:s7.1}\\
\toprule
Outcome & Author & Rank & AI Win Rate (Jeffreys) & Human Win Rate (Jeffreys) & AI Win Rate (Raw) \\
\midrule
\endfirsthead
\toprule
Outcome & Author & Rank & AI Win Rate (Jeffreys) & Human Win Rate (Jeffreys) & AI Win Rate (Raw) \\
\midrule
\endhead
\midrule
\multicolumn{6}{r}{{Continued on next page}} \\
\midrule
\endfoot
\bottomrule
\endlastfoot
quality & Cheryl Strayed & 1 & 0.92 & 0.08 & 0.931 \\
quality & Marilynne Robinson & 2 & 0.91 & 0.09 & 0.917 \\
quality & Han Kang & 3 & 0.88 & 0.12 & 0.889 \\
quality & Salman Rushdie & 4 & 0.84 & 0.16 & 0.847 \\
quality & Lydia Davis & 5 & 0.80 & 0.20 & 0.806 \\
quality & Zadie Smith & 6 & 0.80 & 0.20 & 0.806 \\
quality & Haruki Murakami & 7 & 0.77 & 0.23 & 0.778 \\
quality & Junot Diaz & 8 & 0.77 & 0.23 & 0.778 \\
quality & Orhan Pamuk & 9 & 0.77 & 0.23 & 0.778 \\
quality & Sigrid Nunez & 10 & 0.76 & 0.24 & 0.764 \\
quality & Colson Whitehead & 11 & 0.73 & 0.27 & 0.736 \\
quality & Roxane Gay & 12 & 0.71 & 0.29 & 0.708 \\
quality & Percival Everett & 13 & 0.69 & 0.31 & 0.694 \\
quality & Jhumpa Lahiri & 14 & 0.68 & 0.32 & 0.681 \\
quality & Rachel Cusk & 15 & 0.68 & 0.32 & 0.681 \\
quality & Ben Lerner & 16 & 0.66 & 0.34 & 0.667 \\
quality & Margaret Atwood & 17 & 0.66 & 0.34 & 0.667 \\
quality & Louise Erdrich & 18 & 0.65 & 0.35 & 0.653 \\
quality & George Saunders & 19 & 0.64 & 0.36 & 0.639 \\
quality & Jonathan Franzen & 20 & 0.64 & 0.36 & 0.639 \\
quality & Sally Rooney & 21 & 0.64 & 0.36 & 0.639 \\
quality & Annie Proulx & 22 & 0.60 & 0.40 & 0.597 \\
quality & Chimamanda Ngozi Adichie & 23 & 0.58 & 0.42 & 0.583 \\
quality & Annie Ernaux & 24 & 0.55 & 0.45 & 0.556 \\
quality & Tony Tulathimutte & 25 & 0.55 & 0.45 & 0.556 \\
quality & Yoko Ogawa & 26 & 0.54 & 0.46 & 0.542 \\
quality & Ian McEwan & 27 & 0.51 & 0.49 & 0.514 \\
quality & Ottessa Moshfegh & 28 & 0.51 & 0.49 & 0.514 \\
quality & Kazuo Ishiguro & 29 & 0.49 & 0.51 & 0.486 \\
quality & Min Jin Lee & 30 & 0.49 & 0.51 & 0.486 \\
\midrule
style & Roxane Gay & 1 & 0.98 & 0.02 & 0.986 \\
style & Margaret Atwood & 2 & 0.97 & 0.03 & 0.972 \\
style & Han Kang & 3 & 0.95 & 0.05 & 0.958 \\
style & Ben Lerner & 4 & 0.94 & 0.06 & 0.944 \\
style & Chimamanda Ngozi Adichie & 5 & 0.94 & 0.06 & 0.944 \\
style & Kazuo Ishiguro & 6 & 0.94 & 0.06 & 0.944 \\
style & Lydia Davis & 7 & 0.94 & 0.06 & 0.944 \\
style & Marilynne Robinson & 8 & 0.94 & 0.06 & 0.944 \\
style & Sigrid Nunez & 9 & 0.92 & 0.08 & 0.931 \\
style & Cheryl Strayed & 10 & 0.91 & 0.09 & 0.917 \\
style & Junot Diaz & 11 & 0.91 & 0.09 & 0.917 \\
style & Orhan Pamuk & 12 & 0.91 & 0.09 & 0.917 \\
style & Min Jin Lee & 13 & 0.86 & 0.14 & 0.861 \\
style & Sally Rooney & 14 & 0.86 & 0.14 & 0.861 \\
style & Colson Whitehead & 15 & 0.83 & 0.17 & 0.833 \\
style & Salman Rushdie & 16 & 0.83 & 0.17 & 0.833 \\
style & George Saunders & 17 & 0.80 & 0.20 & 0.806 \\
style & Ian McEwan & 18 & 0.77 & 0.23 & 0.778 \\
style & Haruki Murakami & 19 & 0.76 & 0.24 & 0.764 \\
style & Rachel Cusk & 20 & 0.76 & 0.24 & 0.764 \\
style & Percival Everett & 21 & 0.75 & 0.25 & 0.750 \\
style & Jhumpa Lahiri & 22 & 0.71 & 0.29 & 0.708 \\
style & Zadie Smith & 23 & 0.71 & 0.29 & 0.708 \\
style & Annie Proulx & 24 & 0.66 & 0.34 & 0.667 \\
style & Jonathan Franzen & 25 & 0.66 & 0.34 & 0.667 \\
style & Ottessa Moshfegh & 26 & 0.65 & 0.35 & 0.653 \\
style & Yoko Ogawa & 27 & 0.61 & 0.39 & 0.611 \\
style & Louise Erdrich & 28 & 0.58 & 0.42 & 0.583 \\
style & Annie Ernaux & 29 & 0.51 & 0.49 & 0.514 \\
style & Tony Tulathimutte & 30 & 0.18 & 0.82 & 0.181 \\
\end{longtable}
\endgroup

\noindent\emph{Note.} Jeffreys prior adds 0.5 to successes and 0.5 to failures; differences from raw are small but avoid 0/1 edge cases.

\subsection*{S7.2: Relationship with Corpus Size}

Table~\ref{tab:s7.2} presents OLS regression results examining the relationship between fine-tuning corpus size (ranging from 0.89M tokens for Tulathimutte to 10.9M for Pamuk) and AI preference rates. The near-zero slopes with confidence intervals spanning zero and minimal $R^2$ values indicate no detectable relationship within this token range between training data quantity and model performance.

\begin{table}[htbp]
\centering
\scriptsize
\setlength{\tabcolsep}{6pt}
\caption{OLS regression of author-level AI win rate on fine-tuning corpus size (millions of tokens). Slopes are near zero with CIs spanning zero and very low $R^2$, indicating no detectable relationship between corpus size and AI preference rate.}
\label{tab:s7.2}
\begin{tabular}{lrrrrrrrr}
\toprule
Outcome & Slope & Slope CI (Low) & Slope CI (High) & Intercept & Intercept CI (Low) & Intercept CI (High) & $R^2$ & N Authors \\
\midrule
quality  & 0.0029 & $-$0.0173 & 0.0231 & 0.676 & 0.580 & 0.771 & 0.0031 & 30 \\
style    & 0.0089 & $-$0.0194 & 0.0373 & 0.760 & 0.626 & 0.894 & 0.0153 & 30 \\
\bottomrule
\end{tabular}
\vspace{0.25em}
\footnotesize\emph{Non-parametric check.} Spearman $\rho=0.052$ ($p=0.791$) for quality and $\rho=-0.040$ ($p=0.839$) for style.
\end{table}

This finding has economic implications: authors with limited published works (costing \$22--\$50 to fine-tune) can be emulated as effectively as prolific authors (costing \$200+). The absence of any detectable corpus-size effect ($R^2\approx 0$) suggests that capturing an author's voice depends more on stylistic consistency than sheer data volume.

% =========================
% S8: Detection Mechanisms and Economic Analysis (Figure 4)
% =========================
\section*{S8: Detection Mechanisms and Economic Analysis (Figure 4)}

This section examines the mechanisms linking AI detectability to human preferences (H3) and quantifies the economic implications of fine-tuning. We show that fine-tuning reverses and substantially attenuates the negative association between AI detection and preference observed in in-context prompting, while achieving performance that readers prefer at $\approx$99.7\% lower cost than human writers.

\subsection*{S8.1: AI Detection and Preference Relationship}

Table~\ref{tab:s8.1} presents the full model examining how AI detectability (Pangram score) influences preferences. The key finding is the significant interaction between \emph{pangram\_score} and \emph{setting} (quality: $\hat\beta\approx 2.90$, $p<0.001$), indicating that fine-tuning attenuates the negative association between detectability and preference observed in in-context prompting. The significant three-way interactions (pangram\_score $\times$ setting $\times$ reader\_type) further indicate that the detectability penalty and its attenuation differ across reader groups, with college-educated general readers showing less sensitivity to detectability than MFA-trained readers even before fine-tuning.

\begingroup
\scriptsize
\setlength{\tabcolsep}{5pt}
\renewcommand{\arraystretch}{1.05}
\begin{longtable}{lrrlrrrrrrr}
\caption{Pangram GLM coefficients with CR2 robust standard errors (clustered by reader). Negative coefficients for \emph{pangram\_score} in in-context prompting indicate detection reduces preference; positive interactions with \emph{setting} show that fine-tuning attenuates this penalty. Significant three-way interactions indicate the detectability penalty differs across reader groups.}
\label{tab:s8.1}\\
\toprule
Outcome & $N$ & $N$ Readers & Term & Estimate & SE & $z$ & $p$ & OR & OR (Low) & OR (High) \\
\midrule
\endfirsthead
\toprule
Outcome & $N$ & $N$ Readers & Term & Estimate & SE & $z$ & $p$ & OR & OR (Low) & OR (High) \\
\midrule
\endhead
\midrule
\multicolumn{11}{r}{Continued on next page} \\
\midrule
\endfoot
\bottomrule
\endlastfoot
quality & 10920 & 542 & (Intercept) & 0.990 & 0.169 & 5.87 & 4.43e-09 & 2.69 & 1.93 & 3.74 \\
quality & 10920 & 542 & \textit{pangram\_score} & $-$2.010 & 0.333 & $-$6.03 & 1.65e-09 & 0.13 & 0.07 & 0.26 \\
quality & 10920 & 542 & \textit{setting} (Fine-tuned) & $-$1.025 & 0.175 & $-$5.86 & 4.64e-09 & 0.36 & 0.26 & 0.51 \\
quality & 10920 & 542 & \textit{reader\_type} (College-educated) & $-$1.288 & 0.174 & $-$7.40 & 1.36e-13 & 0.28 & 0.20 & 0.39 \\
quality & 10920 & 542 & pangram\_score $\times$ setting & 2.904 & 0.876 & 3.32 & 9.13e-04 & 18.20 & 3.28 & 102.00 \\
quality & 10920 & 542 & pangram\_score $\times$ reader\_type & 2.613 & 0.344 & 7.59 & 3.24e-14 & 13.60 & 6.94 & 26.80 \\
quality & 10920 & 542 & setting $\times$ reader\_type & 1.313 & 0.180 & 7.28 & 3.34e-13 & 3.72 & 2.61 & 5.29 \\
quality & 10920 & 542 & pangram\_score $\times$ setting $\times$ reader\_type & $-$3.258 & 0.908 & $-$3.59 & 3.31e-04 & 0.04 & 0.01 & 0.23 \\
style & 10920 & 535 & (Intercept) & 0.909 & 0.146 & 6.23 & 4.79e-10 & 2.48 & 1.86 & 3.31 \\
style & 10920 & 535 & \textit{pangram\_score} & $-$1.845 & 0.292 & $-$6.32 & 2.57e-10 & 0.16 & 0.09 & 0.28 \\
style & 10920 & 535 & \textit{setting} (Fine-tuned) & $-$0.938 & 0.158 & $-$5.95 & 2.65e-09 & 0.39 & 0.29 & 0.53 \\
style & 10920 & 535 & \textit{reader\_type} (College-educated) & $-$0.925 & 0.152 & $-$6.10 & 1.09e-09 & 0.40 & 0.29 & 0.53 \\
style & 10920 & 535 & pangram\_score $\times$ setting & 2.556 & 0.810 & 3.15 & 1.61e-03 & 12.90 & 2.63 & 63.00 \\
style & 10920 & 535 & pangram\_score $\times$ reader\_type & 1.878 & 0.304 & 6.18 & 6.26e-10 & 6.54 & 3.61 & 11.90 \\
style & 10920 & 535 & setting $\times$ reader\_type & 0.921 & 0.163 & 5.65 & 1.63e-08 & 2.51 & 1.82 & 3.46 \\
style & 10920 & 535 & pangram\_score $\times$ setting $\times$ reader\_type & $-$1.766 & 0.844 & $-$2.09 & 3.64e-02 & 0.17 & 0.03 & 0.89 \\
\end{longtable}
\endgroup
\noindent\emph{Notes.} CR2 robust standard errors clustered by reader. Readers: 28 MFA-trained and 516 college-educated general readers.

\subsection*{S8.2: Predicted Probabilities Across Detection Levels}

Tables~\ref{tab:s8.2a} and \ref{tab:s8.2b} show model-predicted probabilities at different AI detection levels, corresponding to Figure~2F. For MFA-trained readers evaluating \emph{quality}, high detection scores (0.9) reduce AI preference to $\sim$31\% in in-context prompting but have a point estimate of $\sim$68\% after fine-tuning (95\% CI $0.37$--$0.89$), demonstrating that fine-tuning makes outputs robust to detection-based skepticism. College-educated general readers show a weaker detectability penalty in the in-context condition and a modest positive slope after fine-tuning, with predicted probabilities reaching 0.55--0.67 at high detection scores.

\begin{table}[htbp]
\centering
\scriptsize
\setlength{\tabcolsep}{8pt}
\caption{Predicted AI preference probabilities for \textbf{style} at different Pangram detection scores. Fine-tuning attenuates the relationship with detection score for MFA-trained readers (with wider uncertainty at high detection). College-educated general readers show a weak relationship with detectability in the in-context condition and a positive slope after fine-tuning.}
\label{tab:s8.2a}
\begin{tabular}{rllrrr}
\toprule
Bin & Setting & Reader Type & Prob & Prob (Low) & Prob (High) \\
\midrule
0.1 & In-context Prompting & MFA-trained & 0.674 & 0.621 & 0.722 \\
0.5 & In-context Prompting & MFA-trained & 0.497 & 0.494 & 0.500 \\
0.9 & In-context Prompting & MFA-trained & 0.320 & 0.273 & 0.372 \\
0.1 & Fine-tuned & MFA-trained & 0.511 & 0.491 & 0.530 \\
0.5 & Fine-tuned & MFA-trained & 0.581 & 0.435 & 0.714 \\
0.9 & Fine-tuned & MFA-trained & 0.648 & 0.380 & 0.847 \\
0.1 & In-context Prompting & College-educated & 0.497 & 0.481 & 0.513 \\
0.5 & In-context Prompting & College-educated & 0.500 & 0.500 & 0.500 \\
0.9 & In-context Prompting & College-educated & 0.503 & 0.487 & 0.520 \\
0.1 & Fine-tuned & College-educated & 0.512 & 0.505 & 0.519 \\
0.5 & Fine-tuned & College-educated & 0.593 & 0.544 & 0.641 \\
0.9 & Fine-tuned & College-educated & 0.670 & 0.581 & 0.748 \\
\bottomrule
\end{tabular}
\end{table}

\begin{table}[htbp]
\centering
\scriptsize
\setlength{\tabcolsep}{8pt}
\caption{Predicted AI preference probabilities for \textbf{quality}. MFA-trained readers show strong detection sensitivity in in-context prompting ($\sim$0.69 $\to$ $\sim$0.31) that attenuates after fine-tuning (point estimate at 0.9 $\sim$0.68; 95\% CI 0.37--0.89). College-educated general readers show a weaker in-context penalty and modest positive slopes after fine-tuning.}
\label{tab:s8.2b}
\begin{tabular}{rllrrr}
\toprule
Bin & Setting & Reader Type & Prob & Prob (Low) & Prob (High) \\
\midrule
0.1 & In-context Prompting & MFA-trained & 0.688 & 0.628 & 0.742 \\
0.5 & In-context Prompting & MFA-trained & 0.496 & 0.493 & 0.499 \\
0.9 & In-context Prompting & MFA-trained & 0.306 & 0.254 & 0.363 \\
0.1 & Fine-tuned & MFA-trained & 0.514 & 0.490 & 0.537 \\
0.5 & Fine-tuned & MFA-trained & 0.602 & 0.431 & 0.751 \\
0.9 & Fine-tuned & MFA-trained & 0.683 & 0.373 & 0.887 \\
0.1 & In-context Prompting & College-educated & 0.441 & 0.424 & 0.457 \\
0.5 & In-context Prompting & College-educated & 0.501 & 0.500 & 0.501 \\
0.9 & In-context Prompting & College-educated & 0.561 & 0.544 & 0.578 \\
0.1 & Fine-tuned & College-educated & 0.504 & 0.497 & 0.510 \\
0.5 & Fine-tuned & College-educated & 0.529 & 0.479 & 0.578 \\
0.9 & Fine-tuned & College-educated & 0.553 & 0.461 & 0.643 \\
\bottomrule
\end{tabular}
\end{table}

\subsection*{S8.3: Stylometric Correlates and Mediation}

Table~\ref{tab:s8.3} shows that clich\'{e} density has the strongest correlation with AI detection (Pearson $r=0.60$), while readability ease is negatively correlated ($r=-0.23$), suggesting AI text is marked by formulaic phrases but simpler syntax. Among MFA-trained readers, where the detectability penalty is concentrated, approximately 16\% of the detection effect on preference is mediated through clich\'{e} density in the in-context prompting condition; after fine-tuning, this mediation drops to a statistically insignificant 1.3\%, indicating that fine-tuning substantially reduces rather than merely masks these stylistic signatures. Because college-educated general readers did not penalize AI-detectable text in either condition, mediation analysis does not apply to this group.

\begin{table}[htbp]
\centering
\scriptsize
\setlength{\tabcolsep}{8pt}
\caption{Correlations between Pangram AI detection scores and stylometric features. These correlations are computed at the excerpt level ($N=330$ unique excerpts) and are unchanged by the expanded reader sample.}
\label{tab:s8.3}
\begin{tabular}{lrrr}
\toprule
Metric & Pearson $r$ & Spearman $\rho$ & $N$ \\
\midrule
readability\_ease       & $-$0.232 & $-$0.312 & 330 \\
cliche\_density         &  0.600 &  0.602 & 330 \\
total\_adjective\_count &  0.104 &  0.129 & 330 \\
num\_cliches            &  0.614 &  0.633 & 330 \\
\bottomrule
\end{tabular}
\end{table}

\subsection*{S8.4: Economic Analysis}

Table~\ref{tab:s8.4} documents the cost structure for AI-based text generation. With fine-tuning costs ranging from \$22.25 to \$272.50 (median \$77.88) plus \$3 for inference to generate 100{,}000 words, the total AI cost represents approximately 0.3\% of the \$25{,}000 a professional writer would charge. This $\sim$99.7\% cost reduction, combined with reader preference for fine-tuned outputs, quantifies the potential economic disruption to creative writing markets.

\begin{table}[htbp]
\centering
\scriptsize
\setlength{\tabcolsep}{8pt}
\caption{Cost comparison for generating 100{,}000 words. Fine-tuning plus inference costs (\$25–\$276) represent about 0.1\% to 1.1\% of professional writer compensation (\$25,000), with a median of approximately 0.3\%.}
\label{tab:s8.4}
\begin{tabular}{lrrrrrr}
\toprule
$N$ Authors & Fine-tune Min & Fine-tune Med & Fine-tune Max & Total Min & Total Med & Total Max \\
\midrule
30 & \$22.25 & \$77.88 & \$272.50 & \$25.25 & \$80.88 & \$275.50 \\
\bottomrule
\end{tabular}
\vspace{0.25em}

\footnotesize\emph{Note.} Inference cost of \$3 per 100k words follows Figure 4c; per-token assumptions are documented in the repository.
\end{table}

\section*{S9. Deviations from Preregistration}

This section documents deviations from the preregistered analysis plan (https://osf.io/zt4ad). All other aspects were implemented exactly as specified.

\begin{itemize}

\item \textbf{Model framework.} The preregistration specified mixed-effects logistic regression with random intercepts for readers and prompts. We implemented logistic GLMs with CR2 cluster-robust standard errors (clustered by reader) due to convergence issues and singularity warnings in the mixed-effects models. Point estimates from both approaches were similar when convergence was achieved, and inference on the preregistered contrasts (H1, H2) remained unchanged.

\item \textbf{Sample sizes (exceeded targets).} We successfully recruited more participants than planned:
\begin{itemize}
    \item MFA-trained readers: 28 (planned: ~25)
    \item College-educated general readers: 516 (planned: 120)
    \item Fine-tuned authors: 30 (planned minimum: 10)
\end{itemize}
All analyses used the full realized sample. No stopping rules were applied or violated; the additional data strengthens the reliability of our findings.

\item \textbf{Additional analyses (not preregistered).} We added:
\begin{itemize}
    \item Writer type × Reader type interaction tests (Type-III Wald with robust covariance) for transparency
    \item Predicted probability visualizations (Figures 2C-D) to complement the odds ratio panels
\end{itemize}
These additions provide fuller context but do not alter the conclusions drawn from the preregistered H1 and H2 contrasts.

\end{itemize}

\section*{S10. Code and Data Availability}

All code and data necessary to reproduce Figures 2-4 and the associated statistical analyses are publicly available:

\begin{itemize}

\item \textbf{GitHub Repository:} \url{https://github.com/tuhinjubcse/Author-Style-Personalization/tree/main}

\item \textbf{Preregistration:} \url{https://osf.io/zt4ad} (Version 2, July 2025)

\item \textbf{Key Analysis Scripts:} The repository contains R scripts for data processing, model fitting, and figure generation. Core analyses can be reproduced by running the numbered scripts in sequence.

\item \textbf{Environment:} Analyses were conducted in R 4.3.1 with key packages including \texttt{clubSandwich} (CR2 robust SEs) and \texttt{emmeans} (contrasts). Full package versions are documented in the repository.

\item \textbf{Data:} Anonymized trial-level data in long format, with documentation of all variable definitions and transformations applied.

\end{itemize}

%%%%%%%%%%%%%%% SUPPLEMENTARY REFERENCES %%%%%%%%%%%%%%%

% Do NOT include a reference list in the supplement.
% All references must be in a single list at the end of the main text.
% The copyeditors will ensure that the correct reference list appears with each version of the paper
% (print, HTML, PDF, mobile app, metadata for bibliographic databases etc.)

\end{document}